%% file: main.tex
\pdfoutput=1

\documentclass[10pt,twocolumn,letterpaper]{article}

\usepackage[pagenumbers]{wacv}
\usepackage[accsupp]{axessibility}

\input{macros}

\usepackage{capt-of}
\usepackage{placeins}

\definecolor{wacvblue}{rgb}{0.21,0.49,0.74}
\usepackage[breaklinks,colorlinks,allcolors=wacvblue]{hyperref}

\def\wacvPaperID{3182}
\def\confName{WACV}
\def\confYear{2027}

\title{Adversarial Latent Diffusion Distillation for One-Step Identity Swapping}

\author{Ivan Shchekotov$^{*1}$ \quad
Timofei Gritsaev$^{*1}$ \quad
Viacheslav Pirogov$^{2}$ \quad
Dmitry Vetrov$^{1}$\\[0.4em]
{\normalsize $^1$Constructor University, Bremen, Germany \quad $^2$Sumsub, Berlin, Germany}
}

\begin{document}

\twocolumn[{%
\renewcommand\twocolumn[1][]{#1}%
\maketitle
\input{figures/figure_main_examples}
\medskip
}]
{\renewcommand{\thefootnote}{\fnsymbol{footnote}}%
\footnotetext[1]{\raggedright Equal contribution. Correspondence: \texttt{ishchekotov@constructor.university}, \texttt{slava.pirogov@sumsub.com}. Code: \url{https://github.com/isdevnull/aladdin}.\par}}

\begin{abstract}
Face swapping transfers a source identity onto a target image while preserving the target's pose, expression, hair, and lighting.
The most realistic current methods are diffusion-based, but they need many network evaluations per image. GAN-based methods are fast, yet trained from scratch they must learn realism from the adversarial signal alone and produce visibly weaker results.
We present ALADDIN (\textbf{A}dversarial \textbf{LA}tent \textbf{D}iffusion \textbf{D}istillation for One-Step \textbf{I}dentity Swappi\textbf{N}g), a one-step face swapper obtained by initialising the generator from a pre-trained multi-step flow-matching model and training it adversarially under identity and target-preservation constraints.
We expose a controllable trade-off between identity transfer and target-attribute preservation by training a student model with triplet pseudo-label supervision.
We additionally identify a bias in the standard evaluation protocol, which favours methods trained against the same identity encoder used to score them, and re-evaluate all baselines with an independent one.
ALADDIN is state-of-the-art among one-step methods in realism and target preservation at comparable identity transfer, and matches the best multi-step model with 28× fewer network evaluations and an order of magnitude fewer training samples.
\end{abstract}

\section{Introduction}
\label{sec:intro}

The goal of face swapping is to transfer the source human identity
onto a target face image while keeping the target's attributes intact: background, hair, pose, expression, lighting, and accessories.
An example of a face swap application is shown in Figure~\ref{fig:ffhq-qualitative-full}.
The task has broad relevance across various areas, for example,
film production~\cite{naruniec2020high} and privacy protection~\cite{wilson2022practical}.

Many prominent face-swapping methods are built on the Generative Adversarial Network (GAN)~\citep{goodfellow2014generative} framework.
GAN-based face-swapping methods can be broadly categorised into two families: \emph{direct} and \emph{inversion-based}.
Direct methods, including SimSwap~\citep{chen2020simswap}, HiFiFace~\citep{wang2021hififace}, FaceDancer~\citep{rosberg2023facedancer}, and InfoSwap~\citep{gao2021information}, train a swapping generator end to end with identity, attribute-preservation, and adversarial objectives.
Inversion-based methods, such as MegaFS~\citep{zhu2021one} and E4S~\citep{Liu_2023_CVPR}, manipulate the latent representation of a pretrained StyleGAN generator~\citep{karras2019style, karras2020analyzing}. However, GANs suffer from unstable training and usually produce less realistic results~\citep{dhariwal2021diffusion} than diffusion-based models~\citep{ho2020denoising, song2020score, lipman2022flow}.

Diffusion models train more stably than GANs but need tens of network function evaluations (NFE) per image, where a GAN needs one. Diffusion models for face swapping were developed in~\citep{kim2025diffface, zhao2023diffswap, baliah2025realistic, han2024face, ye2025dreamid, kang2026attribute}, and many distillation techniques~\citep{song2023consistency, liu2023flow, sauer2024adversarial, sauer2024fast, yin2024improved, yin2024one, geng2026mean} aim to combine their quality with a low inference cost.

In our paper, we combine the high realism of diffusion models with one-step GAN inference: we initialise the generator from a pre-trained diffusion model and train it adversarially with a discriminator that reads diffusion features, a design introduced by adversarial diffusion distillation~\citep{sauer2024adversarial, sauer2024fast} and adopted as an auxiliary loss by improved distribution matching distillation~\citep{yin2024one, yin2024improved}.
Our main contributions are:
\begin{enumerate}
    \item We present \MethodName: \textbf{A}dversarial \textbf{LA}tent \textbf{D}iffusion \textbf{D}istillation for one-step \textbf{I}dentity swappi\textbf{N}g, a simple method that achieves state-of-the-art quality among one-step models.
    \item We outperform almost every multi-step baseline and achieve the same quality as the state-of-the-art diffusion model, APPLE~\citep{kang2026attribute}. Our model requires an order of magnitude fewer training samples than APPLE and 28× fewer network evaluations at inference.
    \item We identify a positive bias in identity similarity estimation in the prior evaluation protocols and address it by re-evaluating all baselines with an independent identity model.
\end{enumerate}

\section{Related work}
\label{sec:related}

\paragraph{GAN-based face swapping methods}

Early neural face-swapping methods predominantly relied on GANs and could be broadly divided into direct generator-based and GAN-inversion approaches.
Direct methods train a swapping network to combine the source identity with target attributes.
SimSwap~\citep{chen2020simswap} introduces an ID-injection module and a weak feature-matching loss to improve identity transfer while preserving target attributes.
InfoSwap~\citep{gao2021information} uses information-bottleneck objectives and an identity-contrastive loss to disentangle source identity from target-specific information.
HiFiFace~\citep{wang2021hififace} incorporates 3D shape and semantic priors to improve face-shape transfer and geometric consistency,
while FaceDancer~\citep{rosberg2023facedancer} employs adaptive feature-fusion attention and identity feature refinement to improve robustness to pose variation and occlusion.
InSwapper~\citep{inswapper2023} is a widely used direct model whose training procedure is undocumented.
CSCS~\citep{huang2024identity} uses two surrogate generators to synthesise a source and a target from one real image, which then serves as the ground-truth swap.
Trained from scratch, direct methods must learn realism from the adversarial signal alone while the identity and attribute objectives compete with it, and they produce local artifacts under large pose, shape, or occlusion changes.

The second branch exploits the high-resolution image prior of a pretrained StyleGAN.
MegaFS~\citep{zhu2021one} encodes source and target into an extended StyleGAN2 latent space, blends their high-level codes with a face-transfer module while keeping the target's low-level codes, and decodes the result with the frozen generator.
E4S~\citep{Liu_2023_CVPR} reenacts the source into the target's pose and expression, represents each face by per-region style codes and a segmentation mask, and swaps only the inner-face regions.
These methods inherit the realism of StyleGAN but also its limits: embedding a face into the extended latent space either loses identity detail or leaves the region in which the generator behaves well~\citep{tov2021designing, roich2022pivotal}.

\myparagraph{Diffusion-based face swapping methods} Diffusion models have recently emerged as an alternative to GANs due to their strong image priors, stable training, and high-fidelity generation. DiffFace~\citep{kim2025diffface} introduces an identity-conditioned DDPM and applies identity, semantic, and gaze guidance during sampling. However, the required gradient-based guidance and iterative denoising make inference expensive. DiffSwap~\citep{zhao2023diffswap} formulates face swapping as 3D-aware masked diffusion, using facial landmarks and masks to preserve target pose and expression.

REFace~\citep{baliah2025realistic} trains latent diffusion as a conditional inpainting model and unrolls several differentiable DDIM steps during training to apply identity and perceptual losses to estimated clean images. FaceAdapter~\citep{han2024face} adapts a pretrained Stable Diffusion model by explicitly separating identity, spatial, and attribute conditions. It maps source face embeddings into cross-attention tokens, while target landmarks, background, and CLIP features provide structural and appearance guidance.

More recent work, DreamID~\citep{ye2025dreamid}, revisits the idea of CSCS~\citep{huang2024identity} by introducing the triplet ID group, which allows existing data to be used as ground truth for face swaps (neither its code nor its weights have been released). APPLE~\citep{kang2026attribute} further studies triplet supervision, noting that the teacher used in DreamID leads to limited target attribute preservation. It trains a diffusion teacher with conditional deblurring and attribute-aware inversion to generate pseudo-labels with better target preservation, and subsequently trains a student on the resulting pseudo-triplets. While DreamID achieves one-step generation by starting from an already-distilled model (SD-Turbo~\citep{sauer2024adversarial}), our method can directly adapt a non-distilled multi-step model, allowing us to use the larger FLUX model~\citep{flux2024} and PuLID~\citep{guo2024pulid} for stronger identity conditioning. Compared to APPLE, which is multi-step and requires an order of magnitude more training samples, \MethodName{} generates swaps in a single step.

\section{Method}
\label{sec:method}

We formulate one-step face swapping as conditional generation. Given a source image $x_{\mathrm{src}}$ and a target image $x_{\mathrm{tgt}}$, the generator $G_\theta$ produces a face swap in a single step
\begin{equation}
    \hat{x}
    = G_\theta(x_{\mathrm{src}}, x_{\mathrm{tgt}})
    = x_{\mathrm{tgt}}
    - \sigma_{\mathrm{gen}}\, v_\theta\!\left(x_{\mathrm{tgt}}, \sigma_{\mathrm{gen}}, x_{\mathrm{src}}\right),
    \label{eq:generator}
\end{equation}
where $\hat{x}$ is the generated face image, which should retain the attributes of $x_{\mathrm{tgt}}$ and carry the identity of $x_{\mathrm{src}}$. Here $v_\theta$ is the velocity predicted by a flow-matching model~\citep{liu2023flow, flux2024} extended with an identity-conditioning branch for $x_\mathrm{src}$.
We treat the target image as a sample at noise level $\sigma_{\mathrm{gen}}$, so that the model maps it to the clean swap in one evaluation. The generator operates on the latents of the frozen FLUX VAE, and Equation~\ref{eq:generator} is applied in that space before decoding. The noise level $\sigma_{\mathrm{gen}}$ is a fixed hyperparameter which we set to $0.2$ (ablated in \suppsec{app:sigma}). At inference, we additionally feed the clean target latent through an OminiControl branch~\citep{tan2025ominicontrol}, which serves as an inference-time control to pull the output closer to the target (Section~\ref{sec:comparison}).

\input{figures/figure_training_full}

\subsection{Training Objectives}

Figure~\ref{fig:training} gives an overview of the training objectives and data groups described in this section.

\myparagraph{Identity loss} Identity transfer is encouraged by minimising the cosine distance between the identity embeddings of the generated and the source image:
\begin{equation}
    \mathcal{L}_{\mathrm{id}}(\hat{x}, x_{\mathrm{src}})
    =
    1 -
    \cos\left(f_{\mathrm{id}}(\hat{x}), f_{\mathrm{id}}(x_{\mathrm{src}})\right),
    \label{eq:id_loss}
\end{equation}
where $f_{\mathrm{id}}$ returns the identity embedding of a frozen face-recognition network~\citep{deng2019arcface}.
This objective is standard in GAN- and diffusion-based face swapping~\citep{chen2020simswap, rosberg2023facedancer, zhu2021one, huang2024identity, kim2025diffface, zhao2023diffswap, baliah2025realistic, ye2025dreamid, kang2026attribute}.

\myparagraph{Realism losses} The identity loss alone yields unrealistic images, so we add adversarial training and, where the data provides a ground-truth answer, a reconstruction loss.

\mysubparagraph{Adversarial loss} Adversarial training supplies the realism signal, but its behaviour is significantly affected by the representations the discriminator sees. Following Latent Adversarial Diffusion Distillation (LADD)~\citep{sauer2024fast}, the discriminator relies on the activations of the frozen diffusion model rather than pixels or raw latents.

Let $\mathcal{E}$ denote the frozen VAE encoder, $z = \mathcal{E}(\hat{x})$ the latent of a generated swap, and $z^{+} = \mathcal{E}(x^{+})$ the latent of a real image $x^{+}$, for which we use the target image of the same batch. Both are corrupted by the same linear interpolation
$z_\sigma = (1 - \sigma)\, z + \sigma \epsilon,
    \ \epsilon \sim \mathcal{N}(0, I)$
at a noise level $\sigma$ shared by the real and the generated branch. We then pass both latents through the frozen diffusion model, the same model that initialises $G_\theta$, and intercept the activations $h^{(1)}, \dots, h^{(K)}$ of $K$ of its blocks. Each activation is read by its own discriminator head, and the heads are aggregated into a single realism score:
\begin{equation}
    D_\varphi(z_\sigma) = \sum_{k=1}^{K} w_k \, D_{\varphi_k}\!\left(h^{(k)}, \sigma\right),
    \quad w = \operatorname{softmax}(\alpha),
    \label{eq:ladd_head}
\end{equation}
so that training itself decides which activations carry usable signal. Only the heads $\{D_{\varphi_k}\}$ and the pooling logits $\alpha$ are trainable. The diffusion model stays frozen. Since it is the same network that initialises the generator, adversarial training introduces no pretrained model beyond the diffusion prior we already depend on.

We use the non-saturating GAN objective~\citep{goodfellow2014generative}, minimising $\mathcal{L}_{D_\varphi}$ over the discriminator parameters $\varphi$ and $\mathcal{L}_{\mathrm{GAN}}$ over the generator parameters $\theta$, where $D_\varphi$ predicts the probability that a sample is real:
\begin{align}
    \mathcal{L}_{D_\varphi}(z^{+}_\sigma, z_\sigma) &=
    - \log D_\varphi(z^{+}_\sigma)
    - \log\left(1 - D_\varphi(z_\sigma)\right),
    \label{eq:gan_d} \\
    \mathcal{L}_{\mathrm{GAN}}(z_\sigma) &= -\log D_\varphi(z_\sigma).
    \label{eq:gan_g}
\end{align}

These are not generic features: the diffusion model was trained to generate images including faces, so its activations at every depth already encode what a plausible face looks like, and a discriminator built on them reuses that prior instead of learning it from the adversarial signal alone. Classifiers derived from diffusion models also show a shape bias close to human perception, whereas discriminative backbones rely on texture~\citep{geirhos2019, jaini2024intriguing}, which motivates the generative features of LADD~\citep{sauer2024fast} over the frozen DINO backbone~\citep{caron2021emerging, oquab2024dinov} of ADD~\citep{sauer2024adversarial}.

\mysubparagraph{Reconstruction loss} As we show below, training data can be arranged so that an existing image serves as the ground-truth swap. For example, if the pair $(x_{\mathrm{src}}, x_{\mathrm{tgt}})$ shares the same identity, then $x_{\mathrm{tgt}}$ is itself the correct answer. In this case, we can add a reconstruction loss:
\begin{equation}
    \mathcal{L}_{\mathrm{rec}}(\hat{x}, x_{\mathrm{tgt}}) = d(\hat{x}, x_{\mathrm{tgt}}),
\end{equation}
where $d$ is a distance metric, for example the $\ell_1$ or $\ell_2$ distance. Empirically, we find that Learned Perceptual Image Patch Similarity (LPIPS)~\citep{zhang2018perceptual} leads to better quantitative results than $\ell_1$.

The triplet group of Section~\ref{sec:training_data}, following~\citep{ye2025dreamid, kang2026attribute}, extends it to the case of differing identities.

\myparagraph{Target preservation losses}
Finally, we need objectives that preserve the target attributes. Three of them must be carried over from the target: head pose, facial expression, and the image background.

\mysubparagraph{Pose loss} We preserve the target head orientation using a
frozen 6DRepNet pose estimator~\citep{10477888}. Given an image $x$, the estimator predicts a rotation matrix $\psi_{\mathrm{pose}}(x)=R_x\in \mathrm{SO}(3)$. We compare the predicted orientations of the generated and target images using their geodesic distance:
\begin{equation}
    \begin{aligned}
    \mathcal{L}_{\mathrm{pose}}(\hat{x}, x_{\mathrm{tgt}}) &= \arccos\bigl(\operatorname{clip}(c, -1+\epsilon, 1-\epsilon)\bigr), \\
    c &= \tfrac{1}{2}\bigl(\operatorname{tr}(R_{\hat{x}}R_{x_\mathrm{tgt}}^{\top})-1\bigr),
    \end{aligned}
    \label{eq:pose_loss}
\end{equation}
where $R_{\hat{x}}=\psi_{\mathrm{pose}}(\hat{x})$ and
$R_{x_{\mathrm{tgt}}}=\psi_{\mathrm{pose}}(x_{\mathrm{tgt}})$.

\mysubparagraph{Expression loss} We preserve the target expression using a
frozen DECA estimator~\citep{DECA:Siggraph2021}. DECA predicts a 50-dimensional FLAME~\citep{FLAME:SiggraphAsia2017} expression vector $\psi_{\mathrm{expr}}(x)$ of blendshape weights that describe expression-related geometry, such as smiles and mouth opening, separately from identity shape, head pose, and appearance. We minimise:
\begin{equation}
    \mathcal{L}_{\mathrm{expr}}(\hat{x}, x_{\mathrm{tgt}})
    = \frac{1}{2d_{\mathrm{expr}}}\left\|\psi_{\mathrm{expr}}(\hat{x})-\psi_{\mathrm{expr}}(x_{\mathrm{tgt}})\right\|_2^2.
\end{equation}
where $d_{\mathrm{expr}} = 50$ is the dimension of the expression vector.

\mysubparagraph{Background loss} Identity transfer requires modifying the face, whereas regions outside the face should remain consistent with the target. We obtain a facial mask $m_{\mathrm{tgt}}$ from a frozen face-parsing model, SegFormer~\citep{xie2021segformer}, dilated and smoothed at its boundary and then binarised. The mask covers the facial skin, nose, eyebrows, eyes, mouth, and ears, so its complement $1-m_{\mathrm{tgt}}$ retains the target hair, glasses, hat, neck, clothing, and background.
We minimise the normalised pixel difference over this complementary region:
\begin{equation}
    \mathcal{L}_{\mathrm{bg}}(\hat{x}, x_{\mathrm{tgt}})
    =
    \frac{
        \sum_{c,h,w}\left(1-m_{\mathrm{tgt}}^{h,w}\right)
        \left|\hat{x}^{c,h,w}-x_{\mathrm{tgt}}^{c,h,w}\right|
    }{
        \max\!\left(1,\sum_{h,w}\left(1-m_{\mathrm{tgt}}^{h,w}\right)\right)
    },
    \label{eq:background_loss}
\end{equation}
where the superscripts index the channel and the two spatial dimensions.

\subsection{Training Data Groups}
\label{sec:training_data}

\myparagraph{Cross Identity Group} Having established all loss objectives, we start with the general case $\mathrm{ID}(x_{\mathrm{src}}) \neq \mathrm{ID}(x_{\mathrm{tgt}})$, for which we construct the cross identity objective as:
{
\small
\begin{align*}
    &\mathcal{L}_{\mathrm{cross}}(\hat{x}, x_{\mathrm{src}}, x_{\mathrm{tgt}})
        = \lambda_{\mathrm{id}} \mathcal{L}_{\mathrm{id}}(\hat{x}, x_{\mathrm{src}}) + \lambda_{\mathrm{GAN}} \mathcal{L}_{\mathrm{GAN}}(\hat{x}) +
        \\ &\quad
        \lambda_{\mathrm{pose}} \mathcal{L}_{\mathrm{pose}}(\hat{x}, x_{\mathrm{tgt}}) +
        \lambda_{\mathrm{expr}} \mathcal{L}_{\mathrm{expr}}(\hat{x}, x_{\mathrm{tgt}})
        +
        \lambda_{\mathrm{bg}} \mathcal{L}_{\mathrm{bg}}(\hat{x}, x_{\mathrm{tgt}}).
\end{align*}
}
The identity loss drives identity transfer, the adversarial loss makes the one-step generation realistic, and the last three losses preserve the target attributes. Figure~\ref{fig:training} (left) presents training with the cross identity group.

\myparagraph{Same Identity Group} When the pair $(x_{\mathrm{src}}, x_{\mathrm{tgt}})$ shares the same identity, we construct the same identity objective as:
{
\small
\begin{align*}
    \mathcal{L}_{\mathrm{same}}(\hat{x}, x_{\mathrm{src}}, x_{\mathrm{tgt}})
        &= \lambda_{\mathrm{id}} \mathcal{L}_{\mathrm{id}}(\hat{x}, x_{\mathrm{src}}) + \lambda_{\mathrm{GAN}} \mathcal{L}_{\mathrm{GAN}}(\hat{x}) \\
        &\quad + \lambda_{\mathrm{rec}} \mathcal{L}_{\mathrm{rec}}(\hat{x}, x_{\mathrm{tgt}}).
\end{align*}
}
The reconstruction loss supplies a direct supervision signal and thereby improves image quality. A model trained on the same identity group alone, however, cannot perform well in the general case, since the generator attains a perfect score by naively copying the target image.

\myparagraph{Triplet Group} To apply the reconstruction loss when the identities differ, we use the triplet group of~\citep{ye2025dreamid, kang2026attribute}, illustrated in Figure~\ref{fig:training} (right). It requires two distinct images $(x_{\mathrm{src}}^{A}, x_{\mathrm{tgt}}^{A})$ of the same identity $A$. We first generate a pseudo-target using a frozen face swap teacher $\textcolor{beautifulblue}{G_{\theta_T}}$
\begin{equation*}
    x_{\mathrm{tgt}}^{A\to B} = \textcolor{beautifulblue}{G_{\theta_T}}\!\left(x_{\mathrm{src}}^{B}, x_{\mathrm{tgt}}^{A}\right)
\end{equation*}
carrying the source identity $B$ and the target attributes of $x_{\mathrm{tgt}}^{A}$. We then generate a final image from the real image $x_{\mathrm{src}}^{A}$ and the pseudo-target $x_{\mathrm{tgt}}^{A\to B}$:
\begin{equation*}
    \hat{x} = G_\theta\!\left(x_{\mathrm{src}}^{A}, x_{\mathrm{tgt}}^{A\to B}\right),
\end{equation*}
where $\hat{x}$ has the identity of $A$ and the target attributes of $x_{\mathrm{tgt}}^{A\to B}$, and hence those of $x_{\mathrm{tgt}}^{A}$. The construction therefore yields a ground-truth reconstruction target, $x_{\mathrm{tgt}}^{A}$. The triplet objective is:
{
\begin{align*}
    &\mathcal{L}_{\mathrm{triplet}}(\hat{x}, x_{\mathrm{src}}^{A}, x_{\mathrm{tgt}}^{A})
        = \lambda_{\mathrm{id}} \mathcal{L}_{\mathrm{id}}(\hat{x}, x_{\mathrm{src}}^{A}) +
        \\ &\quad
        \lambda_{\mathrm{GAN}} \mathcal{L}_{\mathrm{GAN}}(\hat{x}) +
        \lambda_{\mathrm{rec}} \Bigl(\mathcal{L}_{\mathrm{rec}}(\hat{x}, x_{\mathrm{tgt}}^{A}) +
        \\ &\quad
        \lambda_{\mathrm{pose}} \mathcal{L}_{\mathrm{pose}}(\hat{x}, x_{\mathrm{tgt}}^{A}) +
        \lambda_{\mathrm{expr}} \mathcal{L}_{\mathrm{expr}}(\hat{x}, x_{\mathrm{tgt}}^{A})\Bigr).
\end{align*}}

\myparagraph{Training pipeline} We first train the \emph{teacher} on a mixture of the cross identity and same identity groups. The \emph{student} generator is initialised from the teacher, with a freshly initialised discriminator, and trained on the triplet group alone, with the frozen teacher $G_{\theta_T}$ generating the pseudo-targets $x_{\mathrm{tgt}}^{A\to B}$ on the fly. Increasing $\lambda_{\mathrm{rec}}$ of the student trades identity transfer against realism and target preservation, so we report students at several settings, written $\text{Student}_{\lambda}$.

\section{Experiments}
\label{sec:experiments}

\subsection{Experimental setup}

\myparagraph{Implementation details}
We follow the architecture of APPLE~\citep{kang2026attribute} and initialise the generator with FLUX.1-Krea-dev~\citep{flux2024}, using PuLID~\citep{guo2024pulid} as the identity encoder.
The same frozen FLUX.1-Krea-dev transformer supplies the discriminator features of Equation~\ref{eq:ladd_head}.
We train LoRA adapters of rank $64$ on the attention projections of all transformer blocks, together with PuLID's cross-attention layers, on top of the frozen backbone, and all experiments run at a resolution of $512 \times 512$. The OminiControl branch is never trained. At inference, and when the frozen Teacher generates pseudo-targets, it feeds the target latent a second time through the same weights, LoRA included, at timestep $0$.
For $\mathcal{L}_{\mathrm{id}}$, we use an ArcFace~\citep{deng2019arcface} model trained on Glint360K~\citep{an2021partial}. Following REFace~\citep{baliah2025realistic} and APPLE~\citep{kang2026attribute}, the source image is masked before it enters the identity encoder, so that target attributes do not leak into the embedding. Table~\ref{tab:ffhq-bench1k-WebFaceR50} reports the teacher and the students with $\lambda_{\mathrm{rec}} \in \{1, 3\}$. The figures and the appendix add $\lambda_{\mathrm{rec}} \in \{2, 7\}$. Further details are in \suppsec{app:implementation_details}.

\myparagraph{Dataset filtering} Our face swapping model is trained on VGGFace2-HQ~\citep{chen2023simswap++} and Arc2Face~\citep{paraperas2024arc2face}. Both datasets provide several images per identity, which the same identity and triplet groups require. Batches are drawn from Arc2Face and VGGFace2-HQ in a 4:1 ratio, and after the filtering described below the training set contains 4.1M images of 540K identities. Both datasets contain low-quality images and identities that mix photographs of different people.
We filter them with an aesthetic-score threshold~\citep{schuhmann2022aestheticpredictor} as in~\citep{kang2026attribute} and an identity-consistency check described in \suppsec{app:implementation_details}.

\myparagraph{Evaluation protocol} We follow the protocol of~\citep{baliah2025realistic, ye2025dreamid, kang2026attribute} and evaluate on the fixed list of $1{,}000$ source--target pairs from FFHQ~\citep{karras2019style} used by REFace, reporting four aspects of a swap. Realism is measured by the Fr\'echet distance~\citep{heusel2017fid} between CLIP ViT-B/32 features~\citep{kynkaanniemi2023role} of the $1{,}000$ generated images and of all $70{,}000$ FFHQ images at $512 \times 512$, computed with the evaluation code released with REFace~\citep{baliah2025realistic}. We refer to it as FID throughout. Identity transfer is measured in two ways: the cosine similarity between the identity embeddings of the generated and the source face, and retrieval accuracy, obtained by ranking all source faces by that similarity and reporting how often the correct one appears first (Top-1) or within the first five (Top-5). Following~\citep{baliah2025realistic}, we mask out the region outside the face in both images before extracting these embeddings, so that shared background and hair cannot affect the score. Target preservation is measured by the $\ell_2$ distance between the generated and target images in head pose, estimated with Hopenet~\citep{Ruiz_2018_CVPR_Workshops}, and in expression, estimated with Deep3DFace~\citep{deng2019accurate}. Both are applied to the whole frame without face detection, and both are independent of the estimators used for training. To provide a fair comparison, we generate face swaps with the available baseline models and score them under our protocol. \suppsec{app:celebahq} reports the same comparison on CelebA-HQ and FaceForensics++.

\myparagraph{Identity encoder bias in prior evaluation protocols}
Prior works~\citep{baliah2025realistic,
ye2025dreamid, kang2026attribute} evaluate identity with
\textit{MS1M--IR-SE50}, the encoder that REFace~\citep{baliah2025realistic} and
InfoSwap~\citep{gao2021information} are optimised against.
Table~\ref{tab:identity_bias} shows the gap: REFace and InfoSwap lose about $0.2$ under either encoder they were not trained against (\suppsec{app:identity_bias}).
Scoring all methods with \textit{MS1M--IR-SE50} therefore rewards the ones that were optimised against it.
\input{figures/table_identity_bias}

We instead evaluate with \textit{WebFace--R50}, an ArcFace encoder with an iResNet-50 backbone trained on WebFace600K, which performs strongly overall and across demographic subsets~\citep{insightface_model_zoo}. To our knowledge no compared method is trained against it.
InSwapper uses it for conditioning, so Table~\ref{tab:ffhq-bench1k-WebFaceR50} also reports its identity under \textit{MS1M--IR-SE50}.

\textit{WebFace--R50} serves for evaluation only. Identity conditioning and $\mathcal{L}_{\mathrm{id}}$ use \textit{Glint--R100}, an iResNet-100 ArcFace trained on Glint360K~\citep{deng2019arcface, an2021partial}, so our method is never optimised against the encoder that scores it.

\subsection{Comparison with other methods}

\label{sec:comparison}
\input{figures/figure_ffhq_selected_pairs.tex}

\myparagraph{Quantitative comparison} Table~\ref{tab:ffhq-bench1k-WebFaceR50} reports the results. No method dominates on all four metrics, so we report our models across the trade-off set by $\lambda_{\text{rec}}$.

\input{figures/table_metrics_WebFaceR50_wretrieval}

\mysubparagraph{Comparison with one-step methods} Every one of our models improves on MegaFS and E4S in all four metrics. HiFiFace is beaten in all four by our Teacher and \Trip{1}, InfoSwap by \Trip{1}, and SimSwap and FaceDancer by \Trip{3}. Only InSwapper and CSCS reach a higher identity similarity, and both pay for it elsewhere: InSwapper is scored with the encoder it uses for conditioning~\citep{insightface_model_zoo, inswapper2023}, and under \textit{MS1M--IR-SE50} it leads \Trip{1} by $0.03$, while the student leads by a wide margin in FID, pose, and expression. CSCS pastes the source face onto the target, as the qualitative comparison shows, at a clear cost in realism, pose, and expression. On Top-1 retrieval, only CSCS exceeds our Teacher with the OminiControl branch off.

\mysubparagraph{Comparison with multi-step methods} DiffSwap and REFace are beaten in all four metrics by every one of our models, and FaceAdapter by our Teacher and \Trip{1}. APPLE is the only baseline that remains competitive. The APPLE row is its released student model, sampled with 28 steps. \Trip{3} matches it in identity transfer at $0.40$ and in FID ($3.13$ against $3.22$), and improves on it in pose ($1.45$ against $1.86$) and expression ($0.56$ against $0.63$). Our model reaches this with a single network evaluation instead of APPLE's $28$, with the same architecture.

\mysubparagraph{Training cost against APPLE} APPLE trains its teacher on 1.04M training samples and its student on a further 0.24M, reporting 12.78 days on four A6000 GPUs, of which 5.78 generate pseudo-triplets offline with the 28-step teacher. ALADDIN uses 0.10M samples for the teacher, about 4.5 hours on four H100 GPUs, and 0.015M for the student, with no offline stage because the teacher is itself one-step.
At inference, one swap takes $0.29$\,s on one H100 GPU against $5.80$\,s for the APPLE student (\suppsec{app:timing}).

\input{figures/figure_operating_points}

\myparagraph{Choosing the operating point at inference} A trained \MethodName{} model exposes two controls at inference that need no retraining: the PuLID identity weight, which scales the injected identity features and moves the output along the identity--attribute trade-off, and the OminiControl branch, which pulls the output toward the target. Figure~\ref{fig:operating-points} shows the trained models under both settings. The two settings join into one frontier: with the OminiControl branch on, the models cover identity similarities from $0.36$ to $0.48$ at the best realism and attribute scores in Table~\ref{tab:ffhq-bench1k-WebFaceR50}. With it off, the same models continue the curve to $0.56$ with a realism that no one-step baseline reaches at that identity, as the Teacher row with the branch off in Table~\ref{tab:ffhq-bench1k-WebFaceR50} shows at $0.55$.

\input{figures/figure_pulid_sweep}

\myparagraph{One model covers the frontier} The students in Table~\ref{tab:ffhq-bench1k-WebFaceR50} are separately trained at different reconstruction weights. Figure~\ref{fig:pulid-sweep} shows an alternative: sweeping the PuLID weight of \Trip{3} alone from $0.75$ to $1.25$, with the OminiControl branch on and then off, traces the frontier of all separately trained students up to an identity similarity of $0.52$, and runs slightly below it while the branch is on. Only the highest-identity end, above $0.52$, still requires a model trained with a smaller reconstruction weight, since the sweep bends upward there. Student training still matters: the swept \Trip{3} lies below the swept Teacher at every identity (\suppsec{app:controls}). One trained model therefore covers the operating points of the whole family, chosen at inference through the PuLID weight.

\myparagraph{Qualitative comparison}
Figure~\ref{fig:qualitative-selected-compact} compares our method against the selected strong baselines across six axes, and \suppsec{app:discussion} extends it with further pairs, the omitted baselines, and the full comparison grids. The first two probe realism (an unusual target hairstyle and a large age difference between source and target), the remaining four the preservation of glasses, tattoo, head pose, and expression. GAN-based methods introduce severe artifacts. CSCS shows \textit{copy-paste} behaviour with little adaptation to the target. InSwapper transfers identity well at the cost of the 128$\times$128 fidelity of the original model. InfoSwap alters the hair (row 1), renders the glasses transparent (row 3), and perturbs pose and expression (rows 5 and 6). Diffusion-based baselines are more realistic and preserve attributes better than the GAN-based methods. REFace produces unrealistic texture and under-transfers the source identity in most rows. APPLE is the strongest baseline overall. After triplet training, ALADDIN matches APPLE's realism and attribute preservation at a fraction of its inference cost. Higher $\lambda_{\text{rec}}$ improves realism and attribute retention at the cost of identity transfer.

\subsection{Ablation study}
\label{sec:ablation}

We ablate the core design choices of teacher training (Table~\ref{tab:ablation}) as well as the individual loss terms of both the teacher (Table~\ref{tab:ablation-teacher-losses}) and the student (Table~\ref{tab:ablation-student-losses}). The reconstruction and PuLID weights are studied in Section~\ref{sec:comparison}, the OminiControl branch and student sweeps in \suppsec{app:controls}, the generator noise level $\sigma_{\mathrm{gen}}$ in \suppsec{app:sigma}, and dataset filtering in \suppsec{app:filtering}.

\input{figures/table_ablation}

\myparagraph{Data mixture} We vary the proportion of cross identity and same identity batches in Table~\ref{tab:ablation}. Same identity pairs provide a ground-truth target image, and including $25\%$ same identity batches improves FID ($4.07$ vs.\ $4.57$), pose ($1.71$ vs.\ $1.98$), and expression ($0.68$ vs.\ $0.83$) over training on cross identity pairs alone. Increasing the same identity share to $50\%$ changes every metric by less than $0.1$, whereas an overly large share encourages the generator to copy the target image rather than transfer the source identity. We therefore fix the mixture at $75\%$ cross identity and $25\%$ same identity.

\myparagraph{Latent discriminator} Following LADD~\citep{sauer2024fast}, our discriminator reads intermediate activations of the frozen diffusion transformer rather than raw VAE latents. To verify the importance of these features in our setting, we train on cross-identity batches only and replace the diffusion features $h^{(1)}, \dots, h^{(K)}$ in Equation~\ref{eq:ladd_head} with the noised latent $z_\sigma$, leaving all other settings unchanged. Compared against the $1.00\,/\,0.00$ cross-identity baseline, FID degrades sharply from $4.57$ to $17.01$, pose from $1.98$ to $2.55$, and expression from $0.83$ to $1.01$, while identity similarity rises from $0.48$ to $0.52$. Without the pretrained diffusion features, the discriminator fails to provide a useful realism signal (scoring worse than training with no discriminator at all) and leaves the generator to satisfy $\mathcal{L}_{\mathrm{id}}$ through severe visual artifacts.

\myparagraph{Realism signal} Table~\ref{tab:ablation} (bottom) also compares our feature-space adversarial loss against training with no realism objective, replacing $\mathcal{L}_{\mathrm{GAN}}$ with Distribution Matching Distillation (DMD)~\citep{yin2024one} as a regulariser toward the frozen FLUX prior~\citep{rakitin2024regularized}, and combining both objectives as in DMD2~\citep{yin2024improved} (see \suppsec{app:implementation_details} for setup details). At matched identity similarity, removing the adversarial loss degrades FID from $4.07$ to $5.58$ and yields less natural facial textures (Figure~\ref{fig:ablation-qual-realism}). DMD alone improves over having no realism loss ($4.85$ vs.\ $5.58$ FID), but remains worse than $\mathcal{L}_{\mathrm{GAN}}$ alone ($4.07$ FID) while increasing training time per step by $2.7\times$ due to the additional score-network evaluations. Combining DMD with the adversarial loss reaches a similar FID ($4.16$) at $3.6\times$ the training cost, so we use the adversarial loss alone.

\input{figures/table_ablation_teacher_losses}

\input{figures/table_ablation_student_losses}

\myparagraph{Teacher loss terms} Table~\ref{tab:ablation-teacher-losses} removes one preservation objective at a time from the teacher recipe. Alongside the standard benchmark metrics, we report the held-out training expression loss (DECA) and the non-face background $\ell_1$ error (Bg.), described in \suppsec{app:implementation_details}. Removing $\mathcal{L}_{\mathrm{pose}}$ or $\mathcal{L}_{\mathrm{bg}}$ alone causes only a small increase in its corresponding metric (pose error rises from $1.61$ to $1.90$ and background error from $1.08$ to $1.41$) with little visible difference in the generated swaps (Figure~\ref{fig:ablation-qual-preservation}), as the remaining objectives and the $25\%$ same-identity reconstruction batches still anchor the output to the target. Removing $\mathcal{L}_{\mathrm{expr}}$ leaves the evaluation protocol's full-frame Deep3DFace metric almost unchanged ($0.68$ vs.\ $0.67$), but increases the held-out DECA error from $8.4$ to $14.3$, corresponding to visible brow, gaze, and eyelid differences on expressive targets (Figure~\ref{fig:ablation-qual-eyes}). Removing all three preservation losses at once degrades pose, expression, and background simultaneously. If we also drop the same-identity reconstruction batches and train with $\mathcal{L}_{\mathrm{id}}$ and $\mathcal{L}_{\mathrm{GAN}}$ alone, nothing anchors the generator to the target attributes and all preservation metrics degrade sharply (Figure~\ref{fig:ablation-qual-preservation}, last column).

\myparagraph{Student loss terms} At the student stage (Table~\ref{tab:ablation-student-losses}), every triplet provides a real ground-truth target $x_{\mathrm{tgt}}^A$, so the triplet reconstruction loss $\mathcal{L}_{\mathrm{rec}}$ and the identity loss $\mathcal{L}_{\mathrm{id}}$ drive the fine-tuning: removing $\mathcal{L}_{\mathrm{id}}$ causes identity transfer to fail ($0.21$), while without $\mathcal{L}_{\mathrm{rec}}$ the student has no reconstruction supervision at all (since student training uses no same-identity batches) and its FID, pose, and expression degrade below the teacher baseline. Because the student is initialised from the adversarially trained teacher and $\mathcal{L}_{\mathrm{rec}}$ directly constrains realism and target attributes against $x_{\mathrm{tgt}}^A$ over this short $15$K-sample stage, removing $\mathcal{L}_{\mathrm{pose}}$ and $\mathcal{L}_{\mathrm{expr}}$ only slightly increases held-out DECA error (from $7.1$ to $8.2$), and removing $\mathcal{L}_{\mathrm{GAN}}$ barely changes FID ($3.35$ vs.\ $3.31$). In practice, student fine-tuning can therefore be simplified to $\mathcal{L}_{\mathrm{rec}}$ and $\mathcal{L}_{\mathrm{id}}$ alone with minimal loss in quality.

\section{Conclusion and limitations}
\label{sec:conclusion}

We presented \MethodName{}, a one-step face swapper obtained by initialising the generator from a pretrained flow-matching model and training it adversarially, with a frozen copy of the same model serving as the discriminator's feature extractor.
\MethodName{} is state-of-the-art among one-step methods and, at equal identity similarity and FID, improves on APPLE, the strongest multi-step baseline, in pose and expression, with one network evaluation instead of $28$ and an order of magnitude fewer training samples. A trained model covers a whole range of operating points through two inference-time controls. Finally, we identified an identity-encoder bias in prior evaluation protocols and re-evaluated all compared methods with an independent encoder.

\myparagraph{Limitations and future work}
Our method inherits the field's failure modes: identity transfer degrades on unusual source identities and on large age differences, and our models with the weakest target preservation lose fine target detail such as a tattoo's texture (Figure~\ref{fig:qualitative-selected-compact}). All numbers are single runs on the 1,000 FFHQ pairs, with checkpoints selected on the same benchmark. The preservation losses rest on pretrained experts for identity, pose, expression, and parsing, whose biases and failure cases the generator inherits. The next step is extending the method to video swapping, where the per-frame cost is attractive but nothing in our training enforces temporal consistency.

\section*{Societal impact}
Face swapping can be misused to produce non-consensual or deceptive imagery, and a one-step method lowers the cost of doing so at scale. Our interest in the problem comes from identity verification, where realistic swaps are needed to train and stress-test the systems that detect them. Publicly available tools already produce convincing swaps, so defenders need training examples at least as strong as those available to attackers. We therefore release \MethodName{} in stages. First, we openly release a classifier that identifies images generated by \MethodName{} (\url{https://huggingface.co/Sumsub/aladdin-detector}). It is specific to our model and is not a general deepfake detector, and detection alone should be combined with other safeguards. Code is publicly available, while model weights will be available only through gated access for research use, under terms that prohibit impersonation, fraud and non-consensual content. We train only on research datasets used throughout prior face-swapping work, and all qualitative examples use FFHQ images.

\section*{Acknowledgements}
This work was supported by Sumsub through a donation to Constructor University under the Sumsub AI Academic Program. The authors gratefully acknowledge the computing time made available to them on the high-performance computer at the NHR Center of TU Dresden. This center is jointly supported by the Federal Ministry of Research, Technology and Space of Germany and the state governments participating in the NHR (www.nhr-verein.de/unsere-partner).

{
    \small
    \bibliographystyle{ieeenat_fullname}
    \bibliography{main}
}

\input{appendix}

\end{document}

%% file: macros.tex
\usepackage{xcolor}
\definecolor{beautifulblue}{HTML}{007ACC}
\usepackage{graphicx}
\usepackage{booktabs}
\usepackage{algorithm}
\usepackage{algpseudocode}
\usepackage{makecell}

\renewcommand{\dbltopfraction}{0.95}
\renewcommand{\dblfloatpagefraction}{0.6}
\newcommand{\MethodName}{ALADDIN}
\newcommand{\Trip}[1]{Student$_{\lambda=#1}$}

\newcommand{\myparagraph}[1]{\vspace{.2cm} \noindent \textbf{#1} \quad}
\newcommand{\mysubparagraph}[1]{\vspace{.1cm} \noindent \textit{#1} \quad}
\newcommand{\suppsec}[1]{Appendix~\ref{#1}}

%% file: figures/figure_main_examples.tex
\begin{center}
  \includegraphics[width=0.74\textwidth]{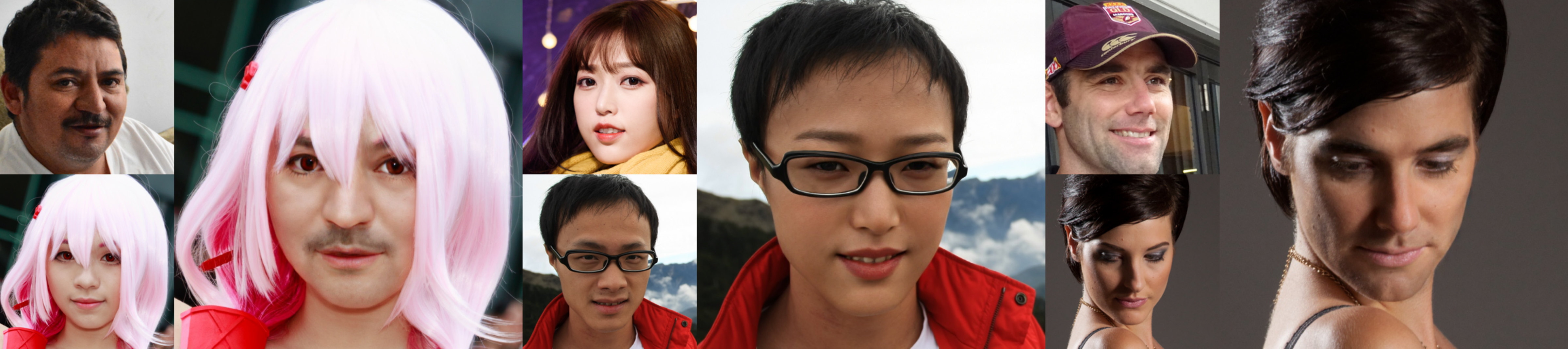}
  \captionof{figure}{\textbf{Qualitative face-swapping results on FFHQ}. Each example shows the source identity (top left), target image (bottom left), and the output of our \Trip{1} (Table~\ref{tab:ffhq-bench1k-WebFaceR50}). The model transfers the source identity while preserving the target's pose, expression, hairstyle, glasses, illumination, and background.}
  \label{fig:ffhq-qualitative-full}
\end{center}

%% file: figures/figure_training_full.tex
\begin{figure*}[t]
  \centering
  \includegraphics[width=\textwidth]{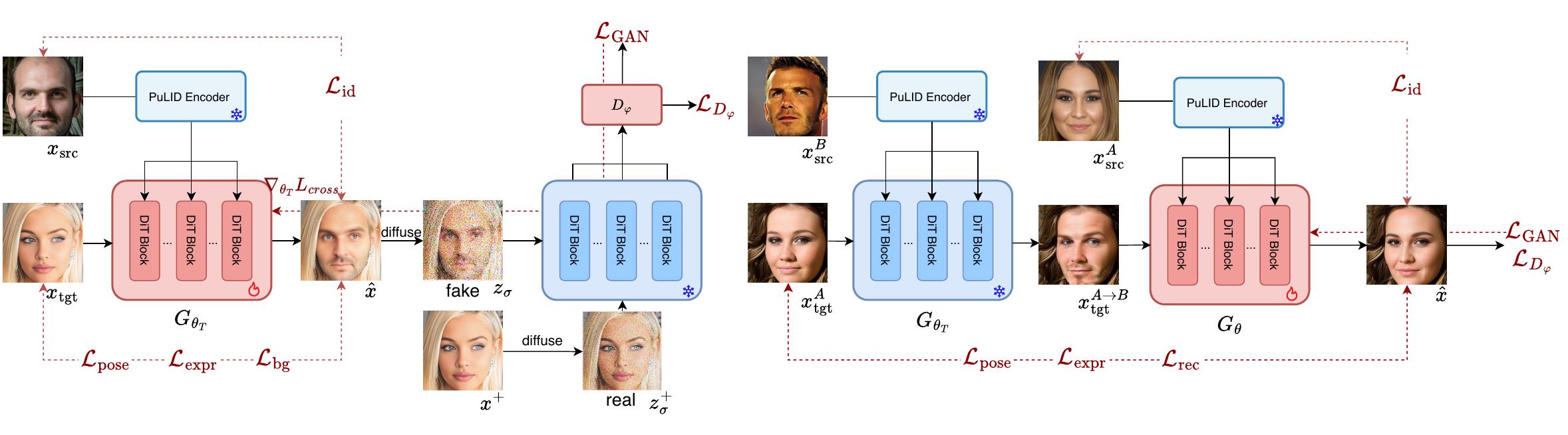}
  \caption{\textbf{Overview of \MethodName\ training.}
  \textbf{\textit{Left (teacher training)}}: the teacher generator $G_{\theta_T}$ receives the target image
  $x_{\mathrm{tgt}}$ and the source identity encoded by frozen PuLID, and produces the one-step swap $\hat{x}$. Identity, pose, expression, and background losses constrain the swap.
  For adversarial training, the latent of $\hat{x}$ and that of a real image $x^+$ are diffused to the same noise
  level $\sigma$ and passed through a frozen diffusion model. The discriminator, $D_\varphi$, classifies its intermediate features as generated or real.
  \textbf{\textit{Right (student training)}}: the frozen teacher constructs the pseudo-target $x_{\mathrm{tgt}}^{A\to B}$ from $x_{\mathrm{src}}^B$ and $x_{\mathrm{tgt}}^A$. The student $G_\theta$, initialised from the teacher,
  swaps identity $A$ back onto this pseudo-target and is supervised toward the real image $x_{\mathrm{tgt}}^A$ by the reconstruction loss.
  }
  \label{fig:training}
\end{figure*}

%% file: figures/table_identity_bias.tex
\begin{table}[t]
  \centering
  \footnotesize
  \setlength{\tabcolsep}{3.5pt}
  \caption{\textbf{Identity encoder bias.} Identity similarity under three ArcFace encoders. $^{\dagger}$ marks the encoder the method is trained against.
  }
  \label{tab:identity_bias}
  \begin{tabular}{lccc}
  \toprule
  Method & \textit{MS1M--IR-SE50} & \textit{Glint--R100} & \textit{WebFace--R50} \\
  \midrule
  REFace       & $0.64^{\dagger}$ & 0.41 & 0.39 \\
  InfoSwap     & $0.58^{\dagger}$ & 0.43 & 0.42 \\
  \Trip{1} (ours) & 0.53 & $0.51^{\dagger}$ & 0.48 \\
  \bottomrule
  \end{tabular}
\end{table}

%% file: figures/figure_ffhq_selected_pairs.tex
\begingroup
\ifdefined\qualseltile\else\newlength{\qualseltile}\fi
\setlength{\qualseltile}{0.081\textwidth}

\newcommand{\qualsellabel}[1]{%
  \raisebox{0.5\qualseltile}{%
    \rotatebox[origin=c]{90}{\makebox[\qualseltile][c]{\small #1}}%
  }%
}

\begin{figure*}[t]
  \centering
  \small
  \setlength{\tabcolsep}{0.5pt}
  \renewcommand{\arraystretch}{0.95}
  \begin{tabular}{@{}l cc @{\hspace{6pt}} *{8}{c}@{}}
    & & & \multicolumn{3}{c}{Ours \MethodName} & & & & & \\
    \cmidrule(lr){4-6}
    & Source & Target
    & Teacher$^{\ddagger}$
    & $\lambda_\text{rec}=1$
    & $\lambda_\text{rec}=3$
    & InSwapper & CSCS & InfoSwap & REFace & APPLE \\
      \qualsellabel{Naturalness}
  &   \includegraphics[width=\qualseltile]{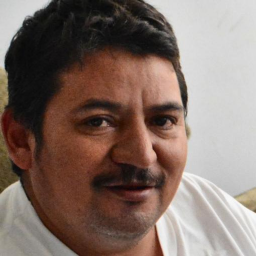}%

  &   \includegraphics[width=\qualseltile]{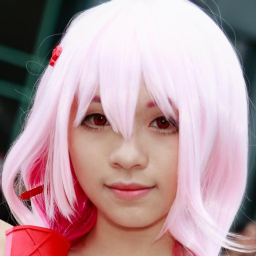}%

  &   \includegraphics[width=\qualseltile]{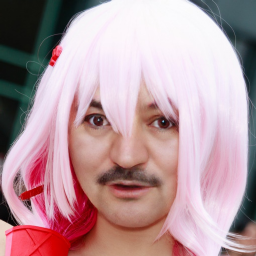}%

  &   \includegraphics[width=\qualseltile]{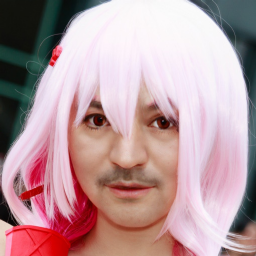}%

  &   \includegraphics[width=\qualseltile]{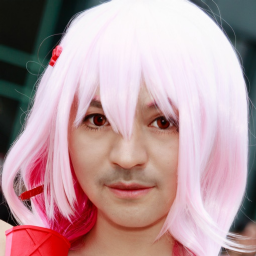}%

  &   \includegraphics[width=\qualseltile]{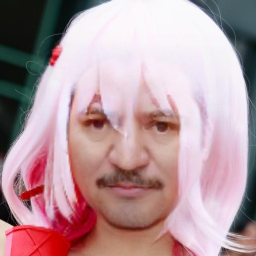}%

  &   \includegraphics[width=\qualseltile]{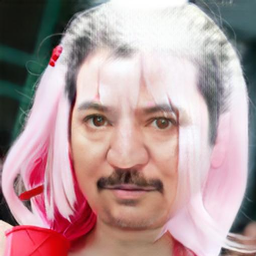}%

  &   \includegraphics[width=\qualseltile]{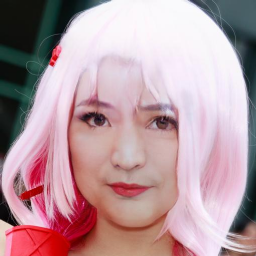}%

  &   \includegraphics[width=\qualseltile]{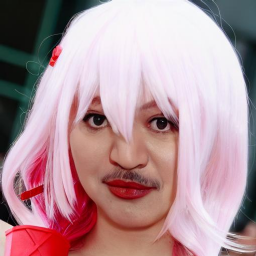}%

  &   \includegraphics[width=\qualseltile]{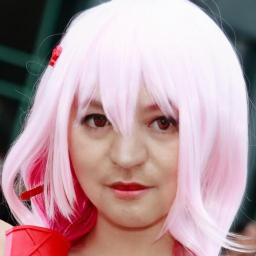}%
 \\

      \qualsellabel{Age}
  &   \includegraphics[width=\qualseltile]{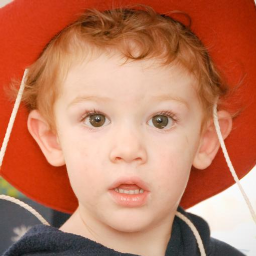}%

  &   \includegraphics[width=\qualseltile]{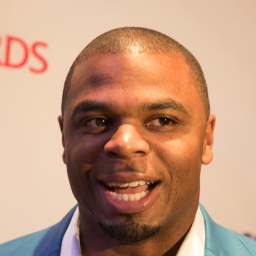}%

  &   \includegraphics[width=\qualseltile]{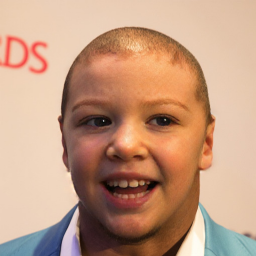}%

  &   \includegraphics[width=\qualseltile]{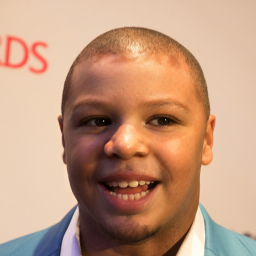}%

  &   \includegraphics[width=\qualseltile]{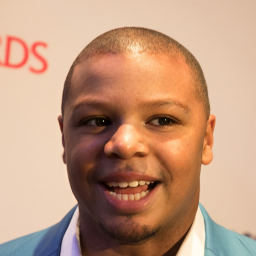}%

  &   \includegraphics[width=\qualseltile]{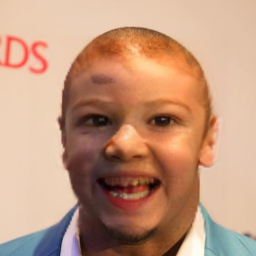}%

  &   \includegraphics[width=\qualseltile]{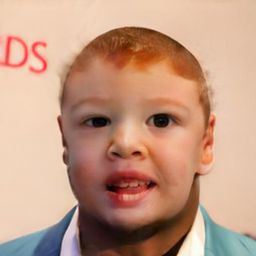}%

  &   \includegraphics[width=\qualseltile]{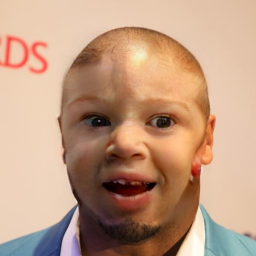}%

  &   \includegraphics[width=\qualseltile]{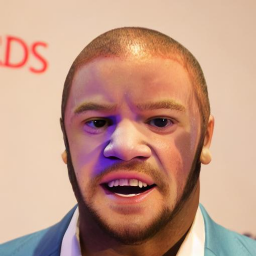}%

  &   \includegraphics[width=\qualseltile]{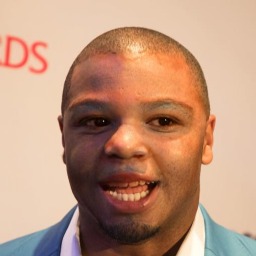}%
 \\

      \qualsellabel{Glasses}
  &   \includegraphics[width=\qualseltile]{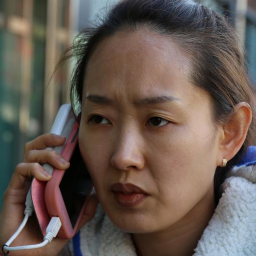}%

  &   \includegraphics[width=\qualseltile]{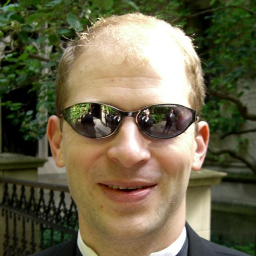}%

  &   \includegraphics[width=\qualseltile]{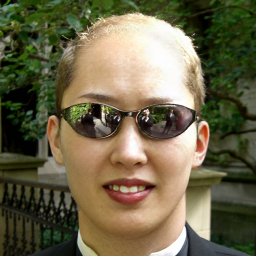}%

  &   \includegraphics[width=\qualseltile]{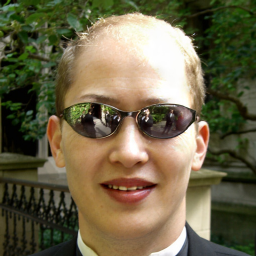}%

  &   \includegraphics[width=\qualseltile]{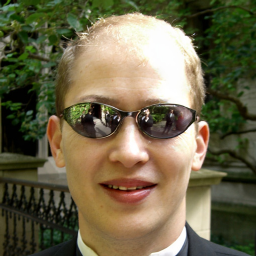}%

  &   \includegraphics[width=\qualseltile]{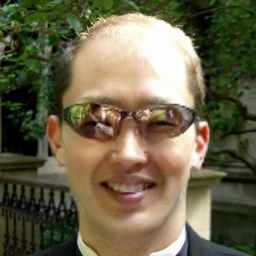}%

  &   \includegraphics[width=\qualseltile]{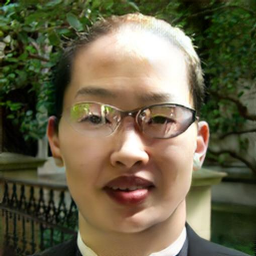}%

  &   \includegraphics[width=\qualseltile]{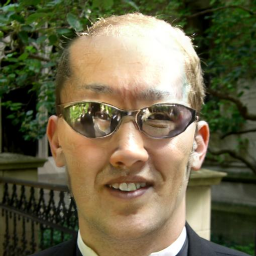}%

  &   \includegraphics[width=\qualseltile]{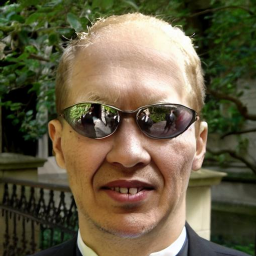}%

  &   \includegraphics[width=\qualseltile]{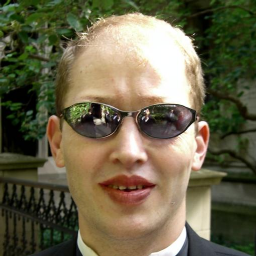}%
 \\

      \qualsellabel{Tattoo}
  &   \includegraphics[width=\qualseltile]{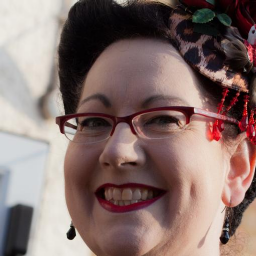}%

  &   \includegraphics[width=\qualseltile]{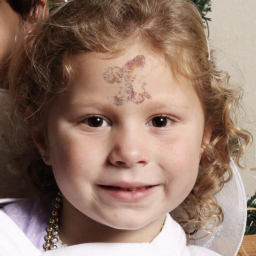}%

  &   \includegraphics[width=\qualseltile]{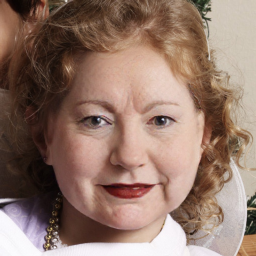}%

  &   \includegraphics[width=\qualseltile]{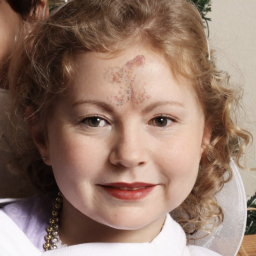}%

  &   \includegraphics[width=\qualseltile]{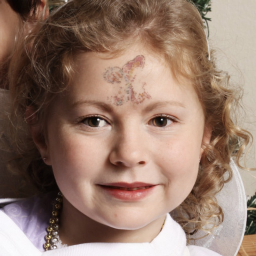}%

  &   \includegraphics[width=\qualseltile]{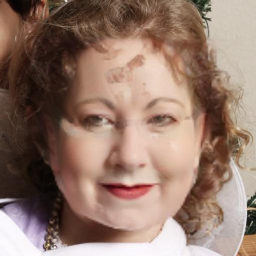}%

  &   \includegraphics[width=\qualseltile]{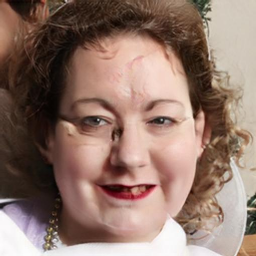}%

  &   \includegraphics[width=\qualseltile]{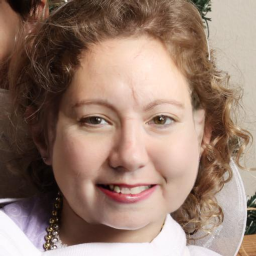}%

  &   \includegraphics[width=\qualseltile]{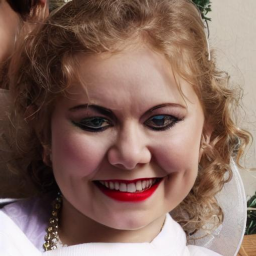}%

  &   \includegraphics[width=\qualseltile]{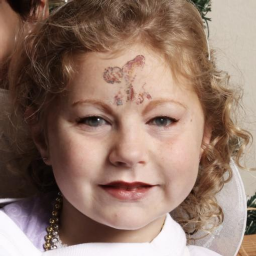}%
 \\

      \qualsellabel{Pose}
  &   \includegraphics[width=\qualseltile]{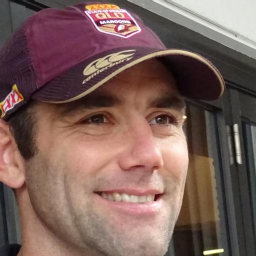}%

  &   \includegraphics[width=\qualseltile]{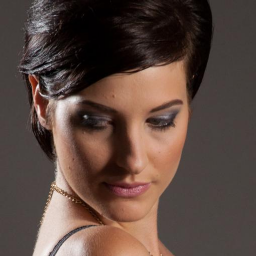}%

  &   \includegraphics[width=\qualseltile]{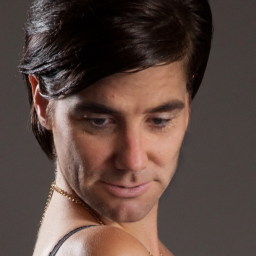}%

  &   \includegraphics[width=\qualseltile]{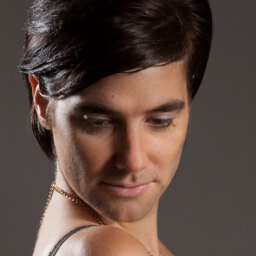}%

  &   \includegraphics[width=\qualseltile]{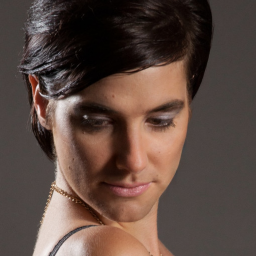}%

  &   \includegraphics[width=\qualseltile]{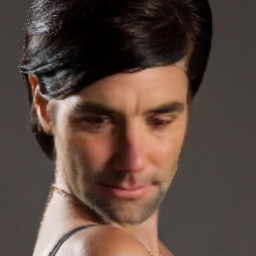}%

  &   \includegraphics[width=\qualseltile]{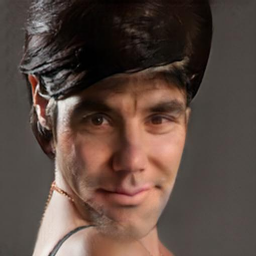}%

  &   \includegraphics[width=\qualseltile]{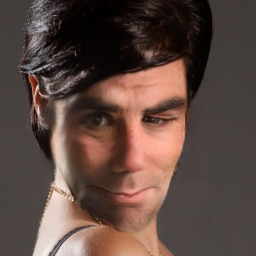}%

  &   \includegraphics[width=\qualseltile]{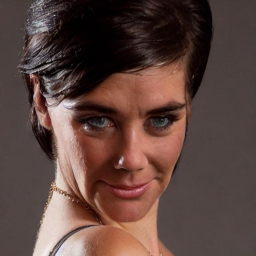}%

  &   \includegraphics[width=\qualseltile]{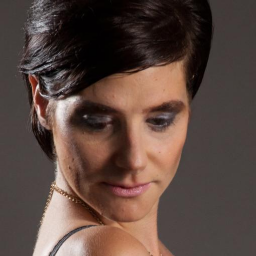}%
 \\

      \qualsellabel{Expression}
  &   \includegraphics[width=\qualseltile]{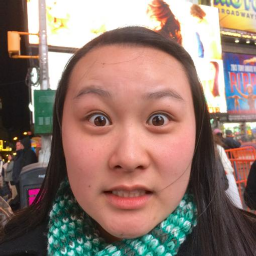}%

  &   \includegraphics[width=\qualseltile]{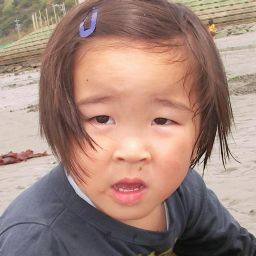}%

  &   \includegraphics[width=\qualseltile]{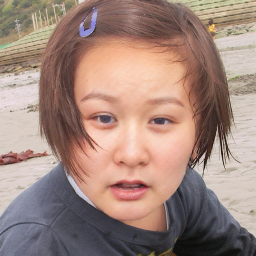}%

  &   \includegraphics[width=\qualseltile]{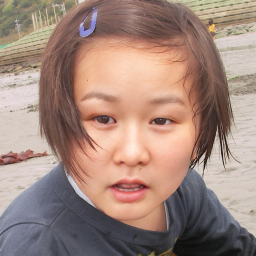}%

  &   \includegraphics[width=\qualseltile]{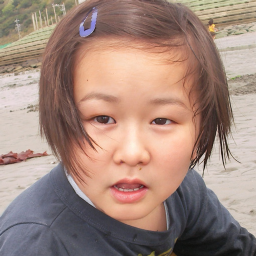}%

  &   \includegraphics[width=\qualseltile]{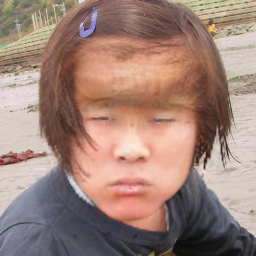}%

  &   \includegraphics[width=\qualseltile]{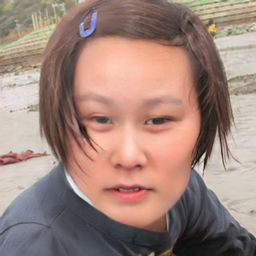}%

  &   \includegraphics[width=\qualseltile]{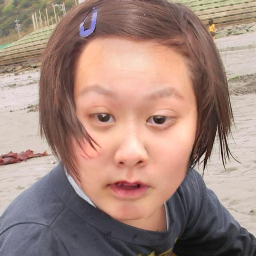}%

  &   \includegraphics[width=\qualseltile]{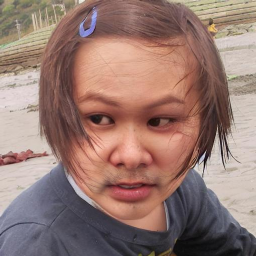}%

  &   \includegraphics[width=\qualseltile]{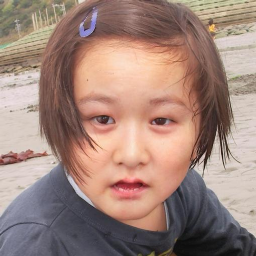}%
 \\

  \end{tabular}

  \caption{\textbf{Qualitative comparison on selected FFHQ source--target
  pairs.} The first two columns give the source and the target. The remaining columns show the swap produced by each method, with Teacher$^{\ddagger}$ the Teacher with the OminiControl branch off as in Table~\ref{tab:ffhq-bench1k-WebFaceR50} and $\lambda_\text{rec}$ the students. Every row probes a different difficulty. \textit{Naturalness} transfers an identity onto a target with unusual hair, and \textit{Age} pairs a source and a target of very different ages. The last four rows test how well the target's attributes survive the swap: glasses, tattoo, head pose, and facial expression.
  }
  \label{fig:qualitative-selected-compact}
\end{figure*}

\endgroup

%% file: figures/table_metrics_WebFaceR50_wretrieval.tex
\begin{table}[t]
  \centering
  \small
  \setlength{\tabcolsep}{3.5pt}
  \caption{
    \textbf{Results on 1,000 pairs from FFHQ}.
    Identity similarity and retrieval are measured with
    \textit{WebFace--R50}. \textbf{Bold} shows the best value in
    each column. The second and third best are \underline{underlined}.
  }
  \label{tab:ffhq-bench1k-WebFaceR50}
  \begin{tabular}{lrrcrr}
    \toprule
    Method
      & FID $\downarrow$
      & ID Sim. $\uparrow$
      & \makecell{Retrieval $\uparrow$ \\ Top-1/5}
      & Pose $\downarrow$
      & Expr. $\downarrow$ \\
    \midrule
    \multicolumn{6}{l}{\textit{One-step methods}} \\
    MegaFS
      & 15.54 & 0.38 & 84.5/89.7 & 4.50 & 1.14 \\
    E4S
      & 12.00 & 0.37 & 77.9/87.9 & 4.47 & 1.30 \\
    HiFiFace
      & 11.08 & 0.45 & 93.1/96.7 & 3.20 & 1.34 \\
    InSwapper$^{\dagger}$
      & 8.67 & $\,^\dagger$\underline{0.60}
      & 90.3/90.5 & 2.80 & 1.52 \\
    CSCS
      & 8.59 & \textbf{0.63}
      & \textbf{99.4}/\textbf{100} & 3.70 & 1.45 \\
    SimSwap
      & 5.90 & 0.38 & 81.9/85.8 & 1.83 & 0.96 \\
    FaceDancer
      & 4.86 & 0.34 & 78.8/90.3 & 2.27 & 0.74 \\
    InfoSwap
      & 3.88 & 0.42 & 91.2/96.1 & 3.18 & 0.99 \\
    \midrule
    \multicolumn{6}{l}{\textit{Multi-step methods}} \\
    FaceAdapter
      & 9.33 & 0.47 & 96.1/\underline{99.0} & 4.01 & 1.20 \\
    DiffSwap
      & 6.94 & 0.26 & 54.3/72.7 & 2.61 & 1.02 \\
    REFace
      & 5.45 & 0.39 & 88.8/96.2 & 3.68 & 1.04 \\
    APPLE
      & \underline{3.22} & 0.40 & 90.0/96.2
      & 1.86 & \underline{0.63} \\
    \midrule
    \multicolumn{6}{l}{\textit{\textbf{Ours \MethodName}, one-step method}} \\
    Teacher$^{\ddagger}$
      & 4.99 & \underline{0.55} & \underline{97.9}/\underline{99.1} & 2.05 & 0.81 \\
    Teacher
      & 4.07 & 0.48 & 95.8/98.0 & \underline{1.71} & 0.68 \\
    \Trip{1}
      & \underline{3.86} & 0.48
      & \underline{96.3}/98.6 & \underline{1.70} & \underline{0.67} \\
    \Trip{3}
      & \textbf{3.13} & 0.40 & 88.9/95.7
      & \textbf{1.45} & \textbf{0.56} \\
    \bottomrule
  \end{tabular}

  \vspace{2pt}
  {\footnotesize
    $^{\dagger}$ InSwapper generates face swaps using the same identity
    model employed here for evaluation. $^{\ddagger}$ OminiControl branch off at inference (Section~\ref{sec:comparison}). All other rows have it on. Under \textit{MS1M--IR-SE50},
    \Trip{1} obtains 0.53 and InSwapper obtains 0.56.
  }
\end{table}

%% file: figures/figure_operating_points.tex
\begin{figure*}[t]
  \centering
  \includegraphics[width=\textwidth]{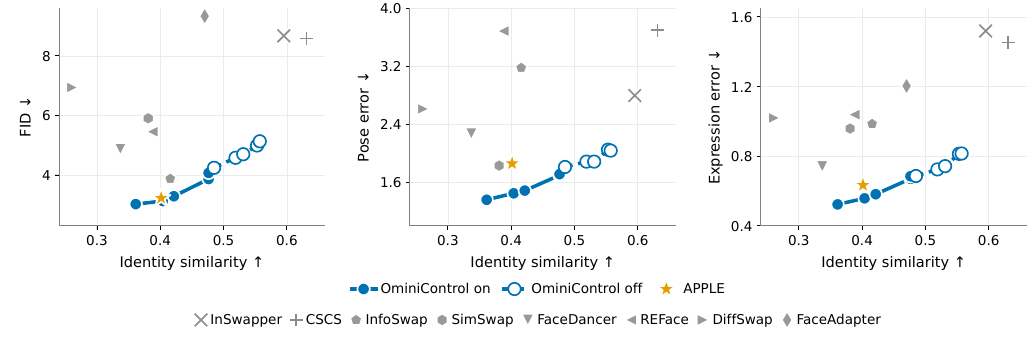}
  \caption{\textbf{Operating points of \MethodName{} on 1,000 pairs from FFHQ}. Lower is better on every vertical axis. Students 7, 3, 2, and 1 and the Teacher with the OminiControl branch on, the setting of Table~\ref{tab:ffhq-bench1k-WebFaceR50}, and off. MegaFS, E4S, and HiFiFace lie outside the FID range.}
  \label{fig:operating-points}
\end{figure*}

%% file: figures/figure_pulid_sweep.tex
\begin{figure}[t]
  \centering
  \includegraphics[width=0.88\linewidth]{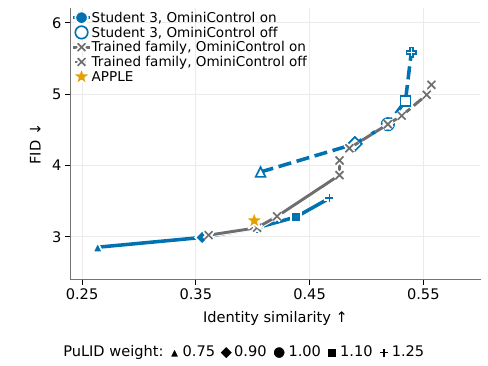}
  \caption{\textbf{A single model covers the frontier of separately trained students.} The PuLID identity weight of \Trip{3} is swept from $0.75$ to $1.25$ with the OminiControl branch on and off. Grey: Students 7, 3, 2, 1 and the Teacher at PuLID weight $1$.}
  \label{fig:pulid-sweep}
\end{figure}

%% file: figures/table_ablation.tex
\begin{table}[t]
  \centering
  \small
  \setlength{\tabcolsep}{2.5pt}
  \caption{\textbf{Ablation study of the teacher training} on 1,000 pairs from FFHQ. Our Teacher uses the $0.75\,/\,0.25$ data mixture, diffusion-feature discriminator, and adversarial (GAN) loss. Each row modifies one component. Checkpoint selection is described in \suppsec{app:implementation_details}.
  }
  \label{tab:ablation}
  \begin{tabular}{lcccc}
  \toprule
  & FID $\downarrow$ & ID Sim. $\uparrow$ & Pose $\downarrow$ & Expr. $\downarrow$ \\
  \midrule
  \textbf{Teacher} & 4.07 & 0.48 & 1.71 & 0.68 \\
  \midrule
  \multicolumn{5}{l}{\textit{Data mixture (cross\,/\,same ID)}} \\
  1.00\,/\,0.00        &  4.57 & 0.48 &  1.98 & 0.83 \\
  0.50\,/\,0.50        &  4.15 & 0.48 &  1.65 & 0.68 \\
  \midrule
  \multicolumn{5}{l}{\textit{Discriminator input}} \\
  Raw latent (1.00\,/\,0.00) & 17.01 & 0.52 &  2.55 & 1.01 \\
  \midrule
  \multicolumn{5}{l}{\textit{Realism signal}} \\
  None$^{\dagger}$ &  5.58 & 0.48 &  1.78 & 0.77 \\
  DMD$^{\dagger}$  &  4.85 & 0.48 &  1.71 & 0.73 \\
  DMD + GAN        &  4.16 & 0.47 &  1.73 & 0.72 \\
  \bottomrule
  \end{tabular}

  \vspace{2pt}
  {\footnotesize $^{\dagger}$ Evaluated at PuLID weight $0.95$ to match the Teacher's $0.48$ identity similarity (\suppsec{app:implementation_details}).}
\end{table}

%% file: figures/table_ablation_teacher_losses.tex
\begin{table*}[t]
  \centering
  \small
  \setlength{\tabcolsep}{7pt}
  \caption{\textbf{Ablation of the Teacher loss terms} on 1,000 pairs from FFHQ. DECA reports the held-out training expression loss $\mathcal{L}_{\mathrm{expr}}$ ($\times 10^3$), and Bg.\ reports the mean pixel-wise $\ell_1$ difference ($\times 10^2$) to the target outside the face on $[0, 1]$-normalised images (encoding and decoding the target through the VAE alone gives a Bg.\ floor of $0.98$). Setup and metric details are in \suppsec{app:implementation_details}.}
  \label{tab:ablation-teacher-losses}
  \begin{tabular}{lcccccc}
  \toprule
  & FID $\downarrow$ & ID Sim. $\uparrow$ & Pose $\downarrow$ & Expr. $\downarrow$ & DECA $\downarrow$ & Bg. $\downarrow$ \\
  \midrule
  \textbf{Full Teacher} & 3.80 & 0.47 & 1.61 & 0.67 & 8.4 & 1.08 \\
  \midrule
  w/o $\mathcal{L}_{\mathrm{pose}}$ & 3.88 & 0.47 & 1.90 & 0.71 & 8.6 & 1.09 \\
  w/o $\mathcal{L}_{\mathrm{expr}}$ & 3.86 & 0.47 & 1.66 & 0.68 & 14.3 & 1.10 \\
  w/o $\mathcal{L}_{\mathrm{bg}}$ & 3.80 & 0.46 & 1.58 & 0.66 & 8.3 & 1.41 \\
  w/o $\mathcal{L}_{\mathrm{pose}}$, $\mathcal{L}_{\mathrm{expr}}$, $\mathcal{L}_{\mathrm{bg}}$ & 4.26 & 0.49 & 2.05 & 0.79 & 19.4 & 1.72 \\
  $\mathcal{L}_{\mathrm{id}}$ and $\mathcal{L}_{\mathrm{GAN}}$ only & 6.53 & 0.55 & 4.62 & 1.17 & 32.6 & 3.79 \\
  \bottomrule
  \end{tabular}
\end{table*}

%% file: figures/table_ablation_student_losses.tex
\begin{table}[t]
  \centering
  \small
  \setlength{\tabcolsep}{2.6pt}
  \caption{\textbf{Ablation of the student loss terms} on 1,000 pairs from FFHQ for \Trip{3}. DECA reports the held-out training expression loss $\mathcal{L}_{\mathrm{expr}}$ ($\times 10^3$, see \suppsec{app:implementation_details}).}
  \label{tab:ablation-student-losses}
  \begin{tabular}{lccccc}
  \toprule
  & FID\,$\downarrow$ & ID Sim.\,$\uparrow$ & Pose\,$\downarrow$ & Expr.\,$\downarrow$ & DECA\,$\downarrow$ \\
  \midrule
  \textbf{Full \Trip{3}} & 3.31 & 0.41 & 1.50 & 0.58 & 7.1 \\
  \midrule
  w/o $\mathcal{L}_{\mathrm{pose}}$, $\mathcal{L}_{\mathrm{expr}}$ & 3.35 & 0.42 & 1.54 & 0.59 & 8.2 \\
  w/o $\mathcal{L}_{\mathrm{GAN}}$ & 3.35 & 0.40 & 1.42 & 0.56 & 6.7 \\
  w/o $\mathcal{L}_{\mathrm{rec}}$ & 4.78 & 0.50 & 1.94 & 0.83 & 10.0 \\
  w/o $\mathcal{L}_{\mathrm{id}}$ & 2.62 & 0.21 & 1.12 & 0.44 & 3.2 \\
  \bottomrule
  \end{tabular}
\end{table}

%% file: appendix.tex
\renewcommand{\topfraction}{0.95}
\renewcommand{\bottomfraction}{0.9}
\renewcommand{\textfraction}{0.05}
\renewcommand{\floatpagefraction}{0.7}
\renewcommand{\dbltopfraction}{1.0}
\renewcommand{\dblfloatpagefraction}{0.8}
\setcounter{totalnumber}{5}
\setcounter{topnumber}{3}
\makeatletter
\setlength{\@fptop}{0pt}
\setlength{\@fpsep}{12pt plus 4pt}
\setlength{\@fpbot}{0pt plus 1fil}
\setlength{\@dblfptop}{0pt}
\setlength{\@dblfpsep}{12pt plus 4pt}
\setlength{\@dblfpbot}{0pt plus 1fil}
\makeatother
\raggedbottom
\appendix
\maketitlesupplementary
\section{Implementation details}
\label{app:implementation_details}
\myparagraph{Main experiments} The generator runs at the fixed noise level $\sigma_{\mathrm{gen}} = 0.2$ of Equation~\ref{eq:generator}, and the discriminator reads $K = 4$ feature maps taken at the double-stream blocks $0$, $6$, $12$, and $18$ of the frozen transformer, at noise levels $\sigma \sim \mathcal{U}[0.01, 0.4]$. Both are rescaled by the scheduler's resolution-dependent shift. Noising the latents before feature extraction makes no clear difference to the identity--realism trade-off, so we keep the noise as in LADD.
We condition FLUX on the fixed prompt ``cinematic full head portrait''. The guidance scale of FLUX, which the model receives as an input embedding, is set to $4.5$ in every forward pass. All models use $\lambda_{\mathrm{id}} = 1$ and $\lambda_{\mathrm{GAN}} = 0.03$, and the discriminator loss $\mathcal{L}_{D_\varphi}$ is scaled by $0.1$. The teacher uses $\lambda_{\mathrm{rec}} = 2$, $\lambda_{\mathrm{bg}} = 0.3$, $\lambda_{\mathrm{pose}} = 2$, and $\lambda_{\mathrm{expr}} = 2$. The students use $\lambda_{\mathrm{pose}} = \lambda_{\mathrm{expr}} = 1$ inside the triplet objective and vary only $\lambda_{\mathrm{rec}}$. We take one discriminator step per generator step. Both networks are optimised with AdamW at a learning rate of $5\times10^{-5}$, weight decay $0.01$, $\beta = (0.9, 0.999)$, and gradient clipping at $1.0$, using $500$ warmup steps followed by a constant schedule.
The estimators used for training, 6DRepNet and DECA, differ from those used for evaluation, so the reported pose and expression metrics are not measured with the networks the generator was optimised against. Experiments run on 4xH100 GPUs with a batch size of $5$ per GPU, giving an effective batch of $20$.

\myparagraph{OminiControl branch} With the branch on, the clean target latent enters the transformer a second time as a third token stream next to the text and image streams, in the way OminiControl~\citep{tan2025ominicontrol} appends condition tokens: the tokens pass the same patch embedding, carry the same positional indices as the image tokens, and attend jointly with the text and image tokens in every double- and single-stream block. The stream uses the same weights as the image stream, including our LoRA adapters, and differs from it in two respects: its timestep embedding is computed for noise level $0$ instead of $\sigma_{\mathrm{gen}}$, with the same guidance embedding and pooled prompt, and the PuLID cross-attention acts on the image stream alone. The velocity is read from the image stream. The branch has no parameters of its own and nothing is trained for it, since training runs with the branch off. OminiControl and APPLE~\citep{kang2026attribute} instead train a LoRA that acts on the condition tokens only. The frozen Teacher that generates the pseudo-targets of the triplet group runs with the branch on and PuLID weight $1$. With the branch on, the transformer processes twice as many image tokens. The timing in Section~\ref{app:timing} is measured in this setting.

\myparagraph{Identity loss input} The masked source and generated images are resized to $112\times112$ without further alignment before entering the identity encoder of $\mathcal{L}_{\mathrm{id}}$.

\myparagraph{Dataset filtering} We discard images whose aesthetic score~\citep{schuhmann2022aestheticpredictor} is below $5.1$, the threshold of~\citep{kang2026attribute}. To clean an identity, we take as anchor the face with the highest average identity similarity to the others and keep the faces whose similarity to it lies between $0.6$ and $0.95$, admitting a face only if its similarity to every already accepted face is below $0.95$ to avoid duplicates.

\myparagraph{Training time} Our teacher model is trained on 100K training samples, which takes about 4.5 hours of wall-clock time on 4xH100. Each student is trained on 15K training samples in about 50 minutes on 4xH100.

\myparagraph{Evaluation on FFHQ} The $1{,}000$ pairs swap FFHQ image $69000+i$ onto image $68000+i$ for $i = 0, \dots, 999$. For HiFiFace, FID, pose, and expression are computed on $999$ images, as one swap failed.

\myparagraph{Evaluation on CelebA-HQ} The $1{,}000$ pairs are the test list of REFace: image $29000+i$ is swapped onto image $28000+i$ for $i = 0, \dots, 999$. The official $1024 \times 1024$ images are resized to $512 \times 512$ without re-alignment, face masks are computed as for FFHQ, and FID uses CLIP features of all $30{,}000$ CelebA-HQ images. Every baseline runs with the settings used on FFHQ.

\myparagraph{Ablation runs} Quality saturates after roughly $70$K training samples, and further training does not improve it. Beyond that point, individual checkpoints differ slightly in how they trade realism against identity transfer, which would confound a comparison between ablation runs. In Table~\ref{tab:ablation}, we therefore select for each run of the data mixture ablation a post-convergence checkpoint whose identity similarity closely matches that of the others, so that the differences fall on FID, pose, and expression rather than on identity. The reported teacher of the main experiments is the checkpoint at $100$K samples, the cross-identity checkpoint ($1.00\,/\,0.00$) is selected at $70$K samples, and the equal-mixture checkpoint ($0.50\,/\,0.50$) at $80$K samples. The raw-latent discriminator run is trained for $70$K samples on cross-identity pairs and its final checkpoint is reported. The realism-signal runs in Table~\ref{tab:ablation} use the checkpoint at $100$K samples. Because removing $\mathcal{L}_{\mathrm{GAN}}$ (None) or replacing it with DMD slightly increases identity similarity at default PuLID weight $1.0$ ($0.50$ vs.\ $0.48$, with FID $6.23$ and $5.34$), Table~\ref{tab:ablation} evaluates those two rows at PuLID weight $0.95$, where their identity similarity matches the Teacher's ($0.48$).

\myparagraph{DMD realism baselines} In the realism-signal ablation of Table~\ref{tab:ablation}, distribution matching distillation (DMD)~\citep{yin2024one} replaces or complements the adversarial loss in the same manner as RDMD~\citep{rakitin2024regularized}, acting as a realism regulariser rather than distilling a multi-step face-swapping teacher. The real-score network is the frozen FLUX.1-Krea-dev model conditioned on the fixed portrait prompt without source or target conditioning and without classifier-free guidance, while the fake-score network is the same model equipped with a rank-$64$ LoRA updated five times per generator step. The DMD loss weight is set to $1$ with noise levels sampled from $[0.01, 0.4]$. In the combined DMD + GAN variant, following DMD2~\citep{yin2024improved}, the discriminator heads read intermediate features from the fake-score network and are updated alongside it. On a single GPU without compilation, DMD and DMD + GAN increase wall-clock training time per step by $2.7\times$ and $3.6\times$ relative to the adversarial loss alone.

\myparagraph{Loss ablations and held-out attribute metrics} The leave-one-out loss ablations in Tables~\ref{tab:ablation-teacher-losses} and~\ref{tab:ablation-student-losses} are trained for $100$K samples (Teacher) and $15$K samples (\Trip{3}) on a single GPU with batch size $5$ and gradient accumulation over $4$ steps (effective batch size $20$), and are evaluated at default PuLID weight $1.0$ with the OminiControl branch on against the full Teacher and \Trip{3} recipes trained under the same single-GPU setup (which thus differ slightly from the 4-GPU runs in Table~\ref{tab:ffhq-bench1k-WebFaceR50}). To directly evaluate the attributes targeted by $\mathcal{L}_{\mathrm{bg}}$ and $\mathcal{L}_{\mathrm{expr}}$ on the $1{,}000$ held-out FFHQ pairs, Tables~\ref{tab:ablation-teacher-losses} and~\ref{tab:ablation-student-losses} report two additional metrics: Bg.\ ($\times 10^2$, Table~\ref{tab:ablation-teacher-losses}) is the mean pixel-wise $\ell_1$ difference to the target outside the face mask of the evaluation parser dilated by $8$ pixels on $[0, 1]$-normalised images (with glasses included in the background region, where encoding and decoding the target through the FLUX VAE alone sets a lower bound of $0.98 \times 10^{-2}$), and DECA ($\times 10^3$, Tables~\ref{tab:ablation-teacher-losses} and~\ref{tab:ablation-student-losses}) is the held-out training expression loss $\mathcal{L}_{\mathrm{expr}}$. While removing $\mathcal{L}_{\mathrm{expr}}$ increases the held-out DECA error from $8.4$ to $14.3$, the evaluation protocol's Expr.\ metric (Deep3DFace on the full frame) stays at $0.68$ vs.\ $0.67$ because it is largely insensitive to upper-face geometry.

\section{Additional discussion of the identity-encoder bias in prior evaluation protocols}
\label{app:identity_bias}

\myparagraph{Results under the previous, biased protocol, with \textit{MS1M--IR-SE50}}
Table~\ref{tab:ffhq-bench1k-MS1M-IRSE50} shows metrics where the identity similarity is scored with \textit{MS1M--IR-SE50}.
From this perspective, REFace and InfoSwap seem to be very strong methods: InfoSwap achieves very high identity similarity with one of the best FIDs, and REFace achieves even higher identity similarity. Under the incorrect protocol, there is no model that is strictly better than REFace and InfoSwap. As Table~\ref{tab:ffhq-bench1k-WebFaceR50} shows, this conclusion is incorrect under the unbiased protocol. InfoSwap performs worse than our \Trip{1} in all four metrics, and REFace is worse than APPLE and several of our models.
\input{figures/table_metrics_MS1M_IRSE50_appendix}

\myparagraph{Comparison of the identity encoders}
\textit{MS1M--IR-SE50} and \textit{WebFace--R50} share a 50-layer backbone, the ArcFace objective, and $512$-dimensional embeddings,
and differ in architecture and training data.
\textit{MS1M--IR-SE50} uses an IR-SE backbone trained on MS1MV2, a refined version of MS-Celeb-1M with $5{,}822{,}653$ images of $85{,}742$ identities.
\textit{WebFace--R50} uses an iResNet-50 backbone trained on WebFace600K, also known as WebFace12M, derived from the cleaned WebFace42M collection~\citep{zhu2021webface260m} and containing more than $12$ million images of roughly $600{,}000$ identities.
Its training set is therefore about twice as large in images and seven times as large in identities.
Combined with the benchmark results reported in~\citep{insightface_model_zoo},
where \textit{WebFace--R50} outperforms \textit{MS1M--IR-SE50},
it makes \textit{WebFace--R50} the better-suited model for evaluation.

\myparagraph{Systematic over-estimation with \textit{MS1M--IR-SE50}} Beyond favouring the methods trained against it, \textit{MS1M--IR-SE50} assigns higher identity similarity than the other two identity encoders, \textit{Glint--R100} and \textit{WebFace--R50}, to almost every method. A plausible explanation is its smaller and less diverse training corpus, which yields a less discriminative embedding space in which distinct identities are separated less sharply.

\section{Extended discussion of qualitative results}
\label{app:discussion}

In this section, we present more examples showcasing other complex cases in Figure~\ref{fig:qualitative-selected-other}: a gender change, a source and a target of different skin tones, a target with a piercing, and a source wearing glasses.
Figure~\ref{fig:qualitative-selected-omitted} extends Figure~\ref{fig:qualitative-selected-compact} with the methods omitted there. Figures~\ref{fig:ablation-qual-realism} to~\ref{fig:ablation-qual-preservation} provide qualitative comparisons for the teacher ablations of Section~\ref{sec:ablation}, and Figures~\ref{fig:curated-grid} to~\ref{fig:curated-grid-3} compare our models with the strongest baselines on further pairs grouped by scenario: make-up and accessories, occlusion, naturalness and image quality, glasses, and pose and expression.

\myparagraph{More examples} InfoSwap fails to transfer gender. REFace shifts the skin tone toward the source, while the other methods, including ours, keep the target's. \MethodName{} and APPLE keep the target's piercing, and the others lose it. CSCS includes glasses from the source.

\input{figures/figure_ffhq_appendix_both}

\section{Results on CelebA-HQ and FaceForensics++}
\label{app:celebahq}
\input{figures/figure_celebahq_operating_points}

\input{figures/table_celebahq}

\input{figures/table_ffpp}

\myparagraph{CelebA-HQ} Table~\ref{tab:celebahq} reports the same comparison as Table~\ref{tab:ffhq-bench1k-WebFaceR50} on the 1,000 CelebA-HQ~\citep{karras2018progressive} test pairs of REFace~\citep{baliah2025realistic}, using the same metrics and checkpoints. Figure~\ref{fig:celebahq-operating-points} plots the operating points of \Trip{7}, \Trip{3}, \Trip{2}, \Trip{1}, and the Teacher on these pairs at PuLID identity weight $1$ with the OminiControl branch on and off, mirroring Figure~\ref{fig:operating-points} on FFHQ. \Trip{3} matches APPLE in FID ($4.18$ vs.\ $4.12$) and identity similarity ($0.46$ vs.\ $0.45$), while improving on it in pose ($1.42$ vs.\ $1.75$) and expression ($0.52$ vs.\ $0.57$). Switching the OminiControl branch off raises identity similarity by $0.05$ to $0.12$ at the expected trade-off in FID, pose, and expression.

\myparagraph{FaceForensics++} Table~\ref{tab:ffpp} evaluates all methods on FaceForensics++ (FF++)~\citep{rossler2019faceforensics}, using its $1{,}000$ original videos in the high-quality (HQ, c23, lightly compressed H.264) setting and its official $1{,}000$ directed source--target video pairs. FaceShifter~\citep{li2020faceshifter} introduced face-swapping evaluation on FF++ by sampling ten evenly spaced frames per video, a protocol adopted by subsequent methods~\citep{chen2020simswap, wang2021hififace, zhu2021one, rosberg2023facedancer, han2024face} without released frame lists, crops, or evaluation code. To match the $1{,}000$-pair size of our FFHQ and CelebA-HQ benchmarks, we divide each video into ten equal segments and take the frame at the centre of the fifth segment ($45\%$ of the video length) for both the source and the target video, align the largest detected face to $512 \times 512$ with the FFHQ dlib landmark crop, and score all methods on the resulting $1{,}000$ crops using the metrics and estimators of Table~\ref{tab:ffhq-bench1k-WebFaceR50}. FID is computed against $9{,}999$ real face crops extracted in the same way from all ten segments across the $1{,}000$ videos, and each swap's own target crop is excluded from the $1{,}000$-crop retrieval gallery so that, as on FFHQ and CelebA-HQ, the target image is not among the candidates. Across our models, increasing $\lambda_{\mathrm{rec}}$ and enabling the OminiControl branch trace the same trade-off as on the other two datasets: \Trip{3} matches APPLE's identity similarity ($0.53$) while achieving lower FID ($1.77$ vs.\ $2.19$), pose error ($1.33$ vs.\ $1.87$), and expression error ($0.53$ vs.\ $0.74$).

\section{Inference cost}
\label{app:timing}

Table~\ref{tab:timing} reports the wall-clock time of one swap for \MethodName{} and for the APPLE student, measured on the same GPU model with the same boundary: from the decoded source and target images in memory to the swapped image as a tensor on the GPU. The time includes every step the method needs for a new source and target (face parsing, identity embedding, VAE encoding and decoding, and the transformer forward passes) and excludes disk reads and PNG encoding. Both methods run in bf16 at $512\times512$ with batch size $1$ and without compilation. We time $100$ benchmark pairs after $10$ untimed warm-up pairs. APPLE is run with its released code and checkpoint in the environment specified by its repository, with the default $28$ sampling steps and no classifier-free guidance, so that one step is one transformer forward. \MethodName{}'s time is the same for the Teacher and the students, which share the architecture.

\begin{table}[tbp]
  \centering
  \footnotesize
  \setlength{\tabcolsep}{3pt}
  \caption{\textbf{Wall-clock time per swap} on one NVIDIA H100, batch size $1$, $512\times512$, bf16. Mean $\pm$ standard deviation over $100$ swaps.}
  \label{tab:timing}
  \begin{tabular}{lcc}
  \toprule
   & \MethodName{} & APPLE student \\
  \midrule
  NFE & 1 & 28 \\
  Whole swap, mean $\pm$ std & $0.292\pm0.003$\,s & $5.796\pm0.030$\,s \\
  \quad parsing and identity embedding & $0.08$\,s & $0.11$\,s \\
  \quad transformer, text encoder, VAE & $0.21$\,s & $5.67$\,s \\
  Peak GPU memory & $27$\,GB & $34$\,GB \\
  \bottomrule
  \end{tabular}
\end{table}

\section{Generator noise level}
\label{app:sigma}

Table~\ref{tab:sigma-ablation} varies the noise level at which the target is read (Equation~\ref{eq:generator}). All runs use the 0.75\,/\,0.25 mixture of Table~\ref{tab:ablation} and are trained without discriminator latent noise, so the $0.2$ row is a control rather than the Teacher of Table~\ref{tab:ffhq-bench1k-WebFaceR50}. Small values move the model to the attribute-preserving end of the trade-off, with lower FID, pose, and expression errors at lower identity similarity, and values from $0.15$ to $0.5$ favour identity transfer. At $\sigma_{\mathrm{gen}} = 1$ the generator returns the target almost unchanged, and the apparently good realism and attribute scores do not correspond to a swap. $\sigma_{\mathrm{gen}}$ is therefore a second knob on the same trade-off as $\lambda_{\mathrm{rec}}$. We choose $0.2$ from the $0.15$ to $0.5$ range, where identity similarity is adequate and FID is lowest at $0.2$. Between $0.15$ and $0.2$ the choice depends on whether slightly higher identity similarity ($0.15$) or slightly lower FID, pose, and expression errors ($0.2$) are preferred.

\input{figures/table_sigma_ablation}

\section{Dataset filtering}
\label{app:filtering}

We filter the training data in two steps (Algorithm~\ref{alg:dataset-filtering}): images below an aesthetic-score threshold $\tau_{\mathrm{AES}}$ are dropped, as in APPLE~\citep{kang2026attribute}, and each identity group is reduced to an anchor and the images that are similar enough to it ($\tau_{\mathrm{ID}}$) without being near-duplicates, with groups of fewer than two images removed. Table~\ref{tab:dataset-ablation} shows that this step matters little for our training. A teacher trained on the unfiltered data is $0.2$ worse in FID and $0.01$ better in identity similarity than the Teacher of Table~\ref{tab:ffhq-bench1k-WebFaceR50}, with pose and expression errors within $0.02$. This is less than the distance between neighbouring operating points in Figure~\ref{fig:operating-points}. Changing the thresholds moves the numbers by similar amounts. The one larger change, $\tau_{\mathrm{ID}} = 0.7$, lowers the FID, pose, and expression errors together with the identity similarity, which is the usual trade-off and not an improvement. We keep the aesthetic threshold of APPLE for comparability and $\tau_{\mathrm{ID}} = 0.6$.

\input{figures/table_dataset_ablation}

\input{figures/algorithm_dataset_filtering}

\FloatBarrier
\section{Operating-point controls at inference}
\input{figures/figure_sweeps_combined}

\input{figures/figure_pulid_sweep_grid}

\input{figures/figure_pulid_vs_family}

\label{app:controls}

Figure~\ref{fig:operating-points} shows every model of Table~\ref{tab:ffhq-bench1k-WebFaceR50}, and \Trip{2} and \Trip{7}, with the OminiControl branch on and off. Switching the branch off raises identity similarity by $0.07$ to $0.13$ and degrades FID, pose, and expression, and the two settings are the two halves of that frontier. Figure~\ref{fig:pulid-sweep-full} extends Figure~\ref{fig:pulid-sweep} to pose and expression: in both settings the PuLID sweep of \Trip{3} follows the separately trained students, and with the branch off it leaves them only above an identity similarity of $0.52$. Figure~\ref{fig:teacher-vs-student} shows that the swept Teacher lies above the swept \Trip{3} at every identity in all three metrics, so student training moves the curve rather than selecting a point on it.

Figure~\ref{fig:pulid-sweep-grid} shows the same sweep of \Trip{3} on three pairs of Figure~\ref{fig:qualitative-selected-compact} and one further pair with strong make-up. Figures~\ref{fig:pulid-vs-family-on} and~\ref{fig:pulid-vs-family-off} place the sweep next to the trained students on five further pairs with a change of gender, age, or skin tone, with the branch on and off. With the branch on, the two rows are close at every weight. With the branch off, the trained models move further toward the source than the sweep.

\input{figures/figure_ablation_examples}

\clearpage
\input{figures/figure_curated_grids}

%% file: figures/table_metrics_MS1M_IRSE50_appendix.tex
\begin{table}[tbp]
  \centering
  \footnotesize
  \setlength{\tabcolsep}{3pt}
  \caption{
      \textbf{Results on 1,000 pairs from FFHQ with the previous, biased protocol. Identity metric is scored with
      \textit{MS1M--IR-SE50}.} FID, pose, and expression follow the protocol of Table~\ref{tab:ffhq-bench1k-WebFaceR50}. Only the identity evaluator is changed. $^{\ddagger}$: OminiControl branch off at inference, as in Table~\ref{tab:ffhq-bench1k-WebFaceR50}. Methods are ordered by FID within each
      group, with better FID placed lower. \textbf{Bold} shows the best value
      in each column. The second and third best are \underline{underlined}.
  }
  \label{tab:ffhq-bench1k-MS1M-IRSE50}
  \begin{tabular}{lrrrr}
  \toprule
  Method & FID $\downarrow$ & ID Sim. $\uparrow$ & Pose $\downarrow$ & Expr. $\downarrow$ \\
  \midrule
  \multicolumn{5}{l}{\textit{One-step methods}} \\
  MegaFS & 15.54 & 0.48 & 4.50 & 1.14 \\
  E4S & 12.00 & 0.49 & 4.47 & 1.30 \\
  HiFiFace & 11.08 & 0.48 & 3.20 & 1.34 \\
  InSwapper & 8.67 & 0.56 & 2.80 & 1.52 \\
  CSCS & 8.59 & \underline{0.61} & 3.70 & 1.45 \\
  SimSwap & 5.90 & 0.49 & 1.83 & 0.96 \\
  FaceDancer & 4.86 & 0.48 & 2.27 & 0.74 \\
  InfoSwap & 3.88 & 0.58 & 3.18 & 0.99 \\
  \midrule
  \multicolumn{5}{l}{\textit{Multi-step methods}} \\
  FaceAdapter & 9.33 & 0.53 & 4.01 & 1.20 \\
  DiffSwap & 6.94 & 0.34 & 2.61 & 1.02 \\
  REFace & 5.45 & \textbf{0.64} & 3.68 & 1.04 \\
  APPLE & \underline{3.22} & 0.52 & 1.86 & \underline{0.63} \\
  \midrule
  \multicolumn{5}{l}{\textit{\textbf{Ours \MethodName}, one-step method}} \\
  Teacher$^{\ddagger}$ & 4.99 & \underline{0.59} & 2.05 & 0.81 \\
  Teacher & 4.07 & 0.53 & \underline{1.71} & 0.68 \\
  \Trip{1} & \underline{3.86} & 0.53 & \underline{1.70} & \underline{0.67} \\
  \Trip{3} & \textbf{3.13} & 0.47 & \textbf{1.45} & \textbf{0.56} \\
  \bottomrule
  \end{tabular}
\end{table}

%% file: figures/figure_ffhq_appendix_both.tex
\begingroup
\ifdefined\qualseltile\else\newlength{\qualseltile}\fi
\setlength{\qualseltile}{0.084\textwidth}
\newcommand{\qualsellabel}[1]{%
  \raisebox{0.5\qualseltile}{%
    \rotatebox[origin=c]{90}{\makebox[\qualseltile][c]{\small #1}}%
  }%
}

\begin{figure*}[tbp]
  \centering
  \small
  \setlength{\tabcolsep}{0.5pt}
  \renewcommand{\arraystretch}{0.95}
  \begin{tabular}{@{}l cc @{\hspace{6pt}} *{7}{c}@{}}
    & & & \multicolumn{2}{c}{Ours \MethodName\;Students} & & & & & \\
    \cmidrule(lr){4-5}
    & Source & Target
    & $\lambda_\text{rec}=1$
    & $\lambda_\text{rec}=3$
    & InSwapper & CSCS & InfoSwap & REFace & APPLE \\
      \qualsellabel{Gender}
  & \includegraphics[width=\qualseltile]{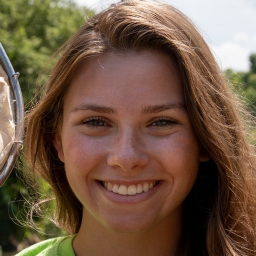} & \includegraphics[width=\qualseltile]{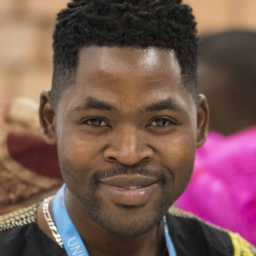}
  & \includegraphics[width=\qualseltile]{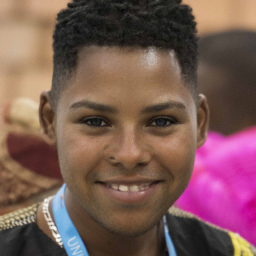} & \includegraphics[width=\qualseltile]{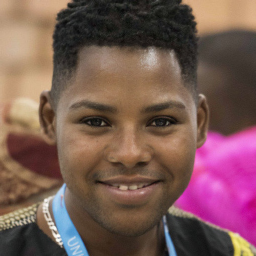}
  & \includegraphics[width=\qualseltile]{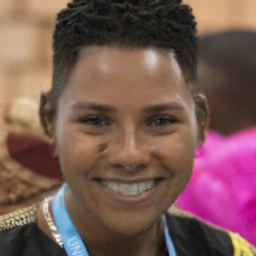} & \includegraphics[width=\qualseltile]{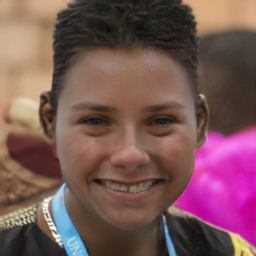}
  & \includegraphics[width=\qualseltile]{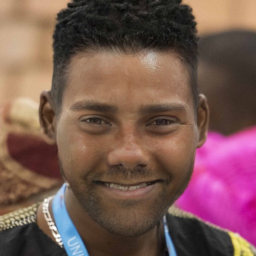} & \includegraphics[width=\qualseltile]{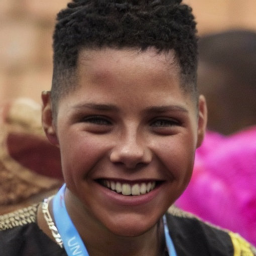} & \includegraphics[width=\qualseltile]{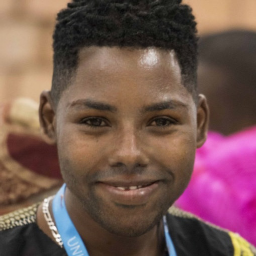} \\

      \qualsellabel{Skin tone}
  & \includegraphics[width=\qualseltile]{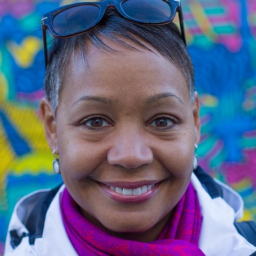} & \includegraphics[width=\qualseltile]{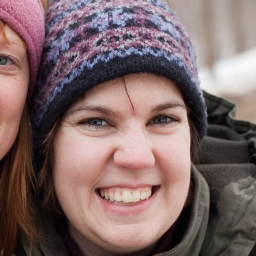}
  & \includegraphics[width=\qualseltile]{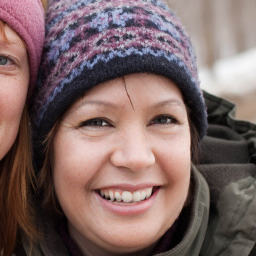} & \includegraphics[width=\qualseltile]{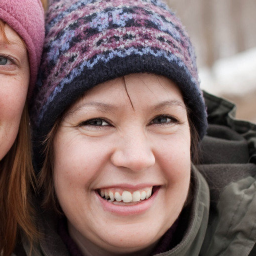}
  & \includegraphics[width=\qualseltile]{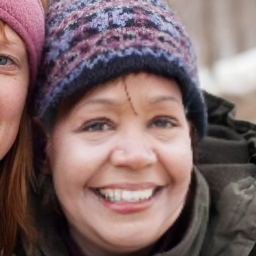} & \includegraphics[width=\qualseltile]{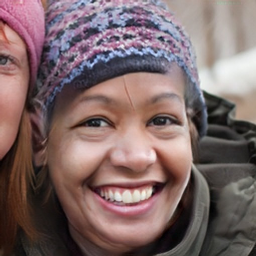}
  & \includegraphics[width=\qualseltile]{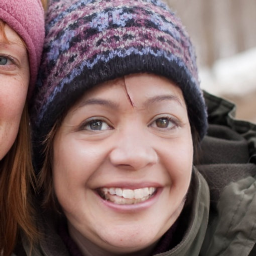} & \includegraphics[width=\qualseltile]{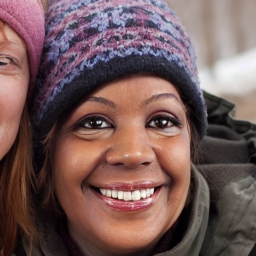} & \includegraphics[width=\qualseltile]{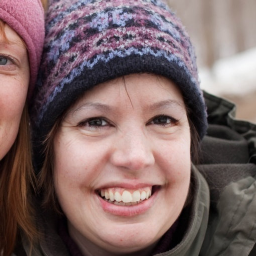} \\

      \qualsellabel{Piercing}
  & \includegraphics[width=\qualseltile]{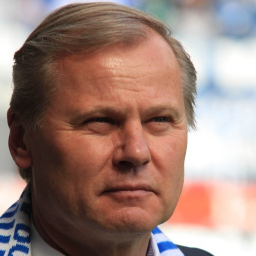} & \includegraphics[width=\qualseltile]{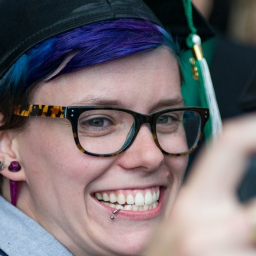}
  & \includegraphics[width=\qualseltile]{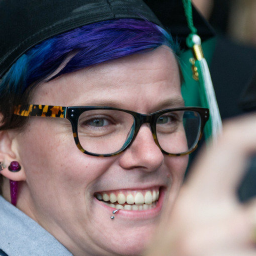} & \includegraphics[width=\qualseltile]{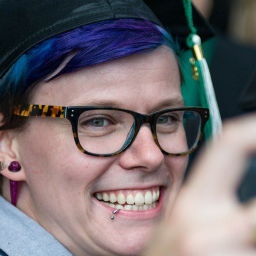}
  & \includegraphics[width=\qualseltile]{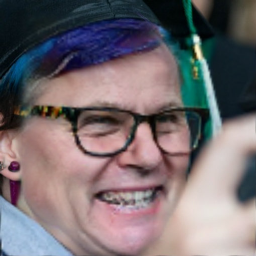} & \includegraphics[width=\qualseltile]{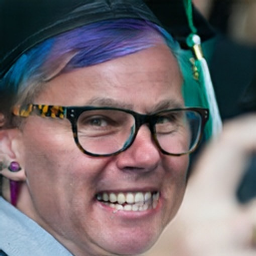}
  & \includegraphics[width=\qualseltile]{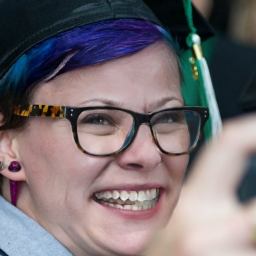} & \includegraphics[width=\qualseltile]{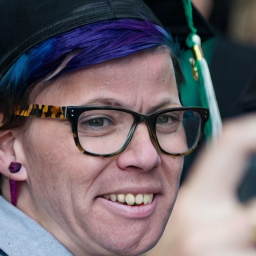} & \includegraphics[width=\qualseltile]{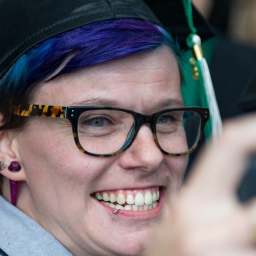} \\

      \qualsellabel{Source glasses}
  & \includegraphics[width=\qualseltile]{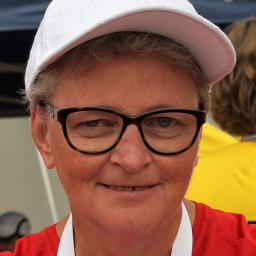} & \includegraphics[width=\qualseltile]{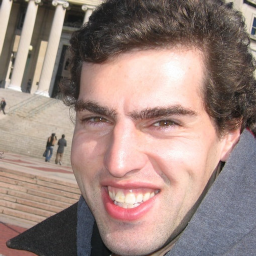}
  & \includegraphics[width=\qualseltile]{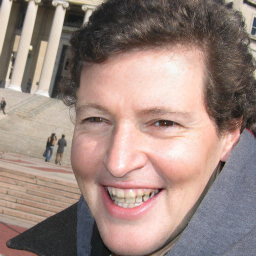} & \includegraphics[width=\qualseltile]{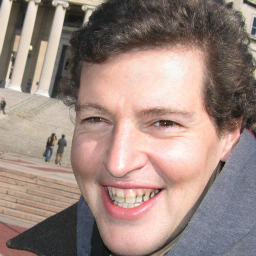}
  & \includegraphics[width=\qualseltile]{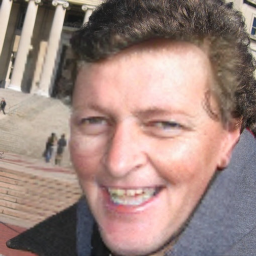} & \includegraphics[width=\qualseltile]{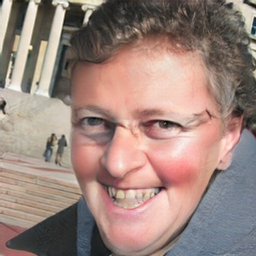}
  & \includegraphics[width=\qualseltile]{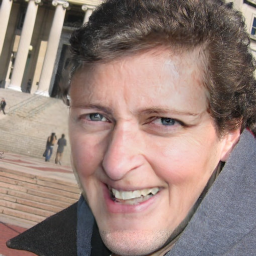} & \includegraphics[width=\qualseltile]{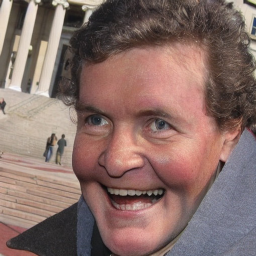} & \includegraphics[width=\qualseltile]{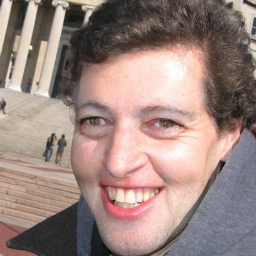} \\

  \end{tabular}
  \caption{\textbf{Qualitative comparison on selected FFHQ source--target
  pairs.} The first two columns give the source and the target. The remaining columns
show the swap produced by each method. Every row probes a different difficulty.
\emph{Gender} transfers an identity onto a target of another perceived gender.
\emph{Skin tone} presents a source--target pair with markedly different skin
tones. \emph{Piercing} tests piercing transfer from the target. \emph{Source glasses} tests
that a method does not transfer glasses from the source.
  }
  \label{fig:qualitative-selected-other}

  \vspace{1.5em}
  \begin{tabular}{@{}l cc @{\hspace{6pt}} *{6}{c}@{}}
    & Source & Target
    & MegaFS
    & E4S
    & HiFiFace & \shortstack{Face-\\Dancer} & \shortstack{Face-\\Adapter} & DiffSwap \\
      \qualsellabel{Naturalness}
  & \includegraphics[width=\qualseltile]{figures/ffhq_selected_examples/complex_source.pdf} & \includegraphics[width=\qualseltile]{figures/ffhq_selected_examples/complex_target.pdf}
  & \includegraphics[width=\qualseltile]{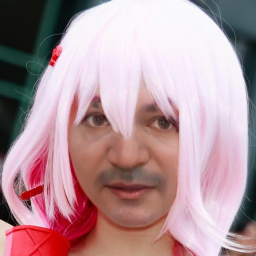} & \includegraphics[width=\qualseltile]{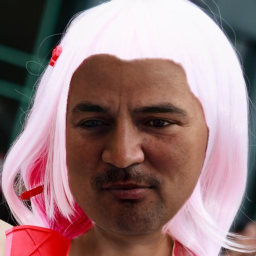}
  & \includegraphics[width=\qualseltile]{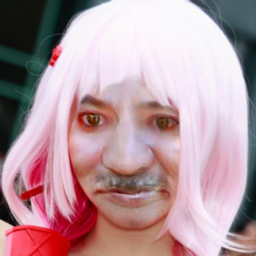} & \includegraphics[width=\qualseltile]{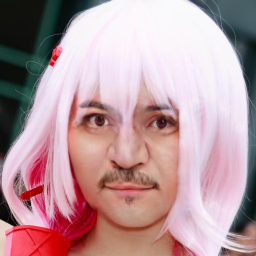}
  & \includegraphics[width=\qualseltile]{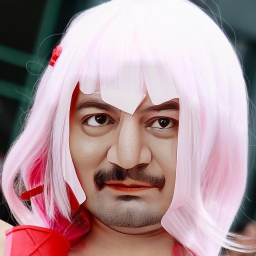} & \includegraphics[width=\qualseltile]{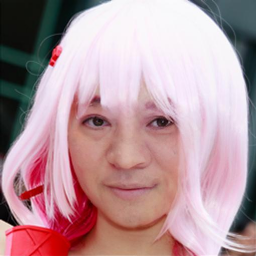} \\

      \qualsellabel{Age}
  & \includegraphics[width=\qualseltile]{figures/ffhq_selected_examples/age_2_source.pdf} & \includegraphics[width=\qualseltile]{figures/ffhq_selected_examples/age_2_target.pdf}
  & \includegraphics[width=\qualseltile]{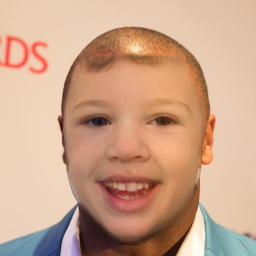} & \includegraphics[width=\qualseltile]{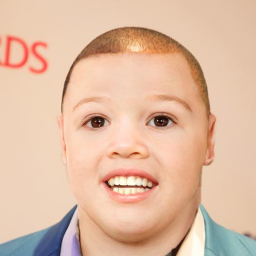}
  & \includegraphics[width=\qualseltile]{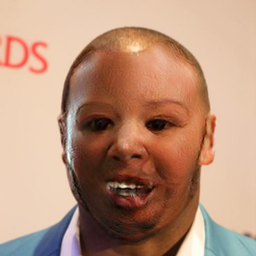} & \includegraphics[width=\qualseltile]{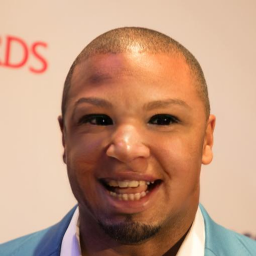}
  & \includegraphics[width=\qualseltile]{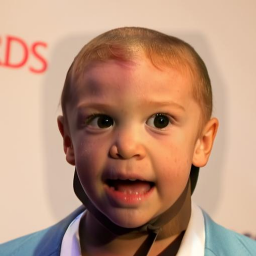} & \includegraphics[width=\qualseltile]{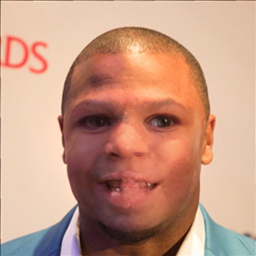} \\

      \qualsellabel{Glasses}
  & \includegraphics[width=\qualseltile]{figures/ffhq_selected_examples/glasses_4_source.pdf} & \includegraphics[width=\qualseltile]{figures/ffhq_selected_examples/glasses_4_target.pdf}
  & \includegraphics[width=\qualseltile]{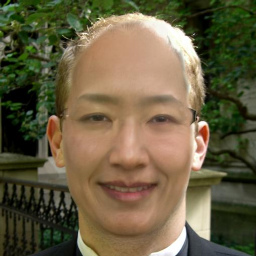} & \includegraphics[width=\qualseltile]{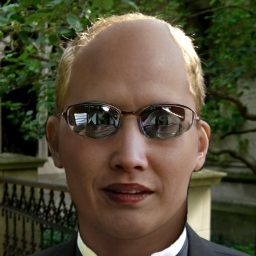}
  & \includegraphics[width=\qualseltile]{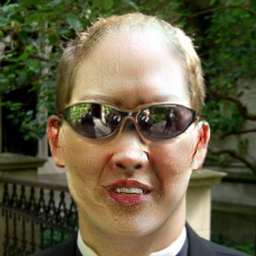} & \includegraphics[width=\qualseltile]{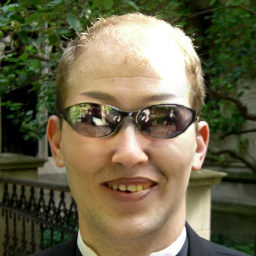}
  & \includegraphics[width=\qualseltile]{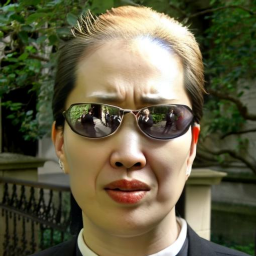} & \includegraphics[width=\qualseltile]{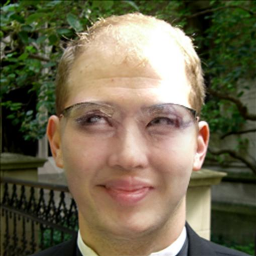} \\

      \qualsellabel{Tattoo}
  & \includegraphics[width=\qualseltile]{figures/ffhq_selected_examples/tattoo_source.pdf} & \includegraphics[width=\qualseltile]{figures/ffhq_selected_examples/tattoo_target.pdf}
  & \includegraphics[width=\qualseltile]{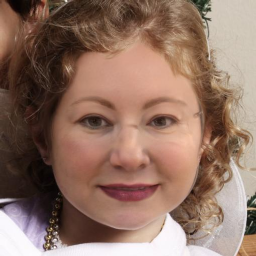} & \includegraphics[width=\qualseltile]{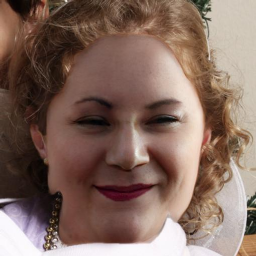}
  & \includegraphics[width=\qualseltile]{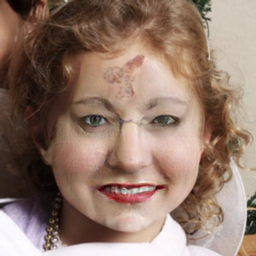} & \includegraphics[width=\qualseltile]{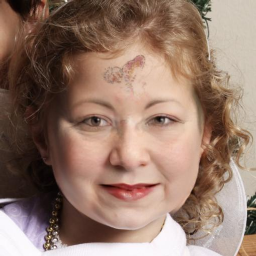}
  & \includegraphics[width=\qualseltile]{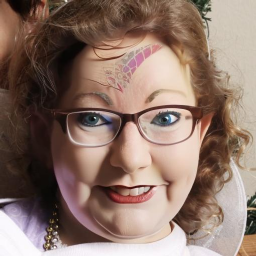} & \includegraphics[width=\qualseltile]{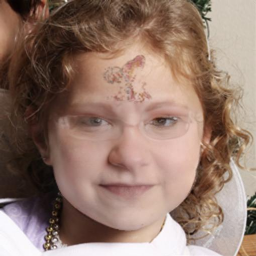} \\

      \qualsellabel{Pose}
  & \includegraphics[width=\qualseltile]{figures/ffhq_selected_examples/pose_1_source.pdf} & \includegraphics[width=\qualseltile]{figures/ffhq_selected_examples/pose_1_target.pdf}
  & \includegraphics[width=\qualseltile]{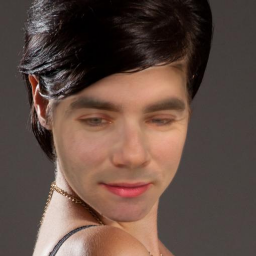} & \includegraphics[width=\qualseltile]{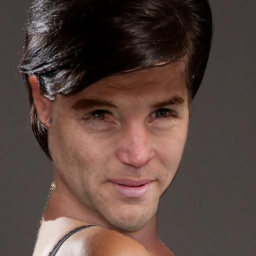}
  & \includegraphics[width=\qualseltile]{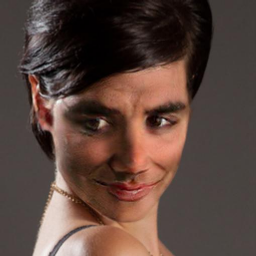} & \includegraphics[width=\qualseltile]{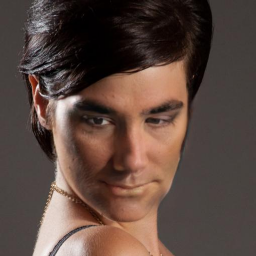}
  & \includegraphics[width=\qualseltile]{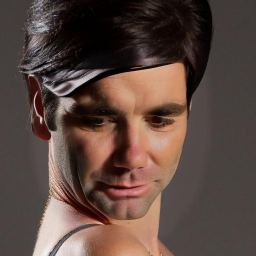} & \includegraphics[width=\qualseltile]{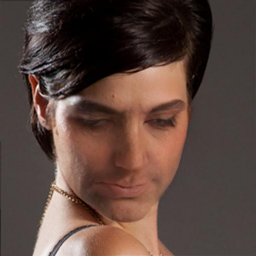} \\

      \qualsellabel{Expression}
  & \includegraphics[width=\qualseltile]{figures/ffhq_selected_examples/pose_2_source.pdf} & \includegraphics[width=\qualseltile]{figures/ffhq_selected_examples/pose_2_target.pdf}
  & \includegraphics[width=\qualseltile]{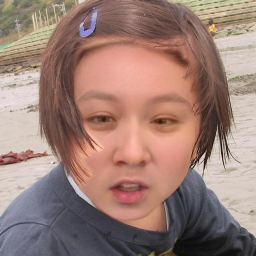} & \includegraphics[width=\qualseltile]{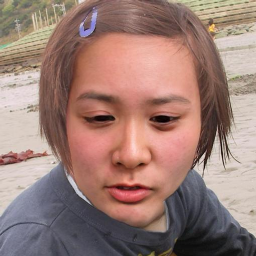}
  & \includegraphics[width=\qualseltile]{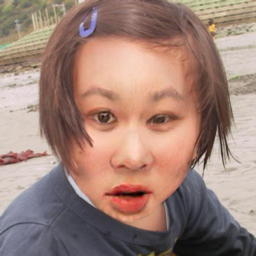} & \includegraphics[width=\qualseltile]{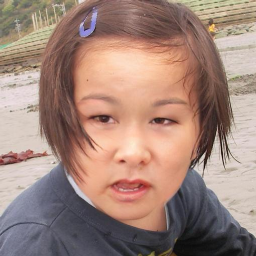}
  & \includegraphics[width=\qualseltile]{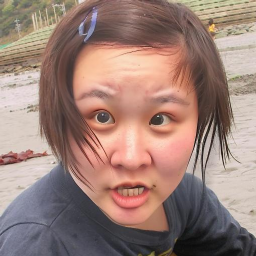} & \includegraphics[width=\qualseltile]{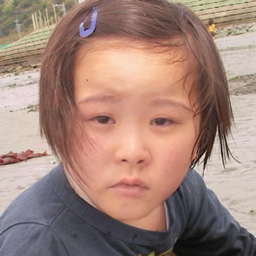} \\

  \end{tabular}
  \caption{\textbf{Qualitative comparison on selected FFHQ source--target
  pairs for the methods omitted from Figure~\ref{fig:qualitative-selected-compact}.} The first two columns give the source and the target. The remaining columns show the swap produced by each method. Every row probes a different difficulty. \textit{Naturalness} transfers an identity onto a target with unusual hair, and \textit{Age} pairs a source and a target of very different ages. The last four rows test how well the target's attributes survive the swap: glasses, tattoo, head pose, and facial expression.
  }
  \label{fig:qualitative-selected-omitted}
\end{figure*}
\endgroup

%% file: figures/figure_celebahq_operating_points.tex
\begin{figure*}[t]
  \centering
  \includegraphics[width=\textwidth]{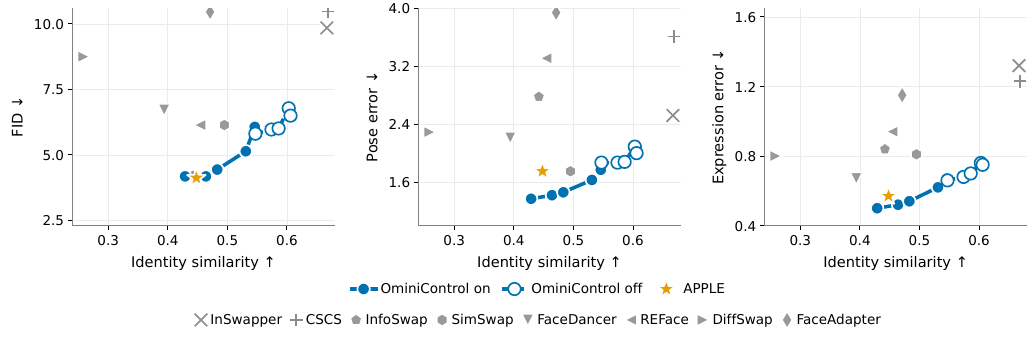}
  \caption{\textbf{Operating points of \MethodName{} on 1,000 pairs from CelebA-HQ}. Lower is better on every vertical axis. Students 7, 3, 2, and 1 and the Teacher with the OminiControl branch on and off, PuLID identity weight $1$, and the baselines of Table~\ref{tab:celebahq}. MegaFS, E4S, and HiFiFace lie outside the FID range.}
  \label{fig:celebahq-operating-points}
\end{figure*}

%% file: figures/table_celebahq.tex
\begin{table}[t]
  \centering
  \small
  \setlength{\tabcolsep}{3.5pt}
  \caption{
    \textbf{Results on 1,000 pairs from CelebA-HQ}~\citep{karras2018progressive}, the test pairs of
    REFace~\citep{baliah2025realistic}. Identity similarity and retrieval
    are measured with \textit{WebFace--R50}. FID uses CLIP features of all
    30,000 CelebA-HQ images. \textbf{Bold} shows the best value in each
    column and the second and third best are \underline{underlined}.
  }
  \label{tab:celebahq}
  \begin{tabular}{lrrcrr}
    \toprule
    Method
      & FID $\downarrow$
      & ID Sim. $\uparrow$
      & \makecell{Retrieval $\uparrow$ \\ Top-1/5}
      & Pose $\downarrow$
      & Expr. $\downarrow$ \\
    \midrule
    \multicolumn{6}{l}{\textit{One-step methods}} \\
    MegaFS
      & 22.06 & 0.37 & 78.3/82.4 & 5.79 & 1.19 \\
    E4S
      & 14.48 & 0.41 & 84.1/91.9 & 4.10 & 1.17 \\
    HiFiFace
      & 11.08 & 0.50 & 96.5/98.5 & 2.83 & 1.14 \\
    CSCS
      & 10.48 & \textbf{0.67}
      & \textbf{99.5}/\textbf{99.9} & 3.61 & 1.23 \\
    InSwapper$^{\dagger}$
      & 9.85 & $\,^\dagger$\textbf{0.67}
      & 96.1/96.4 & 2.52 & 1.32 \\
    FaceDancer
      & 6.71 & 0.39 & 86.9/95.8 & 2.21 & 0.67 \\
    SimSwap
      & 6.14 & 0.49 & 97.3/99.2 & \underline{1.75} & 0.81 \\
    InfoSwap
      & \underline{4.21} & 0.44 & 92.4/96.8 & 2.78 & 0.84 \\
    \midrule
    \multicolumn{6}{l}{\textit{Multi-step methods}} \\
    FaceAdapter
      & 10.46 & 0.47 & 95.9/98.7 & 3.94 & 1.15 \\
    DiffSwap$^{\S}$
      & 8.75 & 0.26 & 44.1/67.5 & 2.29 & 0.80 \\
    REFace
      & 6.14 & 0.45 & 96.6/98.4 & 3.31 & 0.94 \\
    APPLE
      & \textbf{4.12} & 0.45 & 94.8/98.6
      & \underline{1.75} & \underline{0.57} \\
    \midrule
    \multicolumn{6}{l}{\textit{\textbf{Ours \MethodName}, one-step method}} \\
    Teacher$^{\ddagger}$
      & 6.78 & \underline{0.60} & \underline{99.1}/\underline{99.8} & 2.09 & 0.76 \\
    Teacher
      & 6.07 & \underline{0.55} & \underline{98.6}/99.2 & 1.77 & 0.67 \\
    \Trip{1}
      & 5.14 & 0.53
      & 97.6/\underline{99.3} & \underline{1.63} & \underline{0.62} \\
    \Trip{3}
      & \underline{4.18} & 0.46 & 95.4/98.2
      & \textbf{1.42} & \textbf{0.52} \\
    \bottomrule
  \end{tabular}

  \vspace{2pt}
  {\footnotesize
    $^{\dagger}$ InSwapper generates face swaps using the same identity
    model employed here for evaluation. $^{\ddagger}$ OminiControl branch off at inference, as in Table~\ref{tab:ffhq-bench1k-WebFaceR50}. $^{\S}$ DiffSwap detected no face on 5 pairs and is scored on the remaining 995.
  }
\end{table}

%% file: figures/table_ffpp.tex
\begin{table}[t]
  \centering
  \small
  \setlength{\tabcolsep}{3.5pt}
  \caption{
    \textbf{Results on 1,000 pairs from FaceForensics++}~\citep{rossler2019faceforensics}, one frame per video
    (\suppsec{app:celebahq}). Identity similarity and retrieval are measured with \textit{WebFace--R50}.
    FID uses CLIP features of $9{,}999$ FF++ face crops. \textbf{Bold} shows the best value in each
    column and the second and third best are \underline{underlined}.
  }
  \label{tab:ffpp}
  \begin{tabular}{lrrcrr}
    \toprule
    Method
      & FID $\downarrow$
      & ID Sim. $\uparrow$
      & \makecell{Retrieval $\uparrow$ \\ Top-1/5}
      & Pose $\downarrow$
      & Expr. $\downarrow$ \\
    \midrule
    \multicolumn{6}{l}{\textit{One-step methods}} \\
    E4S & 14.82 & 0.49 & 95.5/98.7 & 3.15 & 1.34 \\
    MegaFS & 11.40 & 0.44 & 85.3/87.3 & 4.83 & 1.27 \\
    InfoSwap & 7.62 & 0.56 & 99.4/99.7 & 2.64 & 1.07 \\
    HiFiFace & 7.05 & 0.56 & 98.8/\textbf{100} & 2.58 & 0.98 \\
    CSCS & 6.84 & \underline{0.70} & 98.4/99.4 & 2.67 & 1.16 \\
    FaceDancer & 2.86 & 0.44 & 93.5/98.0 & 1.93 & 0.67 \\
    InSwapper$^{\dagger}$ & 2.69 & $\,^\dagger$\textbf{0.74} & 96.0/96.1 & 2.18 & 0.91 \\
    SimSwap & \underline{2.13} & 0.58 & 99.5/\textbf{100} & \underline{1.45} & \underline{0.66} \\
    \midrule
    \multicolumn{6}{l}{\textit{Multi-step methods}} \\
    REFace & 7.67 & 0.55 & 99.2/\textbf{100} & 2.96 & 1.19 \\
    FaceAdapter & 3.66 & 0.57 & 99.3/99.9 & 3.23 & 1.14 \\
    DiffSwap & 3.46 & 0.31 & 53.8/81.3 & 1.99 & 0.73 \\
    APPLE & 2.19 & 0.53 & 99.4/99.9 & 1.87 & 0.74 \\
    \midrule
    \multicolumn{6}{l}{\textit{\textbf{Ours \MethodName}, one-step method}} \\
    Teacher$^{\ddagger}$ & 3.32 & \underline{0.66} & \textbf{99.8}/99.9 & 1.83 & 0.76 \\
    Teacher & 2.66 & 0.62 & \underline{99.7}/\textbf{100} & 1.63 & 0.67 \\
    \Trip{1} & \underline{2.06} & 0.61 & \textbf{99.8}/99.9 & \underline{1.56} & \underline{0.64} \\
    \Trip{3} & \textbf{1.77} & 0.53 & 97.9/99.8 & \textbf{1.33} & \textbf{0.53} \\
    \bottomrule
  \end{tabular}

  \vspace{2pt}
  {\footnotesize
    $^{\dagger}$ InSwapper generates face swaps using the same identity
    model employed here for evaluation. $^{\ddagger}$ OminiControl branch off at inference, as in Table~\ref{tab:ffhq-bench1k-WebFaceR50}.
  }
\end{table}

%% file: figures/table_sigma_ablation.tex
\begin{table}[t]
  \centering
  \small
  \setlength{\tabcolsep}{4pt}
  \caption{\textbf{Generator noise level $\sigma_{\mathrm{gen}}$} on 1,000 pairs from FFHQ. Teachers trained on the 0.75\,/\,0.25 mixture of Table~\ref{tab:ablation} without discriminator latent noise, each scored at its own training noise level with the OminiControl branch on. Rules separate low, intermediate, and large values. At $\sigma_{\mathrm{gen}}=1$ identity transfer fails, so its realism and attribute scores do not describe a swap.}
  \label{tab:sigma-ablation}
  \begin{tabular}{lcccc}
  \toprule
  $\sigma_{\mathrm{gen}}$ & FID $\downarrow$ & ID Sim. $\uparrow$ & Pose $\downarrow$ & Expr. $\downarrow$ \\
  \midrule
  0.01 & 2.47 & 0.32 & 1.08 & 0.41 \\
  0.05 & 3.11 & 0.36 & 1.26 & 0.52 \\
  0.10 & 2.88 & 0.34 & 1.18 & 0.48 \\
  \midrule
  0.15 & 3.96 & 0.46 & 1.66 & 0.69 \\
  0.20 & 3.53 & 0.43 & 1.55 & 0.61 \\
  0.50 & 5.02 & 0.49 & 1.92 & 0.79 \\
  \midrule
  1.00 & 2.08 & 0.03 & 0.28 & 0.18 \\
  \bottomrule
  \end{tabular}
\end{table}

%% file: figures/table_dataset_ablation.tex
\begin{table}[t]
  \centering
  \small
  \setlength{\tabcolsep}{4pt}
  \caption{\textbf{Dataset filtering} on 1,000 pairs from FFHQ. Teachers trained on the unfiltered data and with the thresholds $\tau_{\mathrm{AES}}$ and $\tau_{\mathrm{ID}}$ of Algorithm~\ref{alg:dataset-filtering}, scored as in Table~\ref{tab:ffhq-bench1k-WebFaceR50}. The $(5.1, 0.6)$ row is the Teacher of Table~\ref{tab:ffhq-bench1k-WebFaceR50}.}
  \label{tab:dataset-ablation}
  \begin{tabular}{lcccc}
  \toprule
  $(\tau_{\mathrm{AES}}, \tau_{\mathrm{ID}})$ & FID $\downarrow$ & ID Sim. $\uparrow$ & Pose $\downarrow$ & Expr. $\downarrow$ \\
  \midrule
  No filtering & 4.28 & 0.49 & 1.70 & 0.70 \\
  $(5.1, 0.6)$, Teacher & 4.07 & 0.48 & 1.71 & 0.68 \\
  \midrule
  $(5.3, 0.6)$ & 4.33 & 0.47 & 1.72 & 0.74 \\
  $(5.5, 0.6)$ & 4.20 & 0.49 & 1.79 & 0.73 \\
  $(5.7, 0.6)$ & 4.03 & 0.47 & 1.73 & 0.68 \\
  \midrule
  $(5.1, 0.5)$ & 4.19 & 0.48 & 1.71 & 0.71 \\
  $(5.1, 0.7)$ & 3.62 & 0.44 & 1.53 & 0.64 \\
  $(5.1, 0.8)$ & 3.98 & 0.46 & 1.72 & 0.71 \\
  \bottomrule
  \end{tabular}
\end{table}

%% file: figures/algorithm_dataset_filtering.tex
\begin{algorithm}[t]
\caption{Dataset filtering}
\label{alg:dataset-filtering}
\small
\begin{algorithmic}[1]
\Require Identity groups $\mathcal{G}$; AES scores $s(x)$; identity embeddings $e_x$; thresholds $\tau_{\mathrm{AES}}$, $\tau_{\mathrm{ID}}$; minimum distance $\delta=0.05$
\Ensure Filtered dataset $\mathcal{D}_{\mathrm{out}}$
\State Define $d(x,y)=1-\cos(e_x,e_y)$
\State $\mathcal{D}_{\mathrm{out}}\gets\varnothing$
\ForAll{$G\in\mathcal{G}$}
    \State $C\gets\{x\in G:s(x)\geq\tau_{\mathrm{AES}}\}$
    \If{$|C|<2$}
        \State \textbf{continue}
    \EndIf
    \State $a\gets\arg\min_{x\in C}\sum_{y\in C}d(x,y)$ \Comment{Anchor}
    \State $S\gets\{a\}$
    \ForAll{$x\in C\setminus\{a\}$ in increasing order of $d(x,a)$}
        \If{$d(x,a)\leq1-\tau_{\mathrm{ID}}$ and $\min_{y\in S}d(x,y)\geq\delta$}
            \State $S\gets S\cup\{x\}$
        \EndIf
    \EndFor
    \If{$|S|\geq2$}
        \State $\mathcal{D}_{\mathrm{out}}\gets\mathcal{D}_{\mathrm{out}}\cup S$
    \EndIf
\EndFor
\State \Return $\mathcal{D}_{\mathrm{out}}$
\end{algorithmic}
\end{algorithm}

%% file: figures/figure_sweeps_combined.tex
\begin{figure*}[p]
  \centering
  \includegraphics[width=0.98\textwidth]{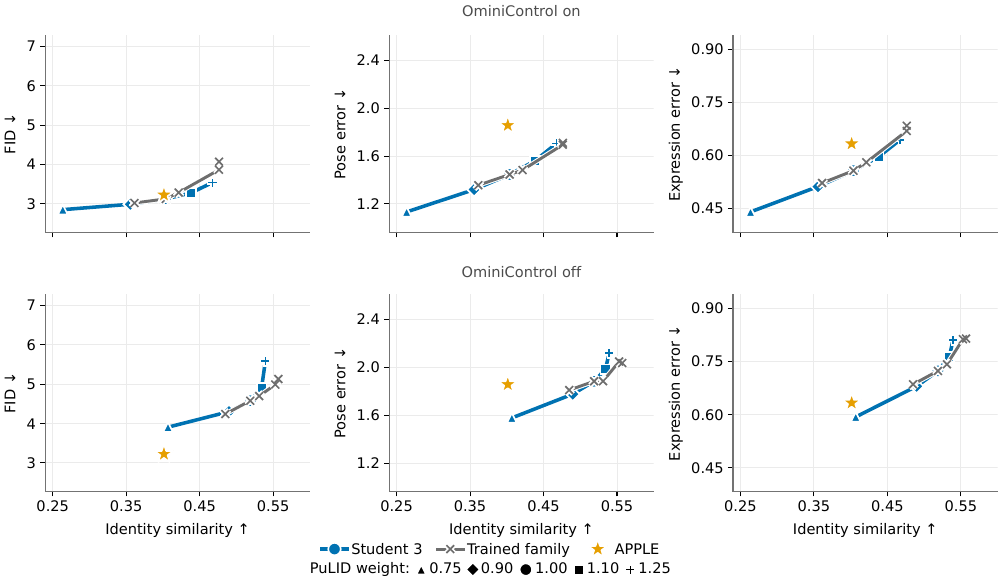}
  \caption{\textbf{PuLID identity weight sweep of \Trip{3}} (blue, weights $0.75$ to $1.25$) against Students 7, 3, 2, 1 and the Teacher at PuLID weight $1$ (grey) with the OminiControl branch on (top) and off (bottom), on 1,000 pairs from FFHQ.}
  \label{fig:pulid-sweep-full}
  \vspace{2pt}
  \includegraphics[width=0.98\textwidth]{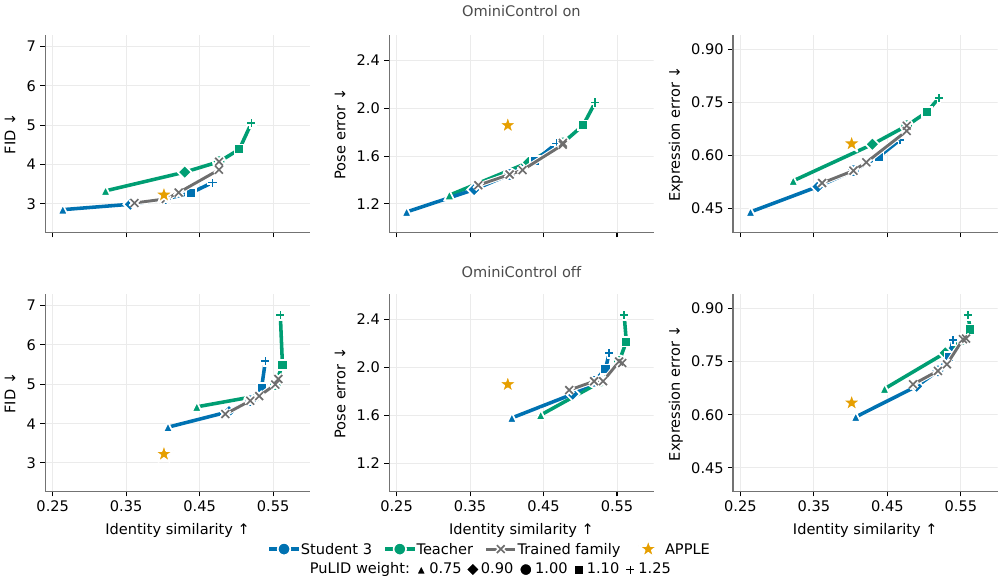}
  \caption{\textbf{Student training moves the curve.} PuLID identity weight sweeps of the Teacher (green) and \Trip{3} (blue) with the OminiControl branch on (top) and off (bottom), on 1,000 pairs from FFHQ. Grey: Students 7, 3, 2, 1 and the Teacher at PuLID weight $1$.}
  \label{fig:teacher-vs-student}
\end{figure*}

%% file: figures/figure_pulid_sweep_grid.tex
\begingroup
\ifdefined\pgridtile\else\newlength{\pgridtile}\fi
\setlength{\pgridtile}{0.124\textwidth}
\newcommand{\pgridcat}[1]{%
  \raisebox{0pt}[0pt][0pt]{\rotatebox[origin=c]{90}{\makebox[2\pgridtile][c]{\small #1}}}%
}
\newcommand{\pgridrow}[1]{%
  \raisebox{0.5\pgridtile}[0pt][0pt]{\rotatebox[origin=c]{90}{\makebox[\pgridtile][c]{\small #1}}}%
}

\begin{figure*}[tbp]
  \centering
  \small
  \setlength{\tabcolsep}{0.5pt}
  \renewcommand{\arraystretch}{0.95}
  \begin{tabular}{@{}l l c @{\hspace{8pt}} l *{5}{c}@{}}
    & & & & \multicolumn{5}{c}{PuLID weight} \\
    \cmidrule(lr){5-9}
    & & & & 0.75 & 0.90 & 1.00 & 1.10 & 1.25 \\
      \pgridcat{Age} & \pgridrow{Source} & \includegraphics[width=\pgridtile]{figures/ffhq_selected_examples/age_2_source.pdf} & \pgridrow{on}
  & \includegraphics[width=\pgridtile]{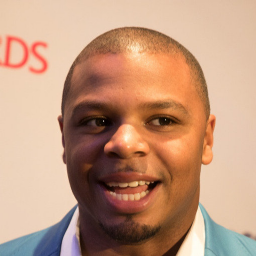} & \includegraphics[width=\pgridtile]{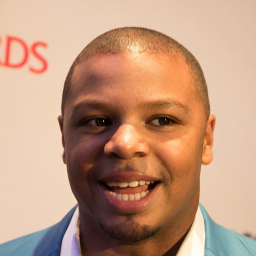} & \includegraphics[width=\pgridtile]{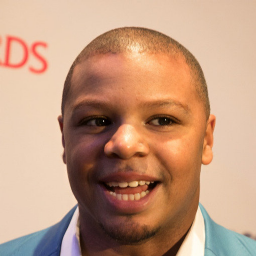} & \includegraphics[width=\pgridtile]{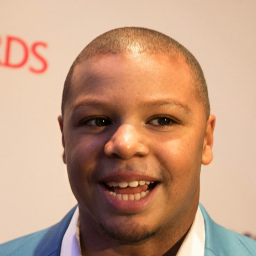} & \includegraphics[width=\pgridtile]{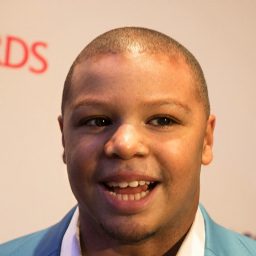} \\
  & \pgridrow{Target} & \includegraphics[width=\pgridtile]{figures/ffhq_selected_examples/age_2_target.pdf} & \pgridrow{off}
  & \includegraphics[width=\pgridtile]{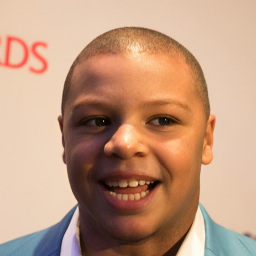} & \includegraphics[width=\pgridtile]{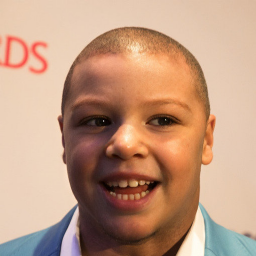} & \includegraphics[width=\pgridtile]{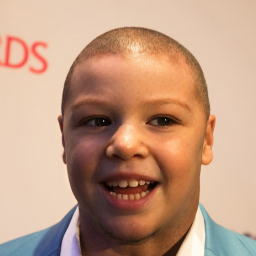} & \includegraphics[width=\pgridtile]{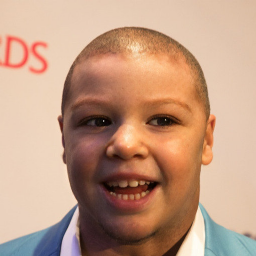} & \includegraphics[width=\pgridtile]{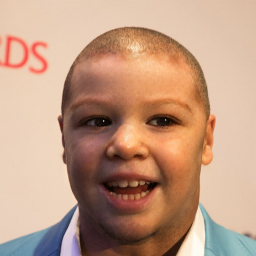} \\[4pt]

      \pgridcat{Make-up} & \pgridrow{Source} & \includegraphics[width=\pgridtile]{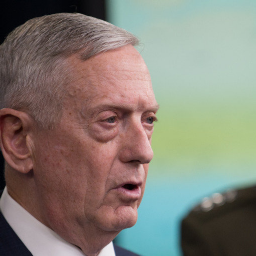} & \pgridrow{on}
  & \includegraphics[width=\pgridtile]{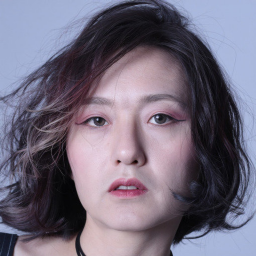} & \includegraphics[width=\pgridtile]{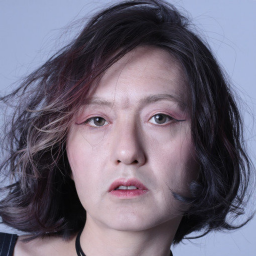} & \includegraphics[width=\pgridtile]{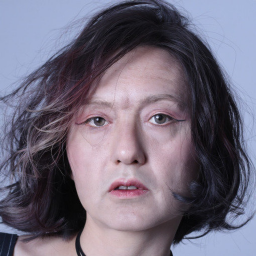} & \includegraphics[width=\pgridtile]{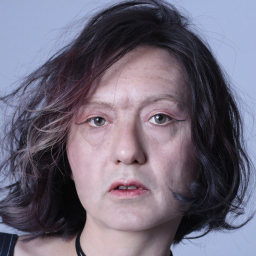} & \includegraphics[width=\pgridtile]{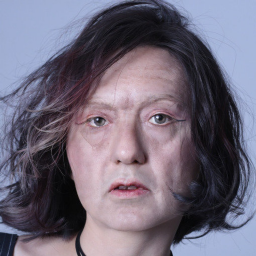} \\
  & \pgridrow{Target} & \includegraphics[width=\pgridtile]{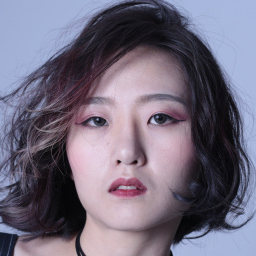} & \pgridrow{off}
  & \includegraphics[width=\pgridtile]{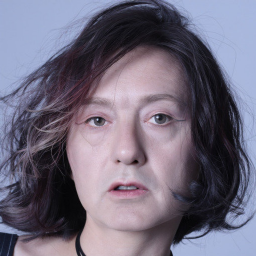} & \includegraphics[width=\pgridtile]{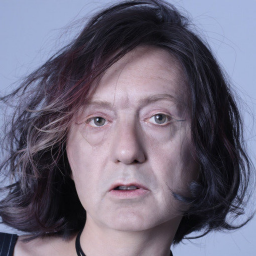} & \includegraphics[width=\pgridtile]{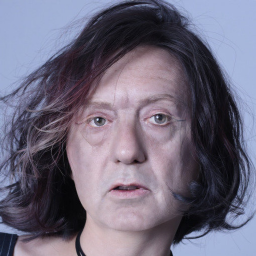} & \includegraphics[width=\pgridtile]{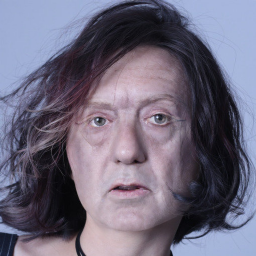} & \includegraphics[width=\pgridtile]{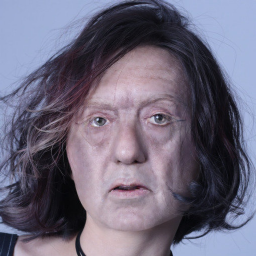} \\[4pt]

      \pgridcat{Tattoo} & \pgridrow{Source} & \includegraphics[width=\pgridtile]{figures/ffhq_selected_examples/tattoo_source.pdf} & \pgridrow{on}
  & \includegraphics[width=\pgridtile]{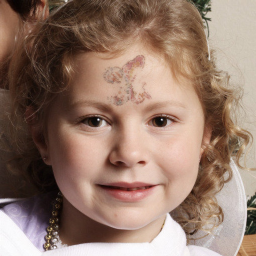} & \includegraphics[width=\pgridtile]{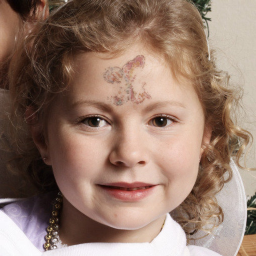} & \includegraphics[width=\pgridtile]{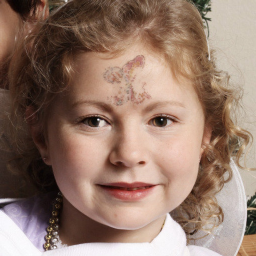} & \includegraphics[width=\pgridtile]{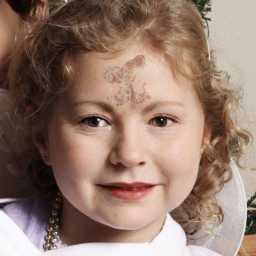} & \includegraphics[width=\pgridtile]{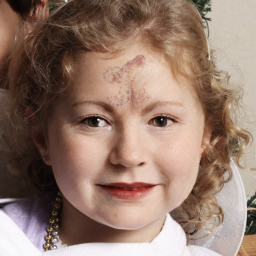} \\
  & \pgridrow{Target} & \includegraphics[width=\pgridtile]{figures/ffhq_selected_examples/tattoo_target.pdf} & \pgridrow{off}
  & \includegraphics[width=\pgridtile]{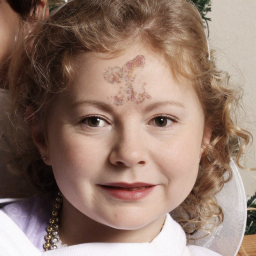} & \includegraphics[width=\pgridtile]{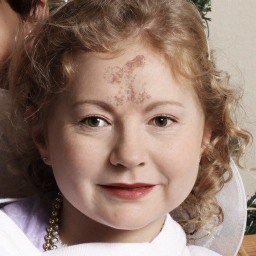} & \includegraphics[width=\pgridtile]{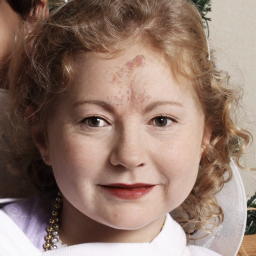} & \includegraphics[width=\pgridtile]{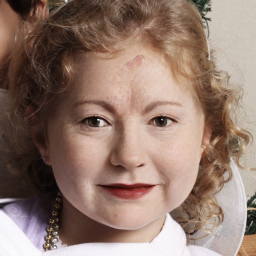} & \includegraphics[width=\pgridtile]{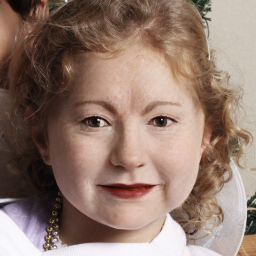} \\[4pt]

      \pgridcat{Expression} & \pgridrow{Source} & \includegraphics[width=\pgridtile]{figures/ffhq_selected_examples/pose_2_source.pdf} & \pgridrow{on}
  & \includegraphics[width=\pgridtile]{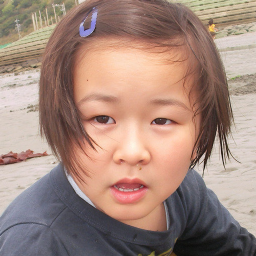} & \includegraphics[width=\pgridtile]{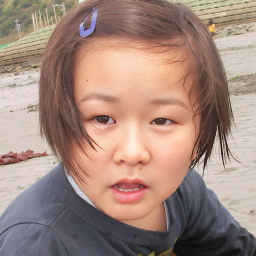} & \includegraphics[width=\pgridtile]{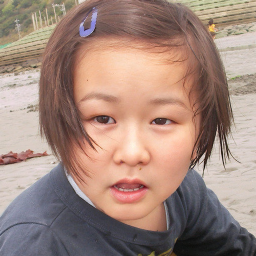} & \includegraphics[width=\pgridtile]{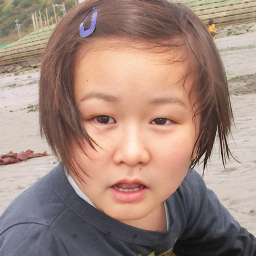} & \includegraphics[width=\pgridtile]{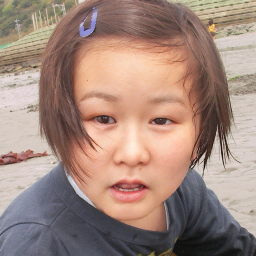} \\
  & \pgridrow{Target} & \includegraphics[width=\pgridtile]{figures/ffhq_selected_examples/pose_2_target.pdf} & \pgridrow{off}
  & \includegraphics[width=\pgridtile]{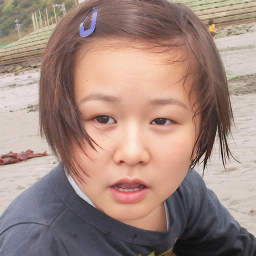} & \includegraphics[width=\pgridtile]{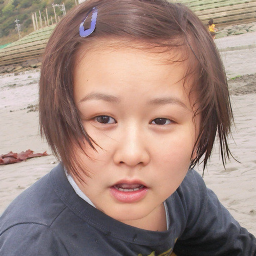} & \includegraphics[width=\pgridtile]{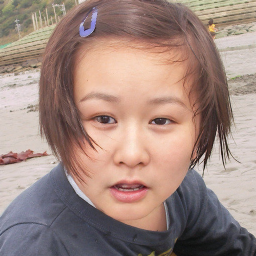} & \includegraphics[width=\pgridtile]{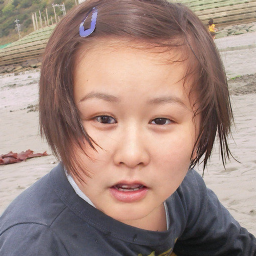} & \includegraphics[width=\pgridtile]{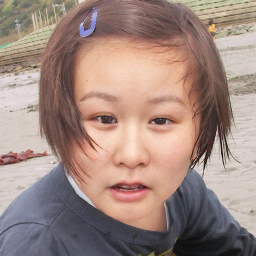} \\[4pt]

  \end{tabular}
  \caption{\textbf{PuLID weight sweep of \Trip{3}, OminiControl branch on and off.} For each FFHQ pair the left column shows the source (upper) and the target (lower). The upper row shows the swaps with the branch on, the lower row with it off, at PuLID weights $0.75$ to $1.25$. The on row at weight $1$ is the \Trip{3} of Table~\ref{tab:ffhq-bench1k-WebFaceR50}. Raising the weight or switching the branch off moves the swap toward the source identity at the cost of target attributes.}
  \label{fig:pulid-sweep-grid}
\end{figure*}
\endgroup

%% file: figures/figure_pulid_vs_family.tex
\begingroup
\ifdefined\pvftile\else\newlength{\pvftile}\fi
\setlength{\pvftile}{0.104\textwidth}
\newcommand{\pvfrow}[1]{%
  \raisebox{0.5\pvftile}[0pt][0pt]{\rotatebox[origin=c]{90}{\makebox[\pvftile][c]{\small #1}}}%
}
\newcommand{\pvfkind}[1]{\raisebox{0.5\pvftile}[0pt][0pt]{\small #1}}
\newcommand{\pvfhead}{%
    & & PuLID & 0.75 & 0.90 & 1.00 & 1.10 & 1.25 \\
    & & Trained & \Trip{7} & \Trip{3} & \Trip{2} & \Trip{1} & Teacher \\[1pt]
}
\newcommand{\pvfcaption}[1]{For each FFHQ pair the left column shows the source (upper) and the target (lower). The PuLID row shows \Trip{3} at PuLID weights $0.75$ to $1.25$, the Trained row the separately trained \Trip{7}, \Trip{3}, \Trip{2}, \Trip{1} and the Teacher at weight $1$, all with the OminiControl branch #1. Both rows are ordered by increasing identity similarity in Figure~\ref{fig:pulid-sweep-full}.}

\begin{figure*}[p]
  \centering
  \small
  \setlength{\tabcolsep}{0.5pt}
  \renewcommand{\arraystretch}{0.95}
  \begin{tabular}{@{}l c @{\hspace{8pt}} l *{5}{c}@{}}
    \pvfhead
      \pvfrow{Source} & \includegraphics[width=\pvftile]{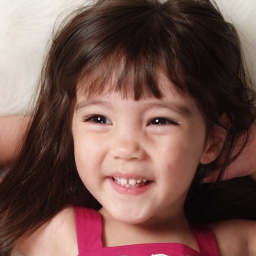} & \pvfkind{PuLID}
  & \includegraphics[width=\pvftile]{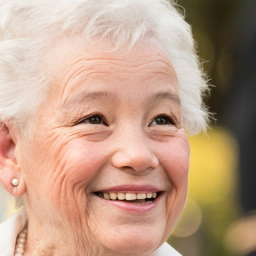} & \includegraphics[width=\pvftile]{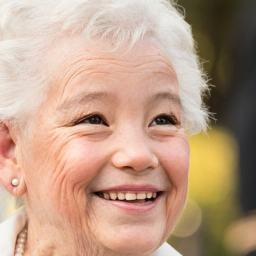} & \includegraphics[width=\pvftile]{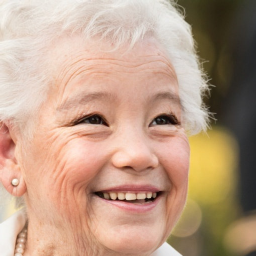} & \includegraphics[width=\pvftile]{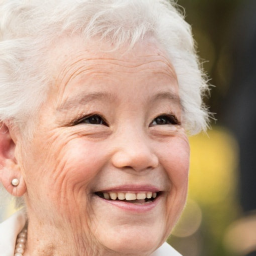} & \includegraphics[width=\pvftile]{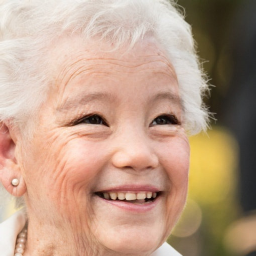} \\
  \pvfrow{Target} & \includegraphics[width=\pvftile]{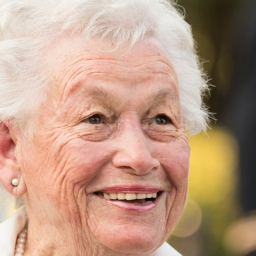} & \pvfkind{Trained}
  & \includegraphics[width=\pvftile]{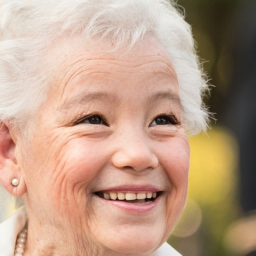} & \includegraphics[width=\pvftile]{figures/pulid_vs_family/p493_on_s3_w1p00.pdf} & \includegraphics[width=\pvftile]{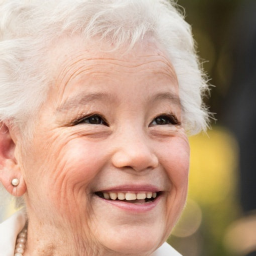} & \includegraphics[width=\pvftile]{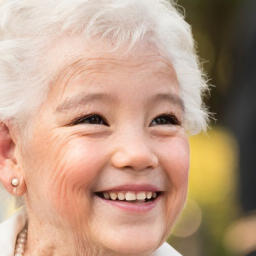} & \includegraphics[width=\pvftile]{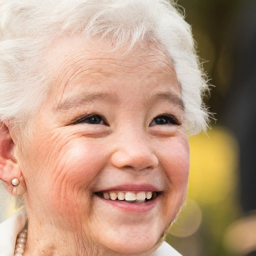} \\[3pt]
  \pvfrow{Source} & \includegraphics[width=\pvftile]{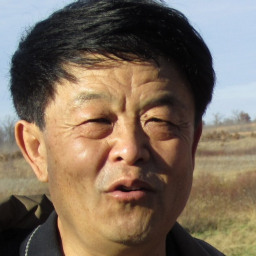} & \pvfkind{PuLID}
  & \includegraphics[width=\pvftile]{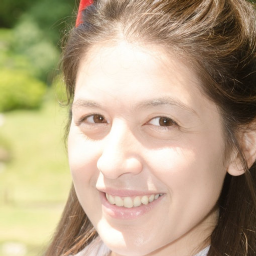} & \includegraphics[width=\pvftile]{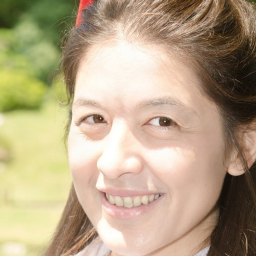} & \includegraphics[width=\pvftile]{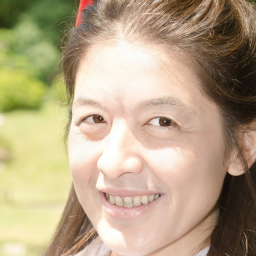} & \includegraphics[width=\pvftile]{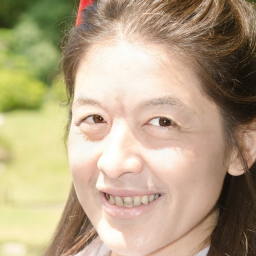} & \includegraphics[width=\pvftile]{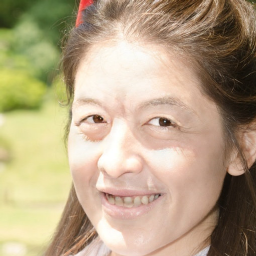} \\
  \pvfrow{Target} & \includegraphics[width=\pvftile]{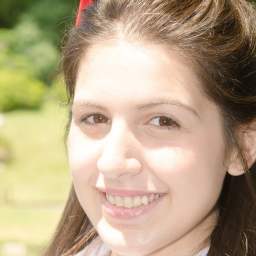} & \pvfkind{Trained}
  & \includegraphics[width=\pvftile]{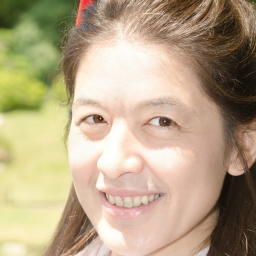} & \includegraphics[width=\pvftile]{figures/pulid_vs_family/p97_on_s3_w1p00.pdf} & \includegraphics[width=\pvftile]{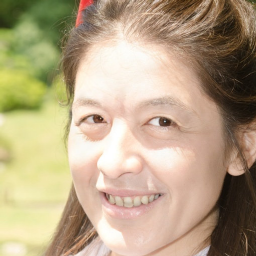} & \includegraphics[width=\pvftile]{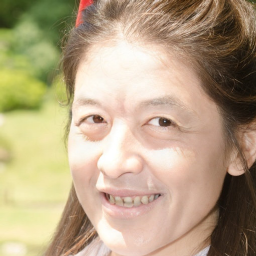} & \includegraphics[width=\pvftile]{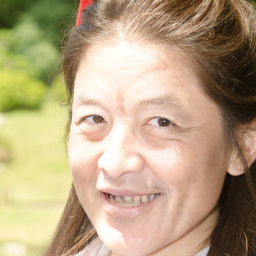} \\[3pt]
  \pvfrow{Source} & \includegraphics[width=\pvftile]{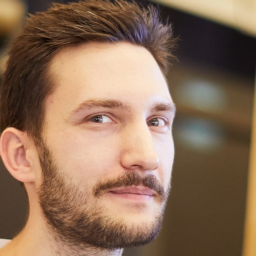} & \pvfkind{PuLID}
  & \includegraphics[width=\pvftile]{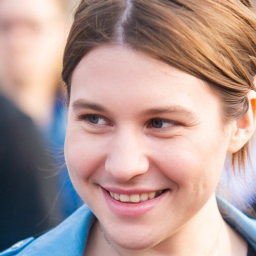} & \includegraphics[width=\pvftile]{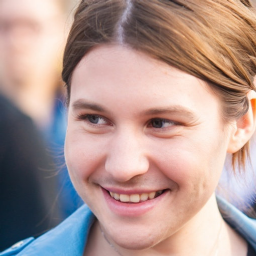} & \includegraphics[width=\pvftile]{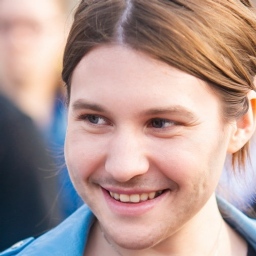} & \includegraphics[width=\pvftile]{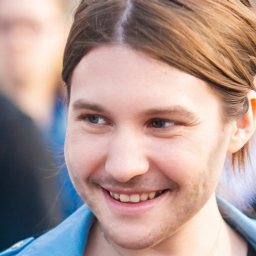} & \includegraphics[width=\pvftile]{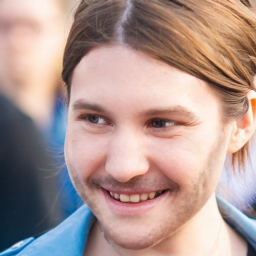} \\
  \pvfrow{Target} & \includegraphics[width=\pvftile]{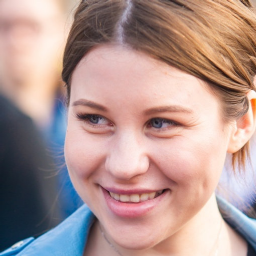} & \pvfkind{Trained}
  & \includegraphics[width=\pvftile]{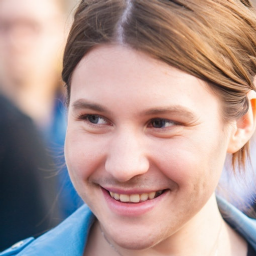} & \includegraphics[width=\pvftile]{figures/pulid_vs_family/p441_on_s3_w1p00.pdf} & \includegraphics[width=\pvftile]{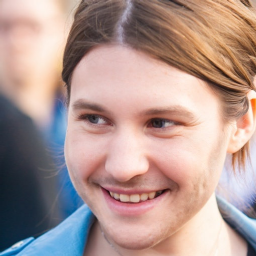} & \includegraphics[width=\pvftile]{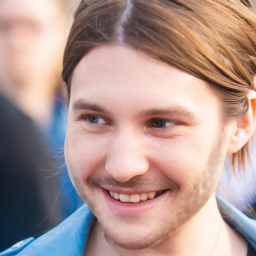} & \includegraphics[width=\pvftile]{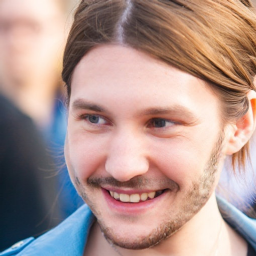} \\[3pt]
  \pvfrow{Source} & \includegraphics[width=\pvftile]{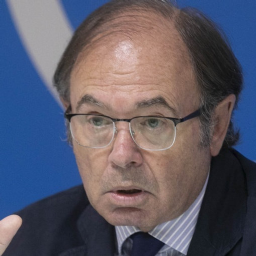} & \pvfkind{PuLID}
  & \includegraphics[width=\pvftile]{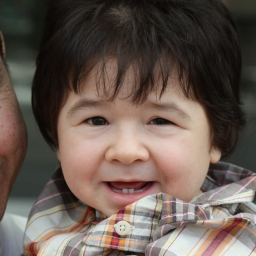} & \includegraphics[width=\pvftile]{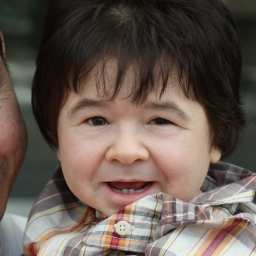} & \includegraphics[width=\pvftile]{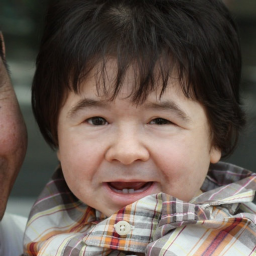} & \includegraphics[width=\pvftile]{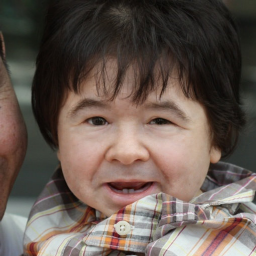} & \includegraphics[width=\pvftile]{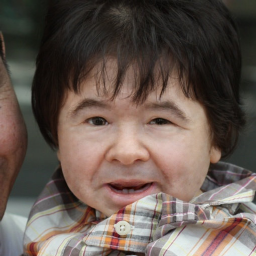} \\
  \pvfrow{Target} & \includegraphics[width=\pvftile]{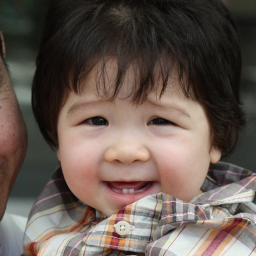} & \pvfkind{Trained}
  & \includegraphics[width=\pvftile]{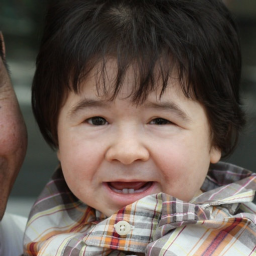} & \includegraphics[width=\pvftile]{figures/pulid_vs_family/p803_on_s3_w1p00.pdf} & \includegraphics[width=\pvftile]{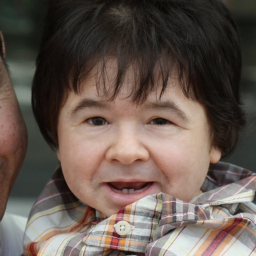} & \includegraphics[width=\pvftile]{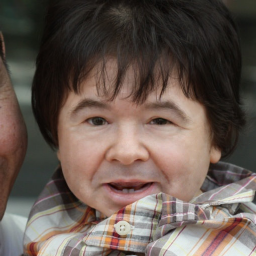} & \includegraphics[width=\pvftile]{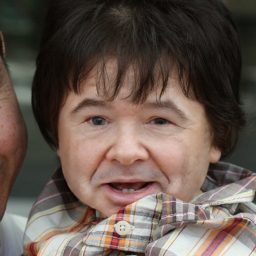} \\[3pt]
  \pvfrow{Source} & \includegraphics[width=\pvftile]{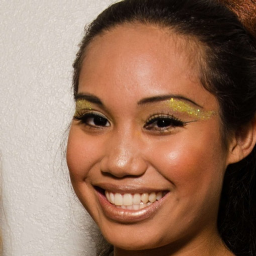} & \pvfkind{PuLID}
  & \includegraphics[width=\pvftile]{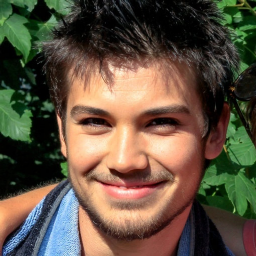} & \includegraphics[width=\pvftile]{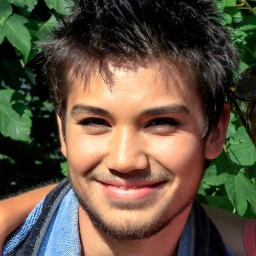} & \includegraphics[width=\pvftile]{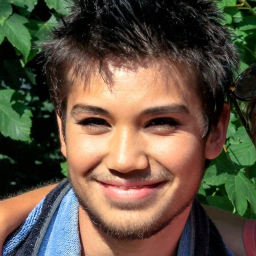} & \includegraphics[width=\pvftile]{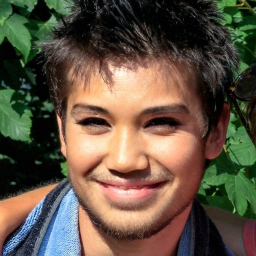} & \includegraphics[width=\pvftile]{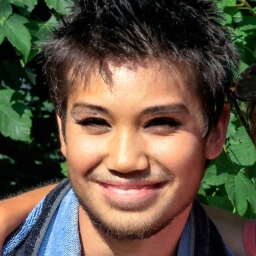} \\
  \pvfrow{Target} & \includegraphics[width=\pvftile]{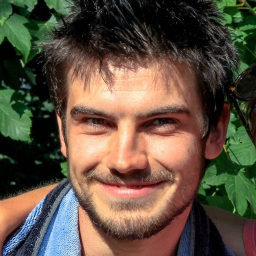} & \pvfkind{Trained}
  & \includegraphics[width=\pvftile]{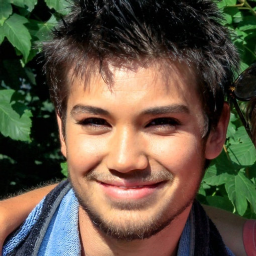} & \includegraphics[width=\pvftile]{figures/pulid_vs_family/p824_on_s3_w1p00.pdf} & \includegraphics[width=\pvftile]{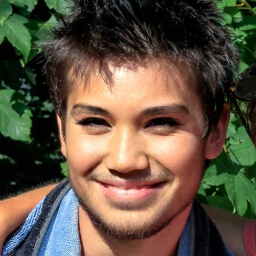} & \includegraphics[width=\pvftile]{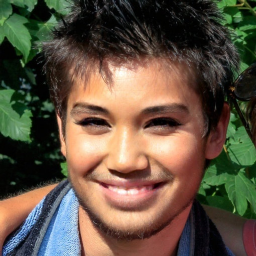} & \includegraphics[width=\pvftile]{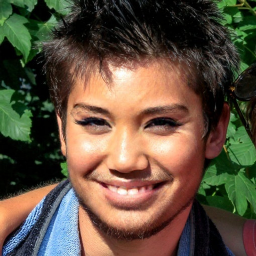} \\[3pt]

  \end{tabular}
  \caption{\textbf{PuLID weight sweep against the trained family, OminiControl branch on.} \pvfcaption{on}}
  \label{fig:pulid-vs-family-on}
\end{figure*}

\begin{figure*}[p]
  \centering
  \small
  \setlength{\tabcolsep}{0.5pt}
  \renewcommand{\arraystretch}{0.95}
  \begin{tabular}{@{}l c @{\hspace{8pt}} l *{5}{c}@{}}
    \pvfhead
      \pvfrow{Source} & \includegraphics[width=\pvftile]{figures/pulid_vs_family/p493_source.pdf} & \pvfkind{PuLID}
  & \includegraphics[width=\pvftile]{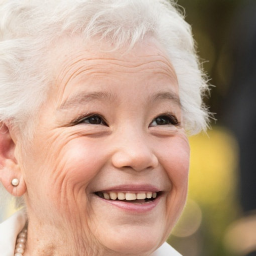} & \includegraphics[width=\pvftile]{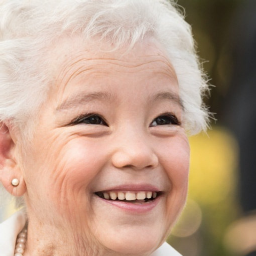} & \includegraphics[width=\pvftile]{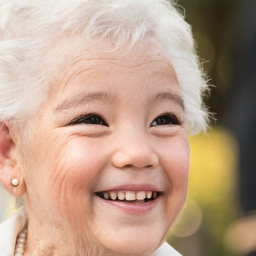} & \includegraphics[width=\pvftile]{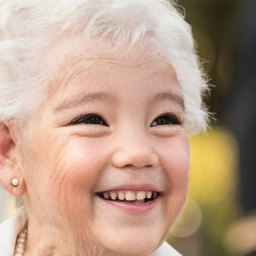} & \includegraphics[width=\pvftile]{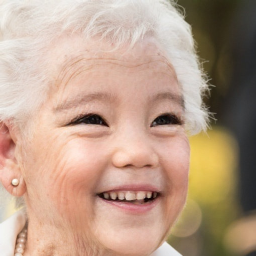} \\
  \pvfrow{Target} & \includegraphics[width=\pvftile]{figures/pulid_vs_family/p493_target.pdf} & \pvfkind{Trained}
  & \includegraphics[width=\pvftile]{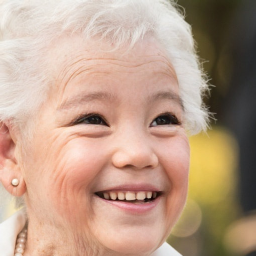} & \includegraphics[width=\pvftile]{figures/pulid_vs_family/p493_off_s3_w1p00.pdf} & \includegraphics[width=\pvftile]{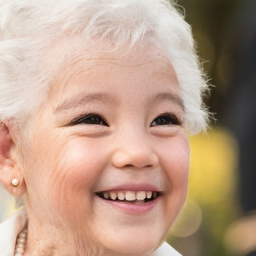} & \includegraphics[width=\pvftile]{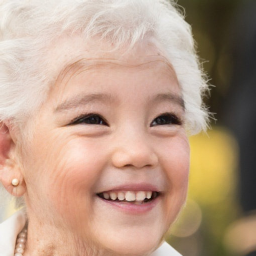} & \includegraphics[width=\pvftile]{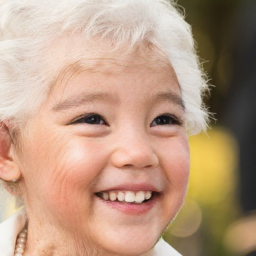} \\[3pt]
  \pvfrow{Source} & \includegraphics[width=\pvftile]{figures/pulid_vs_family/p97_source.pdf} & \pvfkind{PuLID}
  & \includegraphics[width=\pvftile]{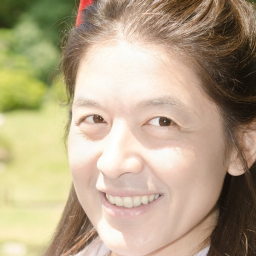} & \includegraphics[width=\pvftile]{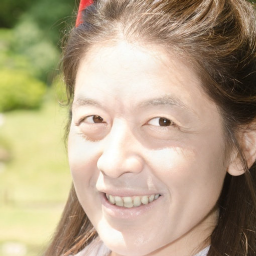} & \includegraphics[width=\pvftile]{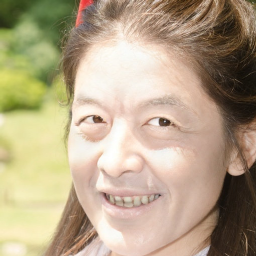} & \includegraphics[width=\pvftile]{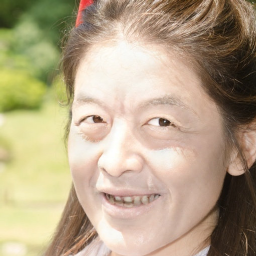} & \includegraphics[width=\pvftile]{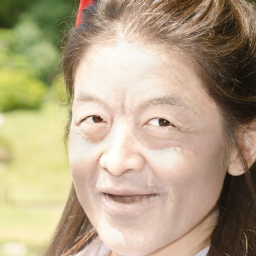} \\
  \pvfrow{Target} & \includegraphics[width=\pvftile]{figures/pulid_vs_family/p97_target.pdf} & \pvfkind{Trained}
  & \includegraphics[width=\pvftile]{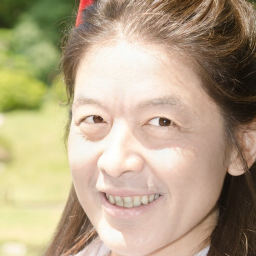} & \includegraphics[width=\pvftile]{figures/pulid_vs_family/p97_off_s3_w1p00.pdf} & \includegraphics[width=\pvftile]{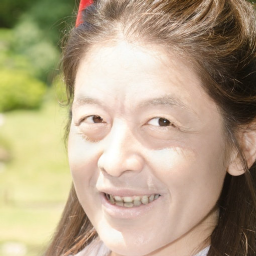} & \includegraphics[width=\pvftile]{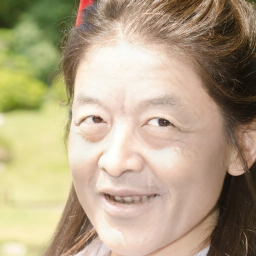} & \includegraphics[width=\pvftile]{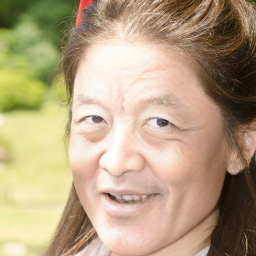} \\[3pt]
  \pvfrow{Source} & \includegraphics[width=\pvftile]{figures/pulid_vs_family/p441_source.pdf} & \pvfkind{PuLID}
  & \includegraphics[width=\pvftile]{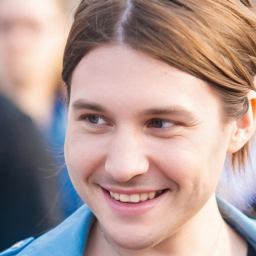} & \includegraphics[width=\pvftile]{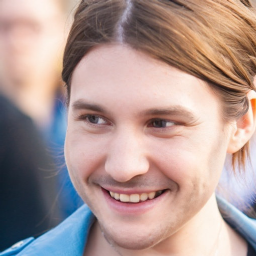} & \includegraphics[width=\pvftile]{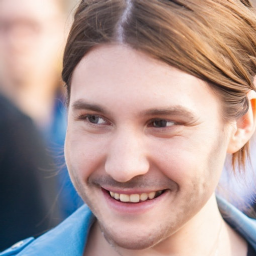} & \includegraphics[width=\pvftile]{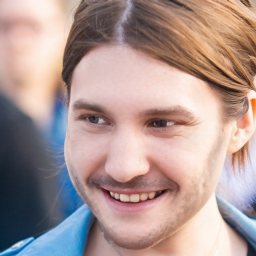} & \includegraphics[width=\pvftile]{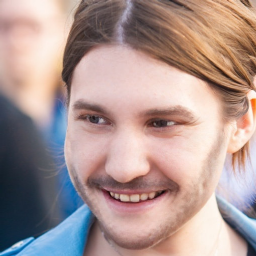} \\
  \pvfrow{Target} & \includegraphics[width=\pvftile]{figures/pulid_vs_family/p441_target.pdf} & \pvfkind{Trained}
  & \includegraphics[width=\pvftile]{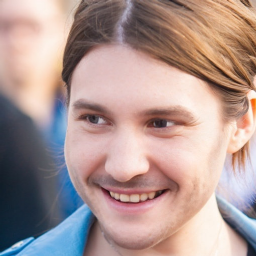} & \includegraphics[width=\pvftile]{figures/pulid_vs_family/p441_off_s3_w1p00.pdf} & \includegraphics[width=\pvftile]{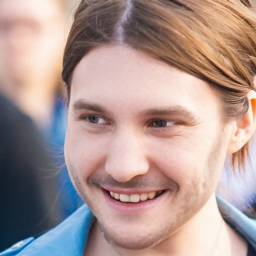} & \includegraphics[width=\pvftile]{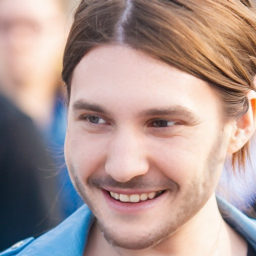} & \includegraphics[width=\pvftile]{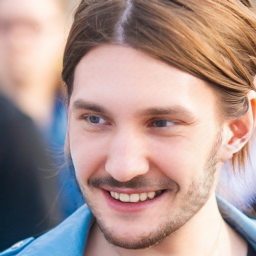} \\[3pt]
  \pvfrow{Source} & \includegraphics[width=\pvftile]{figures/pulid_vs_family/p803_source.pdf} & \pvfkind{PuLID}
  & \includegraphics[width=\pvftile]{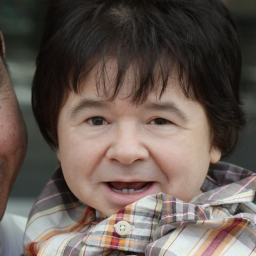} & \includegraphics[width=\pvftile]{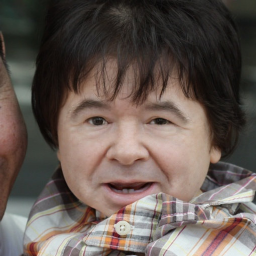} & \includegraphics[width=\pvftile]{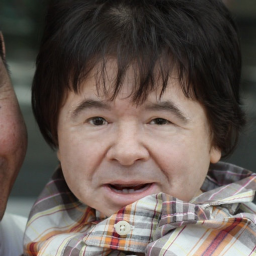} & \includegraphics[width=\pvftile]{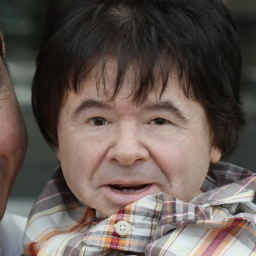} & \includegraphics[width=\pvftile]{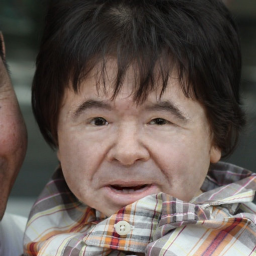} \\
  \pvfrow{Target} & \includegraphics[width=\pvftile]{figures/pulid_vs_family/p803_target.pdf} & \pvfkind{Trained}
  & \includegraphics[width=\pvftile]{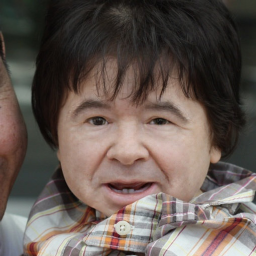} & \includegraphics[width=\pvftile]{figures/pulid_vs_family/p803_off_s3_w1p00.pdf} & \includegraphics[width=\pvftile]{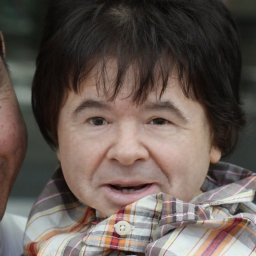} & \includegraphics[width=\pvftile]{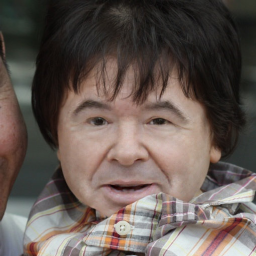} & \includegraphics[width=\pvftile]{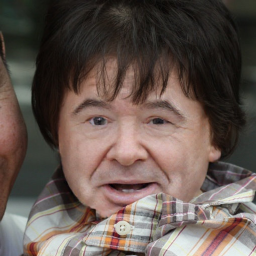} \\[3pt]
  \pvfrow{Source} & \includegraphics[width=\pvftile]{figures/pulid_vs_family/p824_source.pdf} & \pvfkind{PuLID}
  & \includegraphics[width=\pvftile]{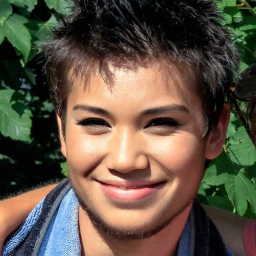} & \includegraphics[width=\pvftile]{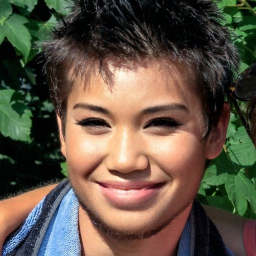} & \includegraphics[width=\pvftile]{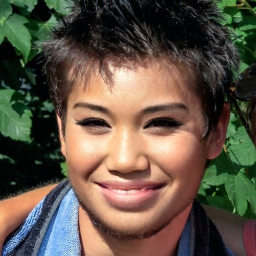} & \includegraphics[width=\pvftile]{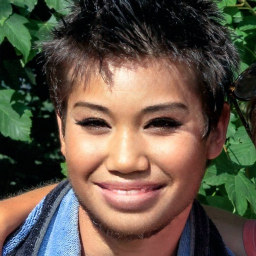} & \includegraphics[width=\pvftile]{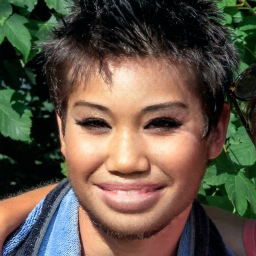} \\
  \pvfrow{Target} & \includegraphics[width=\pvftile]{figures/pulid_vs_family/p824_target.pdf} & \pvfkind{Trained}
  & \includegraphics[width=\pvftile]{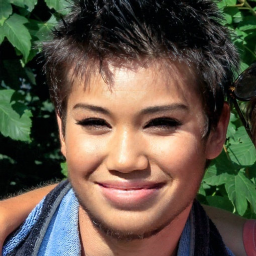} & \includegraphics[width=\pvftile]{figures/pulid_vs_family/p824_off_s3_w1p00.pdf} & \includegraphics[width=\pvftile]{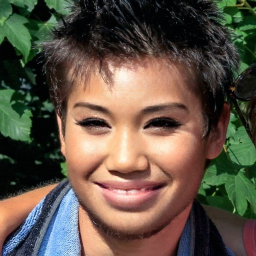} & \includegraphics[width=\pvftile]{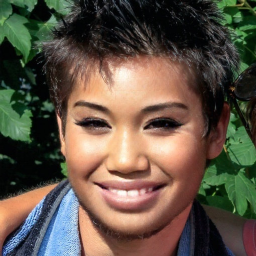} & \includegraphics[width=\pvftile]{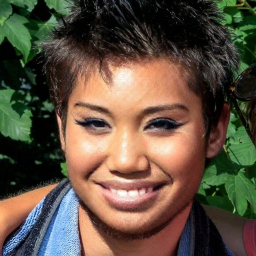} \\[3pt]

  \end{tabular}
  \caption{\textbf{PuLID weight sweep against the trained family, OminiControl branch off.} \pvfcaption{off}}
  \label{fig:pulid-vs-family-off}
\end{figure*}
\endgroup

%% file: figures/figure_ablation_examples.tex
\begingroup

\ifdefined\ablrealtile\else\newlength{\ablrealtile}\fi
\setlength{\ablrealtile}{0.138\textwidth}

\ifdefined\ablexprtile\else\newlength{\ablexprtile}\fi
\setlength{\ablexprtile}{0.185\textwidth}

\begin{figure*}[tbp]
  \centering
  \small
  \setlength{\tabcolsep}{1.5pt}
  \renewcommand{\arraystretch}{0.95}
  \begin{tabular}{@{}cc @{\hspace{6pt}} *{4}{c}@{}}
    Source & Target & None & DMD & DMD + GAN & GAN (Teacher) \\
      \includegraphics[width=\ablrealtile]{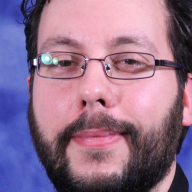} & \includegraphics[width=\ablrealtile]{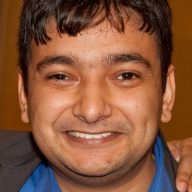}
  & \includegraphics[width=\ablrealtile]{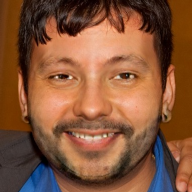} & \includegraphics[width=\ablrealtile]{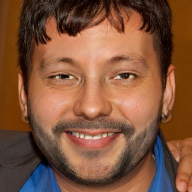}
  & \includegraphics[width=\ablrealtile]{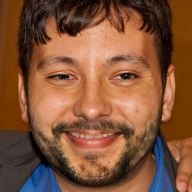} & \includegraphics[width=\ablrealtile]{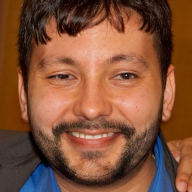} \\

      \includegraphics[width=\ablrealtile]{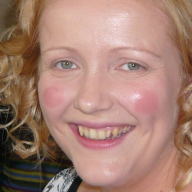} & \includegraphics[width=\ablrealtile]{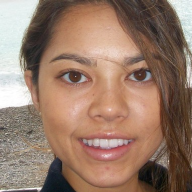}
  & \includegraphics[width=\ablrealtile]{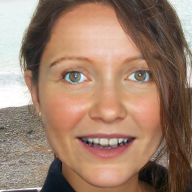} & \includegraphics[width=\ablrealtile]{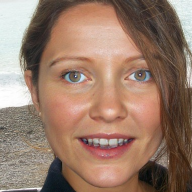}
  & \includegraphics[width=\ablrealtile]{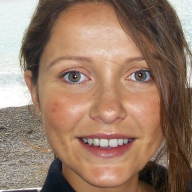} & \includegraphics[width=\ablrealtile]{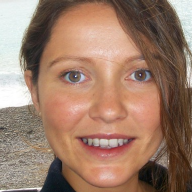} \\

      \includegraphics[width=\ablrealtile]{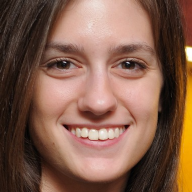} & \includegraphics[width=\ablrealtile]{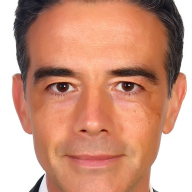}
  & \includegraphics[width=\ablrealtile]{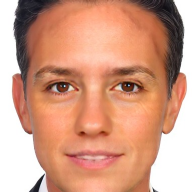} & \includegraphics[width=\ablrealtile]{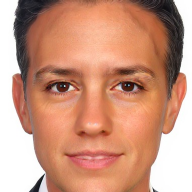}
  & \includegraphics[width=\ablrealtile]{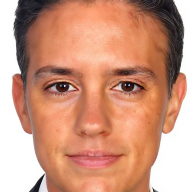} & \includegraphics[width=\ablrealtile]{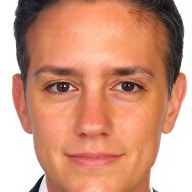} \\

  \end{tabular}
  \caption{\textbf{Qualitative comparison of teacher realism signals} on FFHQ (Table~\ref{tab:ablation}, bottom). Without the adversarial loss (None) or with DMD alone, skin and facial hair appear less natural than with DMD + GAN or the adversarial loss alone (GAN).}
  \label{fig:ablation-qual-realism}

  \vspace{1.2em}
  \begin{tabular}{@{}cc @{\hspace{6pt}} cc@{}}
    Source & Target & Full Teacher & w/o $\mathcal{L}_{\mathrm{expr}}$ \\
      \includegraphics[width=\ablexprtile]{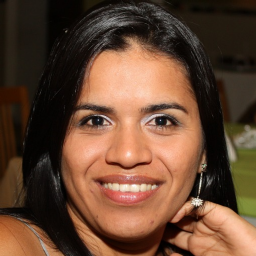} & \includegraphics[width=\ablexprtile]{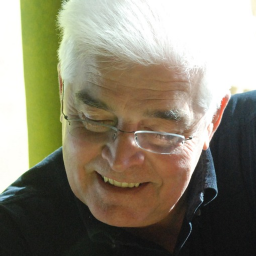}
  & \includegraphics[width=\ablexprtile]{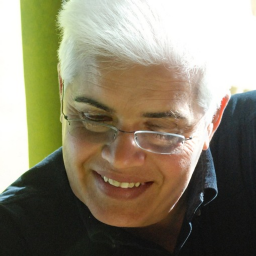} & \includegraphics[width=\ablexprtile]{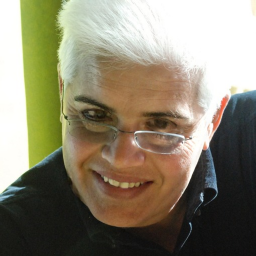} \\[-1pt]
  \includegraphics[width=\ablexprtile]{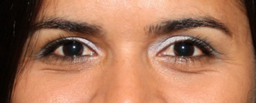} & \includegraphics[width=\ablexprtile]{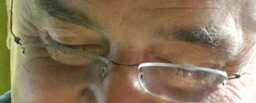}
  & \includegraphics[width=\ablexprtile]{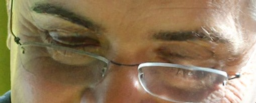} & \includegraphics[width=\ablexprtile]{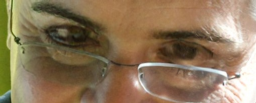} \\[4pt]

      \includegraphics[width=\ablexprtile]{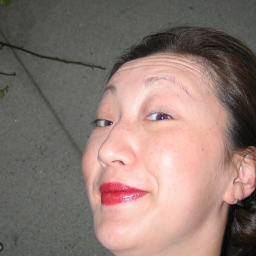} & \includegraphics[width=\ablexprtile]{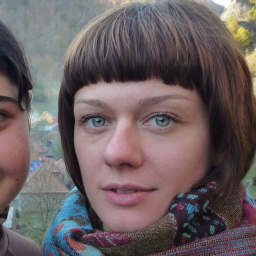}
  & \includegraphics[width=\ablexprtile]{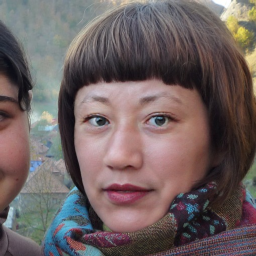} & \includegraphics[width=\ablexprtile]{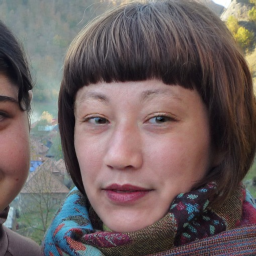} \\[-1pt]
  \includegraphics[width=\ablexprtile]{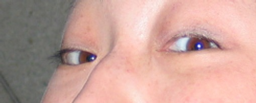} & \includegraphics[width=\ablexprtile]{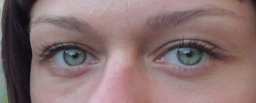}
  & \includegraphics[width=\ablexprtile]{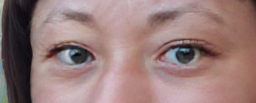} & \includegraphics[width=\ablexprtile]{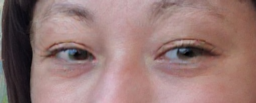} \\[4pt]

  \end{tabular}
  \caption{\textbf{Effect of the expression loss $\mathcal{L}_{\mathrm{expr}}$ on the eye region} (Table~\ref{tab:ablation-teacher-losses}). Each pair shows the full image above a zoomed view of the eyes. Without $\mathcal{L}_{\mathrm{expr}}$, the generator transfers the source's open eyelids (top) or sideways gaze and narrowed eyelids (bottom) instead of preserving the target's.}
  \label{fig:ablation-qual-eyes}
\end{figure*}

\ifdefined\ablprestile\else\newlength{\ablprestile}\fi
\setlength{\ablprestile}{0.118\textwidth}

\begin{figure*}[tbp]
  \centering
  \small
  \setlength{\tabcolsep}{1pt}
  \renewcommand{\arraystretch}{0.95}
  \begin{tabular}{@{}cc @{\hspace{5pt}} *{6}{c}@{}}
    Source & Target & Full Teacher
    & w/o $\mathcal{L}_{\mathrm{pose}}$
    & w/o $\mathcal{L}_{\mathrm{expr}}$
    & w/o $\mathcal{L}_{\mathrm{bg}}$
    & w/o all three
    & $\mathcal{L}_{\mathrm{id}} + \mathcal{L}_{\mathrm{GAN}}$ only \\
      \includegraphics[width=\ablprestile]{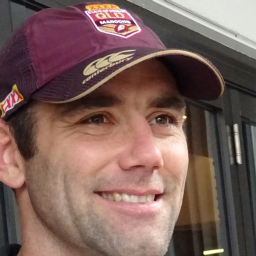} & \includegraphics[width=\ablprestile]{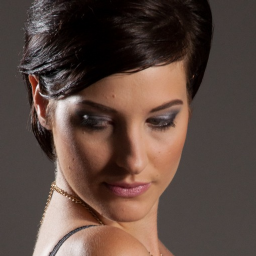}
  & \includegraphics[width=\ablprestile]{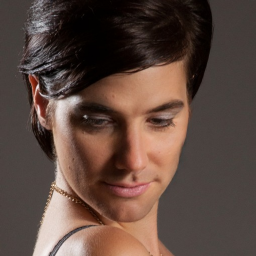} & \includegraphics[width=\ablprestile]{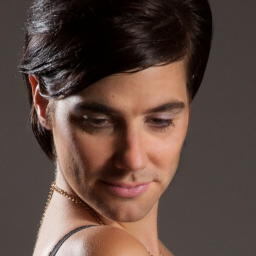}
  & \includegraphics[width=\ablprestile]{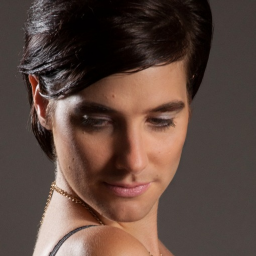} & \includegraphics[width=\ablprestile]{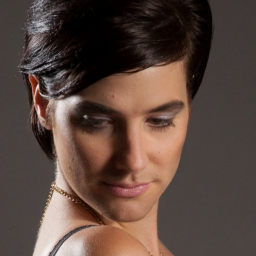}
  & \includegraphics[width=\ablprestile]{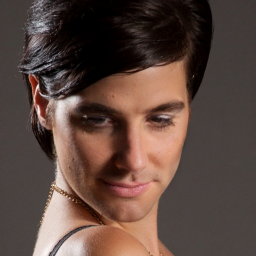} & \includegraphics[width=\ablprestile]{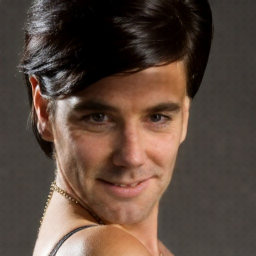} \\

      \includegraphics[width=\ablprestile]{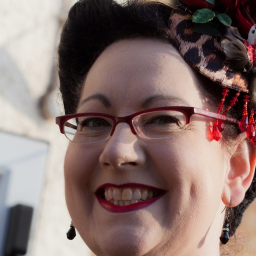} & \includegraphics[width=\ablprestile]{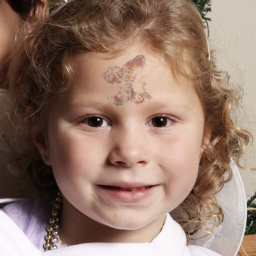}
  & \includegraphics[width=\ablprestile]{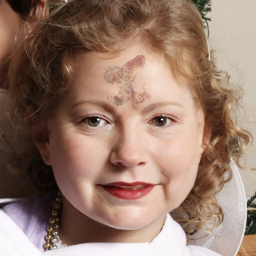} & \includegraphics[width=\ablprestile]{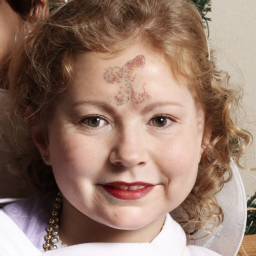}
  & \includegraphics[width=\ablprestile]{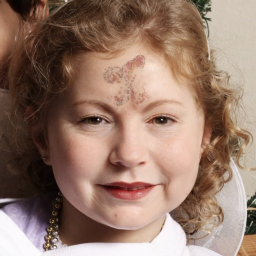} & \includegraphics[width=\ablprestile]{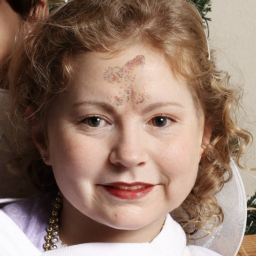}
  & \includegraphics[width=\ablprestile]{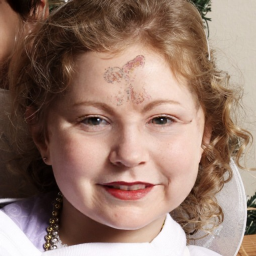} & \includegraphics[width=\ablprestile]{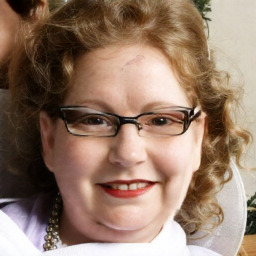} \\

      \includegraphics[width=\ablprestile]{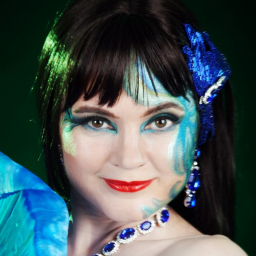} & \includegraphics[width=\ablprestile]{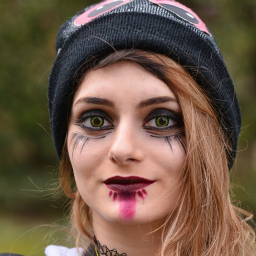}
  & \includegraphics[width=\ablprestile]{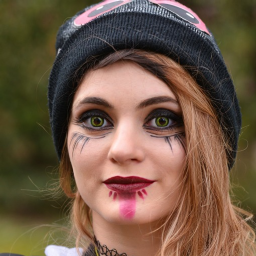} & \includegraphics[width=\ablprestile]{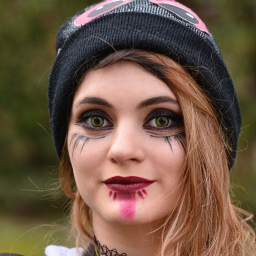}
  & \includegraphics[width=\ablprestile]{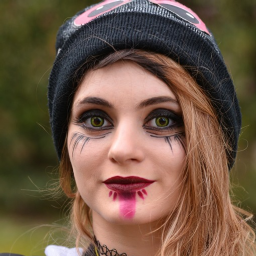} & \includegraphics[width=\ablprestile]{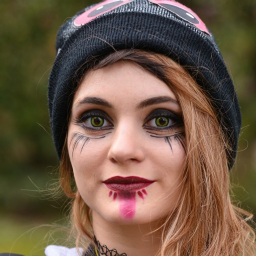}
  & \includegraphics[width=\ablprestile]{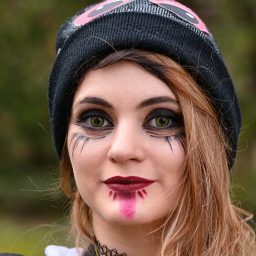} & \includegraphics[width=\ablprestile]{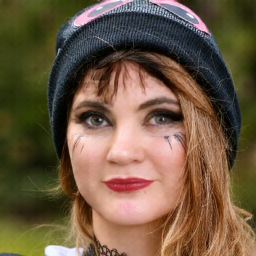} \\

      \includegraphics[width=\ablprestile]{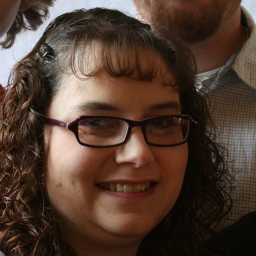} & \includegraphics[width=\ablprestile]{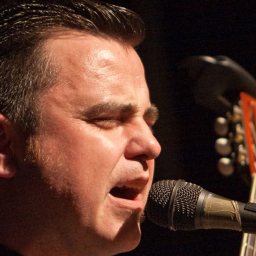}
  & \includegraphics[width=\ablprestile]{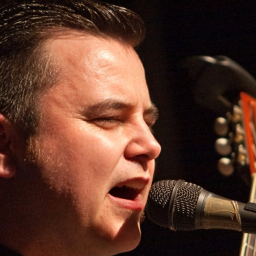} & \includegraphics[width=\ablprestile]{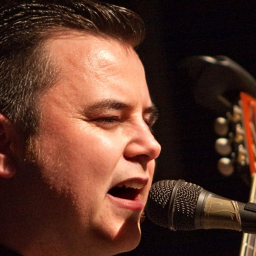}
  & \includegraphics[width=\ablprestile]{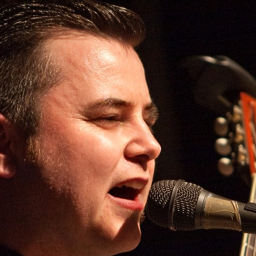} & \includegraphics[width=\ablprestile]{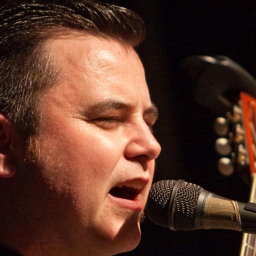}
  & \includegraphics[width=\ablprestile]{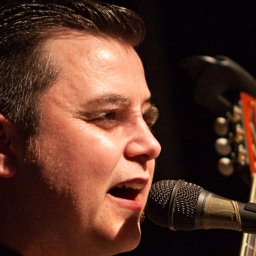} & \includegraphics[width=\ablprestile]{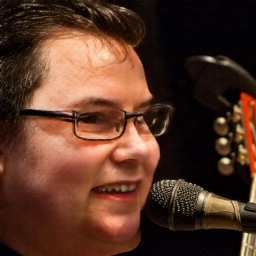} \\

  \end{tabular}
  \caption{\textbf{Qualitative ablation of the Teacher preservation losses} on FFHQ (Table~\ref{tab:ablation-teacher-losses}). Removing a single preservation loss (w/o $\mathcal{L}_{\mathrm{pose}}$, w/o $\mathcal{L}_{\mathrm{expr}}$, or w/o $\mathcal{L}_{\mathrm{bg}}$) produces little visible change outside subtle eyelid differences. Removing all three preservation losses at once (w/o all three) weakens gaze, make-up, and expression retention, and dropping the same-identity reconstruction batches as well ($\mathcal{L}_{\mathrm{id}} + \mathcal{L}_{\mathrm{GAN}}$ only) loses target pose, expression, make-up, and accessories.}
  \label{fig:ablation-qual-preservation}
\end{figure*}

\endgroup

%% file: figures/figure_curated_grids.tex
\begingroup
\ifdefined\cgtile\else\newlength{\cgtile}\fi
\newcommand{\cgband}[2]{%
  \raisebox{\dimexpr -0.5\cgtile*(#1-1)-0.4ex\relax}[0pt][0pt]{\rotatebox[origin=c]{90}{\makebox[#1\cgtile][c]{\footnotesize #2}}}%
}
\newcommand{\cgheadA}{%
    & & & \multicolumn{3}{c}{Ours \MethodName} & & & & & & & \\
    \cmidrule(lr){4-6}
    & Source & Target
    & Teacher$^{\ddagger}$ & $\lambda_\text{rec}=1$ & $\lambda_\text{rec}=3$
    & InSwapper & CSCS & SimSwap & \shortstack{Face-\\Dancer} & InfoSwap & REFace & APPLE \\[1pt]
}
\newcommand{\cgcaption}{Face swapping results on FFHQ compared with existing baselines. Teacher$^{\ddagger}$: OminiControl branch off.}

\begin{figure*}[p]
  \centering\footnotesize
  \setlength{\cgtile}{0.077\textwidth}
  \setlength{\tabcolsep}{0.5pt}
  \renewcommand{\arraystretch}{0.95}
\begin{tabular}{@{}l cc @{\hspace{5pt}} *{10}{c}@{}}
    \cgheadA
    \cgband{8}{Make-up / Accessories}
      & \IfFileExists{figures/curated_examples/p63_source.pdf}{\includegraphics[width=\cgtile]{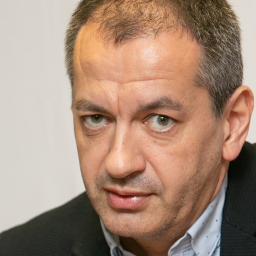}}{\rule{\cgtile}{\cgtile}} & \IfFileExists{figures/curated_examples/p63_target.pdf}{\includegraphics[width=\cgtile]{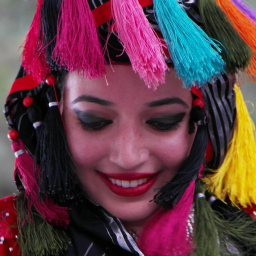}}{\rule{\cgtile}{\cgtile}}
  & \IfFileExists{figures/curated_examples/p63_teacher_ref_off.pdf}{\includegraphics[width=\cgtile]{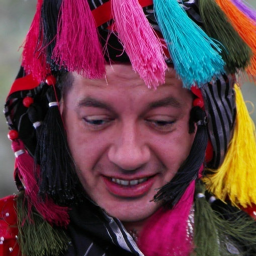}}{\rule{\cgtile}{\cgtile}} & \IfFileExists{figures/curated_examples/p63_ladd_lpips1.pdf}{\includegraphics[width=\cgtile]{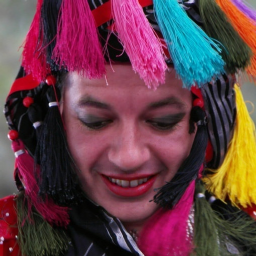}}{\rule{\cgtile}{\cgtile}} & \IfFileExists{figures/curated_examples/p63_ladd_lpips3.pdf}{\includegraphics[width=\cgtile]{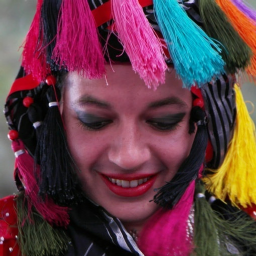}}{\rule{\cgtile}{\cgtile}}
  & \IfFileExists{figures/curated_examples/p63_inswapper.pdf}{\includegraphics[width=\cgtile]{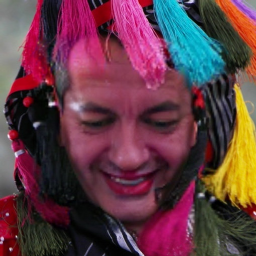}}{\rule{\cgtile}{\cgtile}} & \IfFileExists{figures/curated_examples/p63_cscs.pdf}{\includegraphics[width=\cgtile]{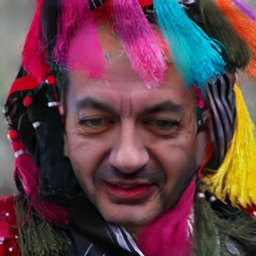}}{\rule{\cgtile}{\cgtile}} & \IfFileExists{figures/curated_examples/p63_simswap.pdf}{\includegraphics[width=\cgtile]{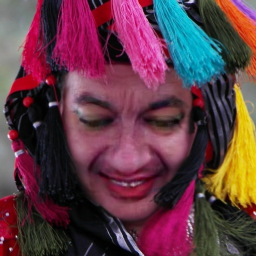}}{\rule{\cgtile}{\cgtile}} & \IfFileExists{figures/curated_examples/p63_facedancer.pdf}{\includegraphics[width=\cgtile]{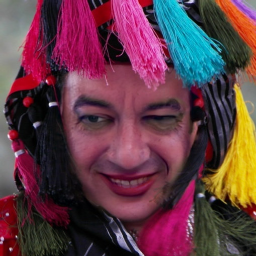}}{\rule{\cgtile}{\cgtile}}
  & \IfFileExists{figures/curated_examples/p63_infoswap.pdf}{\includegraphics[width=\cgtile]{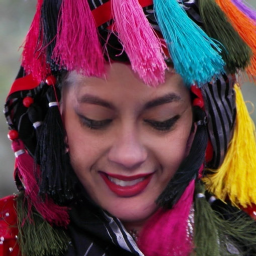}}{\rule{\cgtile}{\cgtile}} & \IfFileExists{figures/curated_examples/p63_reface.pdf}{\includegraphics[width=\cgtile]{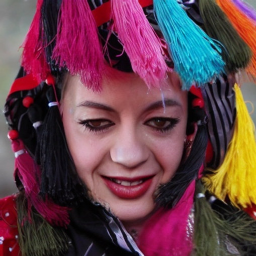}}{\rule{\cgtile}{\cgtile}} & \IfFileExists{figures/curated_examples/p63_apple.pdf}{\includegraphics[width=\cgtile]{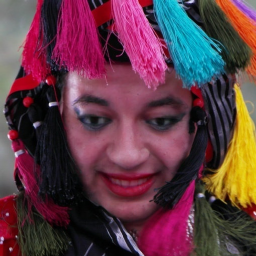}}{\rule{\cgtile}{\cgtile}} \\
\end{tabular}
  \caption{\textbf{Qualitative results.} \cgcaption}
  \label{fig:curated-grid}
\end{figure*}

\begin{figure*}[p]
  \centering\footnotesize
  \setlength{\cgtile}{0.077\textwidth}
  \setlength{\tabcolsep}{0.5pt}
  \renewcommand{\arraystretch}{0.95}
  \begin{tabular}{@{}l cc @{\hspace{5pt}} *{10}{c}@{}}
    \cgheadA
    \cgband{14}{Naturalness / Quality}
      & \IfFileExists{figures/curated_examples/p263_source.pdf}{\includegraphics[width=\cgtile]{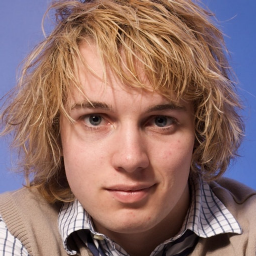}}{\rule{\cgtile}{\cgtile}} & \IfFileExists{figures/curated_examples/p263_target.pdf}{\includegraphics[width=\cgtile]{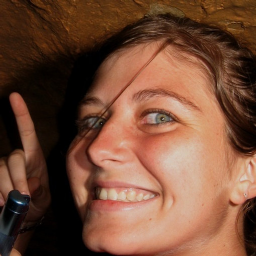}}{\rule{\cgtile}{\cgtile}}
  & \IfFileExists{figures/curated_examples/p263_teacher_ref_off.pdf}{\includegraphics[width=\cgtile]{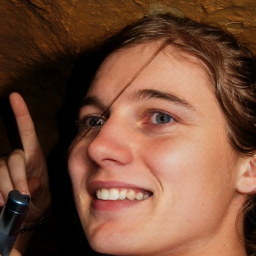}}{\rule{\cgtile}{\cgtile}} & \IfFileExists{figures/curated_examples/p263_ladd_lpips1.pdf}{\includegraphics[width=\cgtile]{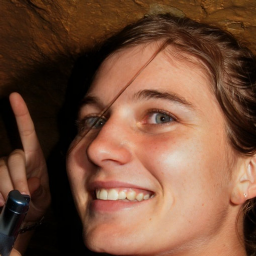}}{\rule{\cgtile}{\cgtile}} & \IfFileExists{figures/curated_examples/p263_ladd_lpips3.pdf}{\includegraphics[width=\cgtile]{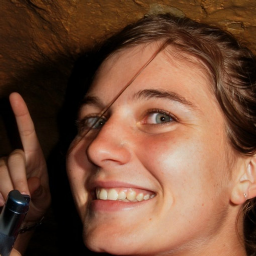}}{\rule{\cgtile}{\cgtile}}
  & \IfFileExists{figures/curated_examples/p263_inswapper.pdf}{\includegraphics[width=\cgtile]{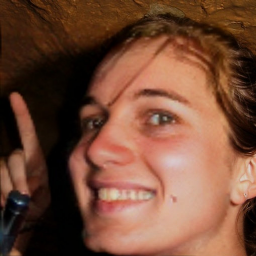}}{\rule{\cgtile}{\cgtile}} & \IfFileExists{figures/curated_examples/p263_cscs.pdf}{\includegraphics[width=\cgtile]{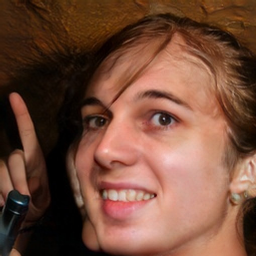}}{\rule{\cgtile}{\cgtile}} & \IfFileExists{figures/curated_examples/p263_simswap.pdf}{\includegraphics[width=\cgtile]{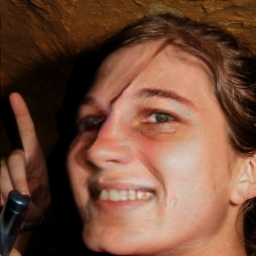}}{\rule{\cgtile}{\cgtile}} & \IfFileExists{figures/curated_examples/p263_facedancer.pdf}{\includegraphics[width=\cgtile]{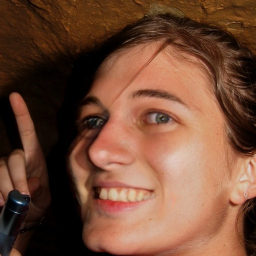}}{\rule{\cgtile}{\cgtile}}
  & \IfFileExists{figures/curated_examples/p263_infoswap.pdf}{\includegraphics[width=\cgtile]{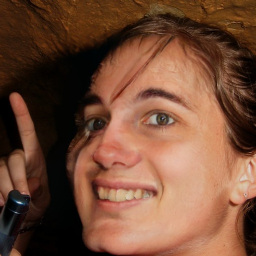}}{\rule{\cgtile}{\cgtile}} & \IfFileExists{figures/curated_examples/p263_reface.pdf}{\includegraphics[width=\cgtile]{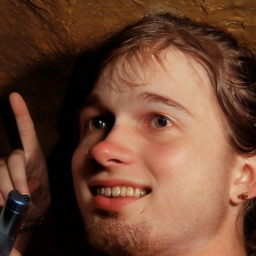}}{\rule{\cgtile}{\cgtile}} & \IfFileExists{figures/curated_examples/p263_apple.pdf}{\includegraphics[width=\cgtile]{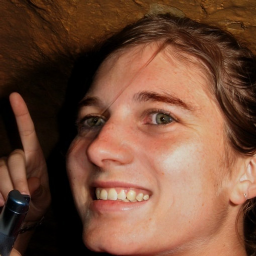}}{\rule{\cgtile}{\cgtile}} \\
\end{tabular}
  \caption{\textbf{Qualitative results.} \cgcaption}
  \label{fig:curated-grid-2}
\end{figure*}
\twocolumn[{
  \centering\footnotesize\vspace{-6pt}
  \setlength{\cgtile}{0.077\textwidth}
  \setlength{\tabcolsep}{0.5pt}
  \renewcommand{\arraystretch}{0.95}
  \begin{tabular}{@{}l cc @{\hspace{5pt}} *{10}{c}@{}}
    \cgheadA
    \cgband{4}{Glasses}
      & \IfFileExists{figures/curated_examples/p127_source.pdf}{\includegraphics[width=\cgtile]{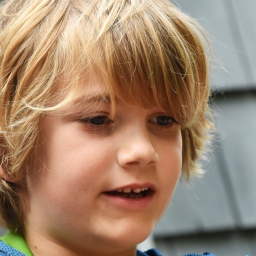}}{\rule{\cgtile}{\cgtile}} & \IfFileExists{figures/curated_examples/p127_target.pdf}{\includegraphics[width=\cgtile]{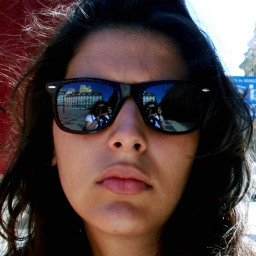}}{\rule{\cgtile}{\cgtile}}
  & \IfFileExists{figures/curated_examples/p127_teacher_ref_off.pdf}{\includegraphics[width=\cgtile]{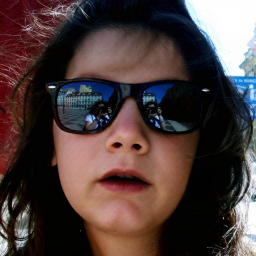}}{\rule{\cgtile}{\cgtile}} & \IfFileExists{figures/curated_examples/p127_ladd_lpips1.pdf}{\includegraphics[width=\cgtile]{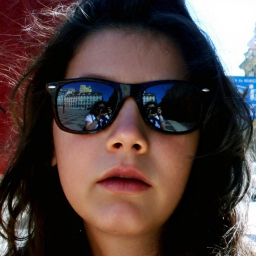}}{\rule{\cgtile}{\cgtile}} & \IfFileExists{figures/curated_examples/p127_ladd_lpips3.pdf}{\includegraphics[width=\cgtile]{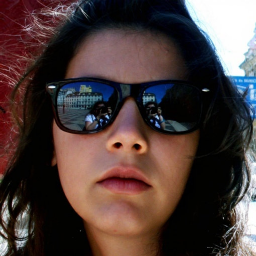}}{\rule{\cgtile}{\cgtile}}
  & \IfFileExists{figures/curated_examples/p127_inswapper.pdf}{\includegraphics[width=\cgtile]{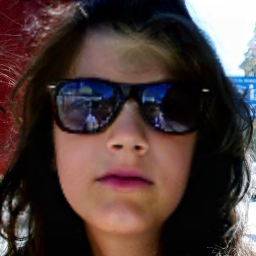}}{\rule{\cgtile}{\cgtile}} & \IfFileExists{figures/curated_examples/p127_cscs.pdf}{\includegraphics[width=\cgtile]{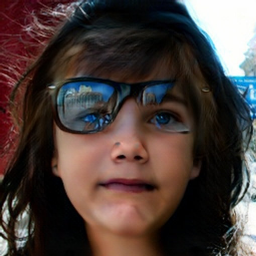}}{\rule{\cgtile}{\cgtile}} & \IfFileExists{figures/curated_examples/p127_simswap.pdf}{\includegraphics[width=\cgtile]{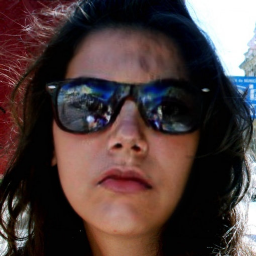}}{\rule{\cgtile}{\cgtile}} & \IfFileExists{figures/curated_examples/p127_facedancer.pdf}{\includegraphics[width=\cgtile]{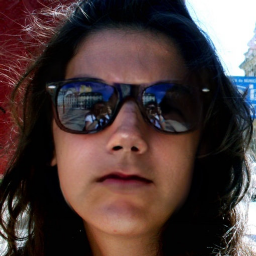}}{\rule{\cgtile}{\cgtile}}
  & \IfFileExists{figures/curated_examples/p127_infoswap.pdf}{\includegraphics[width=\cgtile]{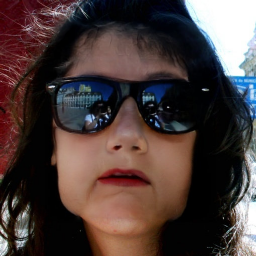}}{\rule{\cgtile}{\cgtile}} & \IfFileExists{figures/curated_examples/p127_reface.pdf}{\includegraphics[width=\cgtile]{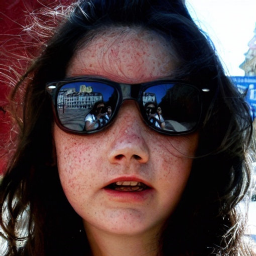}}{\rule{\cgtile}{\cgtile}} & \IfFileExists{figures/curated_examples/p127_apple.pdf}{\includegraphics[width=\cgtile]{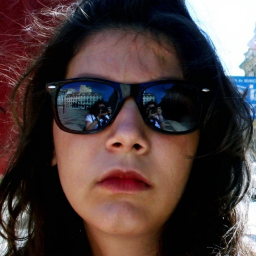}}{\rule{\cgtile}{\cgtile}} \\
\end{tabular}
    \par\vspace{6pt}
  \captionof{figure}{\textbf{Qualitative results.} \cgcaption}
  \label{fig:curated-grid-3}
  \vspace{12pt}
}]
\endgroup